# A Novel Approach for Protein Structure Prediction

The Project Report is submitted in partial fulfillment of the requirements for the award of the degree of

**BACHELOR OF TECHNOLOGY**

Submitted By:

**(Saurabh Sarkar, Prateek Malhotra, Virender Guman)**

Department of Computer Science & Engineering

G.L.Bajaj Institute of Technology & Mgmt.
Greater Noida

Uttar Pradesh Technical University
Lucknow (U. P.) India

# A Novel Approach for Protein Structure Prediction

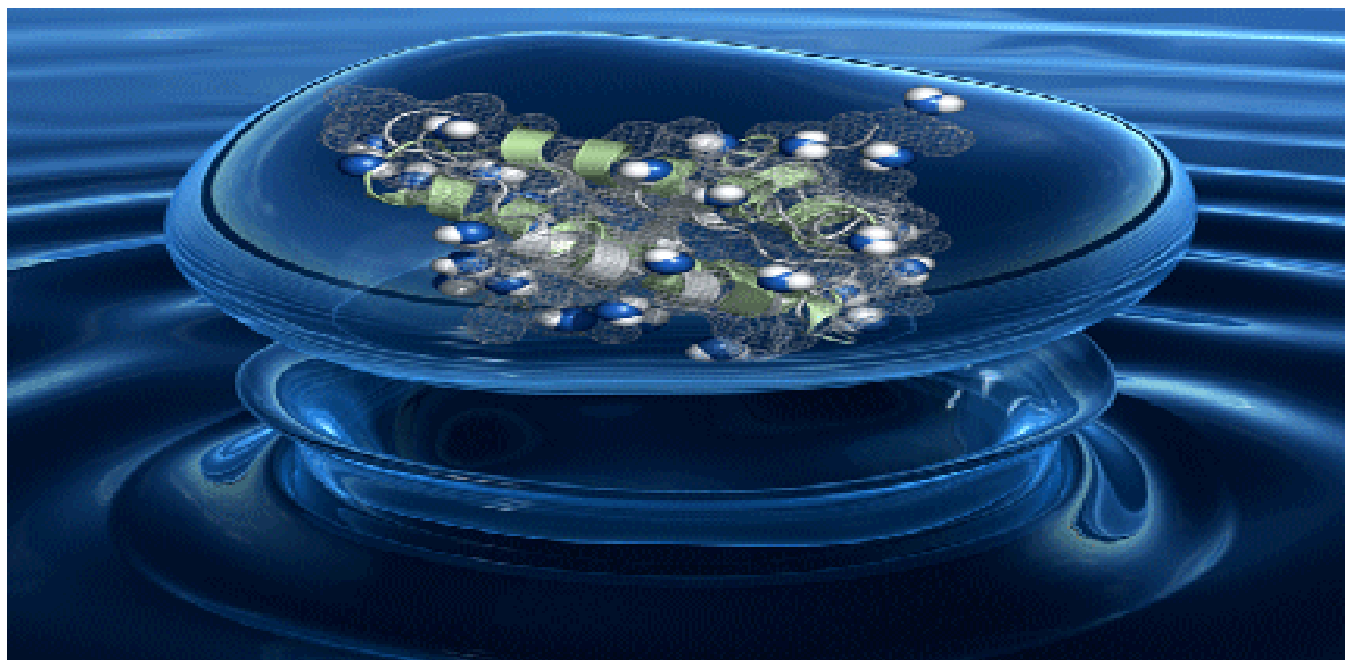

The Project Report is submitted in partial fulfillment of the requirements for the award of the degree of

**BACHELOR OF TECHNOLOGY**

Submitted By:

**(Saurabh Sarkar, Prateek Malhotra, Virender Guman)**

Department of Computer Science & Engineering

G.L.Bajaj Institute of Technology & Mgmt.
Greater Noida

Uttar Pradesh Technical University
Lucknow (U. P.) India

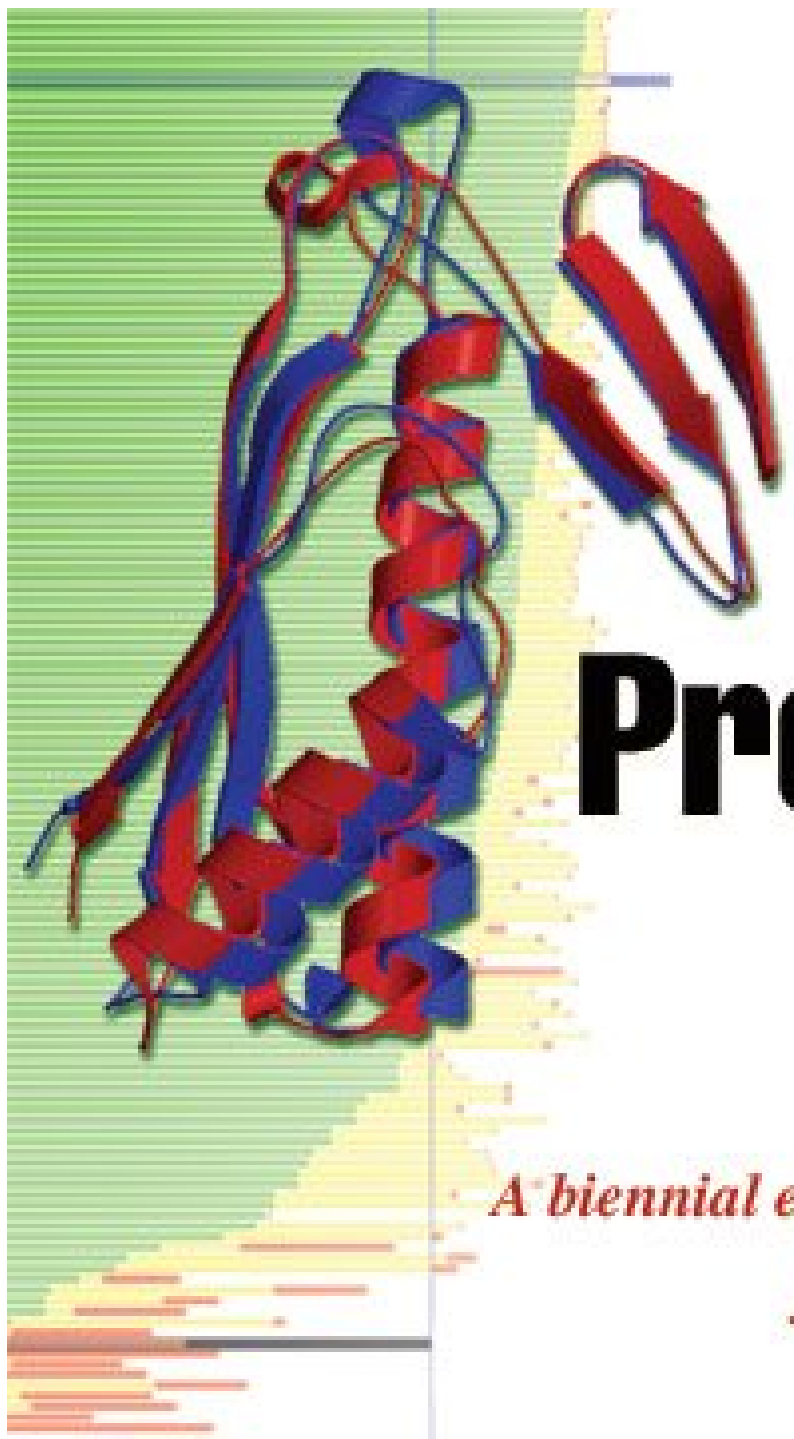

# The Art of Protein Structure Prediction

*A biennial experiment helps scientists evaluate the best methods for predicting the structures of proteins.*

# CERTIFICATE

This is to certify that the project report entitled "A Novel Approach for Protein Structure Prediction" submitted by by Saurabh Sarkar, Prateek Malhortra, Virender Guman in the Department of Computer Science & Engineering, G.L.Bajaj Institute of Technology and Management, Greater Noida for the award of Bachelor of Technology is a record of the bona-fide work carried out by them under my supervision.

**Details of the student(s)**

Mr. Saurabh Sarkar (0619210099)

Mr. Prateek Malhotra (0619210083)

Mr. Virender Guman (0619210118)

**Certified by:**

Prof. K.S. Mehta
HOD (CSE Department)

**Department of the Computer Science & Engineering**
**G.L.B.I.T.M. Greater Noida**

## ACKNOWLEDGEMENT

At the very outset, we are highly indebted to G.L. Bajaj Instt. Of Tech. & Mgmt., Gr. Noida for giving me an opportunity to carry out my project on **"A Novel Approach for Protein Structure Prediction"** at their esteemed institute. We would like to thank our faculty supervisor Mr. Saurabh Singh for his able guidance and support during the course of the project. He gave us the freedom to carry on our research, for which we are highly indebted to him. We would also like to express our profound sense of gratitude to Asst. Prof. Manish Singh (CS Dept., GCET, Gr. Noida) for his valuable guidance during our course of project.

We would like to extend our special thanks to the HOD of CSE Deptt. Professor K.S. Mehta and management GLBITM Gr. Noida for providing us the opportunity to work on this project, and the support staff for their help and cooperation.

**Details of the student(s)**

Mr. Saurabh Sarkar (0619210099)

Mr. Prateek Malhotra (0619210083)

Mr. Virender Guman (0619210118)

**Supervisor:**

Mr. Saurabh Singh
(CSE Department)

## ABSTRACT

The idea of this project is to study the protein structure and sequence relationship using the hidden markov model and artificial neural network. In this context we have assumed two hidden markov models. In first model we have taken protein secondary structures as hidden and protein sequences as observed. In second model we have taken protein sequences as hidden and protein structures as observed. The efficiencies for both the hidden markov models have been calculated. The results show that the efficiencies of first model is greater that the second one .These efficiencies are cross validated using artificial neural network. This signifies the importance of protein secondary structures as the main hidden controlling factors due to which we observe a particular amino acid sequence. This also signifies that protein secondary structure is more conserved in comparison to amino acid sequence.

## LIST OF FIGURES

## LIST OF TABLES

# ABBREVIATIONS

HMM Hidden markov Model

ANN Artificial Neural Network

DSSP Dictionary of Protein Secondary Structure

DNA Deoxy Ribonucleic Acid

RNA Ribonucleic Acid

3D 3 Dimensional

SD Standard Deviation

Q3 Score for three hidden states

# INDEX

# SYNOPSIS

## 1.1 INTRODUCTION AND OBJECTIVE

**Protein structure prediction** is the prediction of the three-dimensional structure of a protein from its amino acid sequence—that is, the prediction of a protein's tertiary structure from its primary structure (structure prediction is fundamentally different from the inverse problem of protein design). Protein structure prediction is one of the most important goals pursued by bioinformatics.

The information for life of most of the organisms is stored in genes by the four different types of nucleotides. Proteins are the other important macromolecules that do all important tasks in all organisms, like catalysis of biochemical reactions, transport of nutrients, recognition and transmission of signals. In this way genes are the blueprint of life and it can be said that the proteins are the functional unit or the machinery of life. Proteins are made up of amino acids joined in a long stretched chain. This long stretched chain is called as primary sequence, a translation of the genes into 20 types of amino acids. In water, the chain folds up to a specific three-dimensional (3D) structure. The main driving force is the necessity to pack residues for which a contact with water is energetically unfavorable into the interior of the molecule.

The three dimensional structure of a protein determines its function and it is well known that the details of the three dimensional structure are uniquely determined by the specificity of the sequence. In principle it can be said that the code could by cracked by calculating the physico-chemical force fields determining the fold. But the required computer time to calculate the three dimensional structure based on this principle is many orders of magnitude. However, it is of practical importance to know the three dimensional structure.

The exchange of a few amino acids can already destabilize a protein this means that if the length of a protein sequence is N the possible number of protein structures will be 20 raised to power N. But nature has not created this number of variety. Random errors in the DNA level information lead to a different translation of proteins. These 'errors' are the basis of evolution. Mutations resulting in a structural change are not likely to be accepted because of it the protein cannot perform its task, if protein can perform its task the mutation will be accepted.

The evolutionary pressures to conserve function and the discontinuity of the universe of structures have the result that the structure is evolutionarily more conserved than sequence.

The idea in this project is to explore the relationship of protein sequence and structure using statistical methods like hidden markov models and artificial neural network.

To explore the relationship two hidden markov models have been assumed:

- In the first model protein secondary structure has been assumed the main controlling factor to determine the protein sequence.
- In the second model protein sequence has been assumed the main controlling factor to determine the protein secondary structure.

The efficiencies for both the models will be calculated and compared. These efficiencies are cross validated using artificial neural network. If the first model will have greater efficiencies in comparison to second model it will show that protein secondary structure is the main hidden controlling factor to determine the protein sequence thus it will state that protein secondary structure is more conserved than protein sequence. If the second model will have greater efficiencies the conclusions will be just opposite.

# LITERATURE SURVEY

## 2.1 INTRODUCTION TO PROTEINS

Proteins are very common in all organisms and are fundamental unit of life. The diversity of protein structure underlies the very large range of their function like enzyme catalysis, transport and storage, coordinated motions, mechanical supports, immune protection, generation and transmission of nerve impulses and control of growth and differentiation.

1. Commonly all the biological reactions are catalyzed by the enzymes. Enzymes are well known to increase the rate of the biological reactions up to the 1 million fold. The number of enzymes identified till today reaches up to several thousands.

2. Proteins play a significant role in transport of the small molecules that are important for the physiological system, for example the transport of the oxygen to the tissues is carried out by the protein hemoglobin. There are many drug molecules that partially bound to serum albumins in the plasma.

3. Muscles are mostly composed of proteins and the mechanism of muscle contraction is mediated by the actin and myosin filaments those are proteins.

4. Proteins also take part to give mechanical support as skin and bone are strengthened by the protein collagen.

5. Proteins are important for the immune system as antibodies are proteins and they are responsible for reacting with specific foreign substances in the body.

6. Some amino acids work as neurotransmitters which transmit electrical signals from one nerve cell to another. Receptors for neurotransmitters, drugs, etc. are protein in nature acetylcholine receptor is a good example for this which is a protein structure that remains embedded in postsynaptic neurons.

7. Proteins play a critical role in the control of growth, cell differentiation and expression of DNA. For example, repressor proteins may bind to specific segments of DNA. In this way it can prevent the expression and thus the formation of the product of that particular DNA segment. Also, many growth factors and hormones that regulate cell function are proteins such as insulin or thyroid stimulating hormone.

Proteins are an important class of biological macromolecules present in all biological organisms, made up primarily from the elements carbon, hydrogen, nitrogen, oxygen, and sulphur. All proteins are polymers of amino acids. Classified by their physical size, proteins are nanoparticles (definition: 1-100 nm). Each protein polymer – also known as a polypeptide – consists of a sequence of 20 different L-α-amino acids, also referred to as residues. To understand the functions of proteins at a molecular level, it is often necessary to determine their three-dimensional structure.

A certain number of residues is necessary to perform a particular biochemical function, and around 40-50 residues appears to be the lower limit for a functional domain size. Protein sizes range from this lower limit to several thousand residues in multi-functional or structural proteins.

## 2.2 PROTEIN STRUCTURE

Proteins are macromolecules made up from 20 different L-α-amino acids which fold into a particular three-dimensional structure that is unique to each protein. This three-dimensional structure is responsible for the function of a protein. Thus in order to understand the details of protein functions it is necessary to understand protein structure. Protein structure is discussed in terms of four levels of organization.

Primary structure is the amino acid sequence of its polypeptide chains. Proteins are large polypeptides of defined amino acid sequence. The particular sequence of amino acids in protein is determined by the gene that encodes it. The gene is transcribed into a messenger ribonucleic acid (mRNA) and the mRNA is translated into a protein with the help of ribosome.

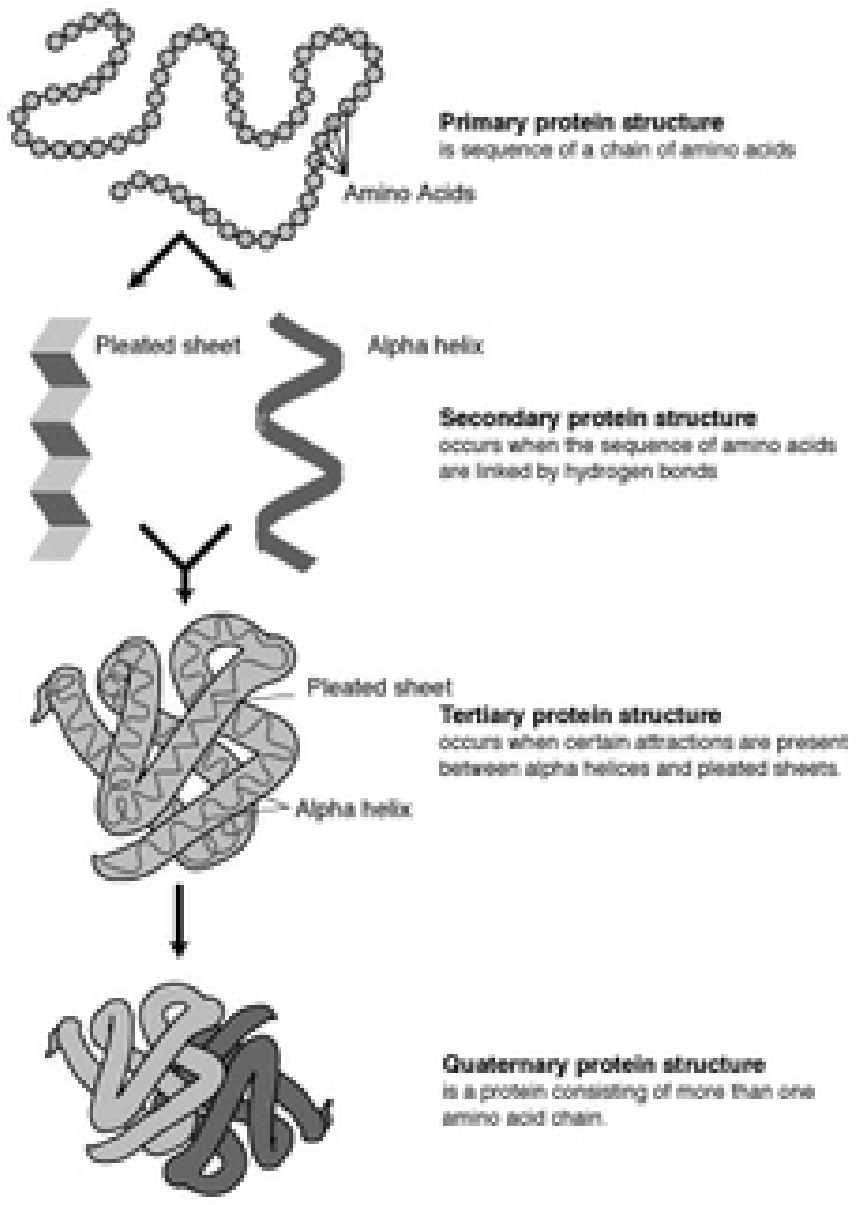

**Levels of protein structure**

**Protein structure**, from primary to quaternary structure**.**

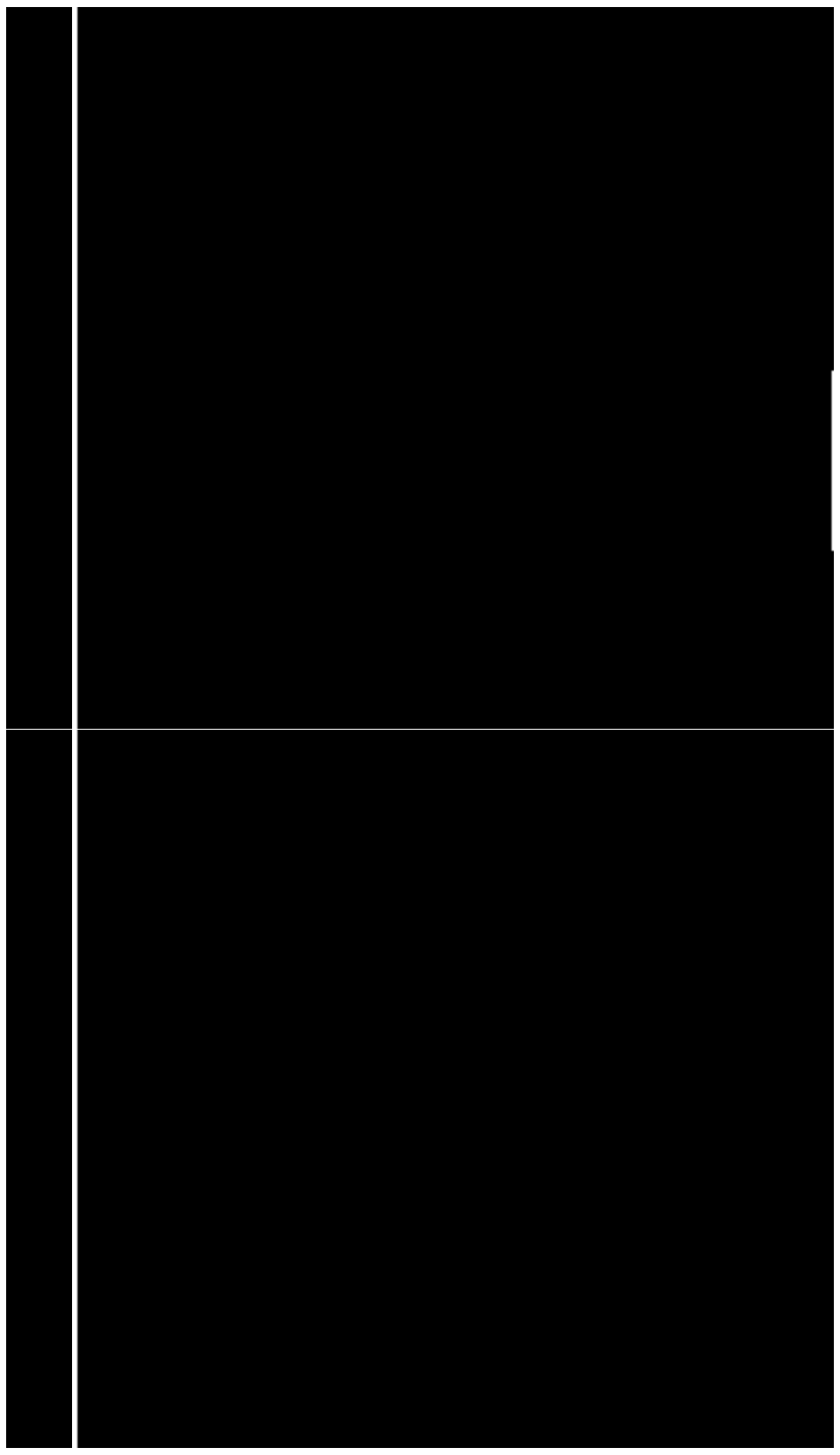

**Primary structure**

the amino acid sequence of the peptide chains.

The sequence of the different amino acids is called the primary structure of the peptide or protein. Counting of residues always starts at the N-terminal end (NH2-group), which is the end where the amino group is involved in a peptide bond. The primary structure of a protein is determined by the gene corresponding to the protein. A specific sequence of nucleotides in DNA is transcribed.

Primary structure is also called as the covalent structure of proteins because all of the covalent bonding within proteins defines the primary structure except disulfide bonds. In contrast, the higher level organizations of protein structure like secondary, tertiary and quaternary structure involve mainly noncovalent interactions.

Peptides (from the Greek πεπτίδια, "small digestibles") are short polymers formed from the linking, in a defined order, of α-amino acids. The link between one amino acid residue and the next is called an amide bond or a peptide bond. Proteins are polypeptide molecules, or consist of multiple polypeptide subunits, each composed of chains containing a specific sequence of the 22 proteinogenic amino acids. The distinction is that peptides are short whereas polypeptides are long.

**Structure of the amino acids**

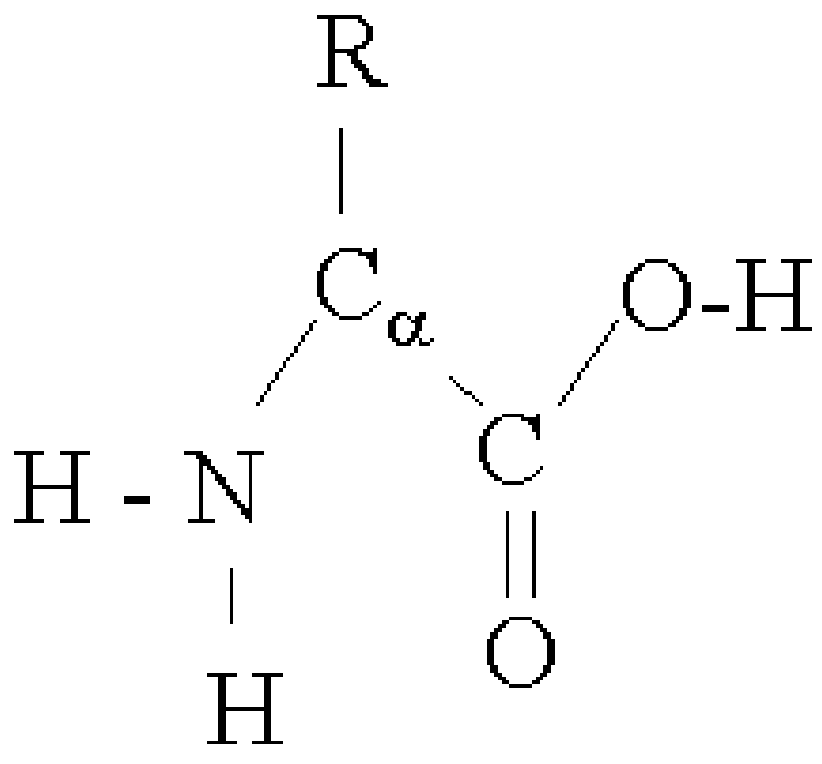

**An α-amino acid**

An α-amino acid consists of a part that is present in all the amino acid types, and a side chain that is unique to each type of residue. The Cα atom is bound to 4 different atoms: a hydrogen atom (the H is omitted in the diagram), an amino group nitrogen, a carboxyl group carbon, and a side chain carbon specific for this type of amino acid.
The side chain determines the chemical properties of the α-amino acid and may be any one of the 20 different side chains:
The 20 naturally occurring amino acids can be divided into several groups based on their chemical properties.

**The peptide bond (amide bond)**

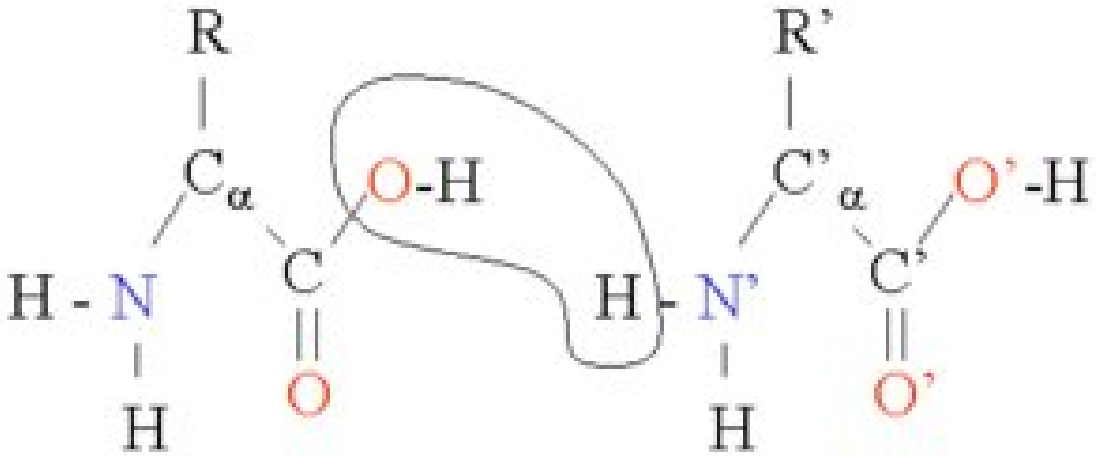

**Two amino acids**

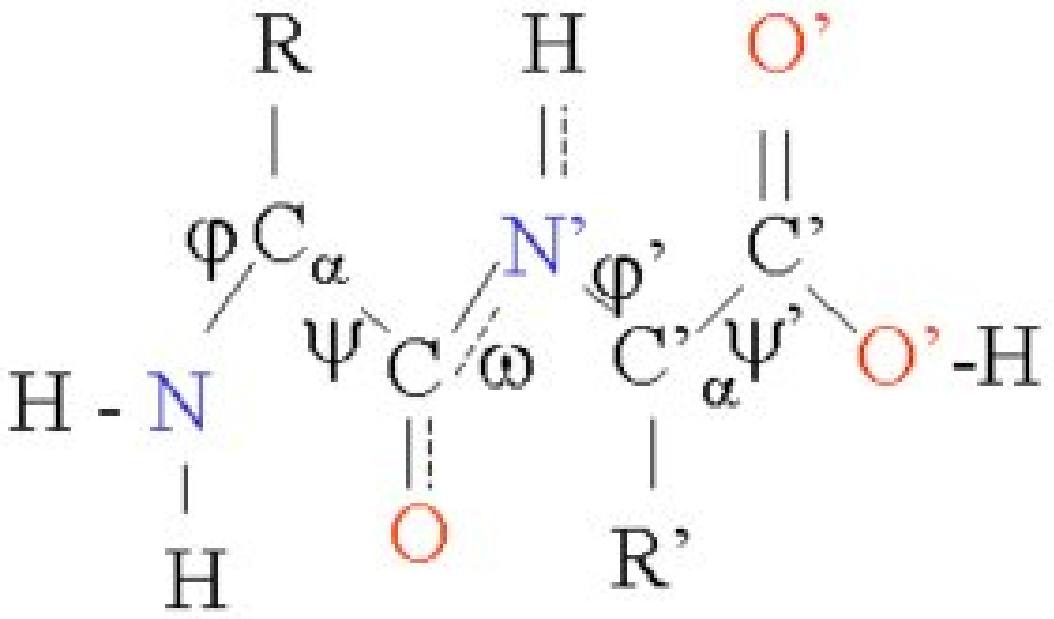

**Bond angles for ψ and ω**

Two amino acids can be combined in a condensation reaction. By repeating this reaction, long chains of residues (amino acids in a peptide bond) can be generated. The rigid peptide dihedral angle, ω (the bond between C1 and N) is always close to 180 degrees. The dihedral angles phi φ (the bond between N and Cα) and psi ψ (the bond between Cα and C1) can have a certain range of possible values. These angles are the degrees of freedom of a protein, they control the protein's three dimensional structure. A few important bond lengths are given in the table below.

| Peptide bond | Average length | Single bond | Average length | Hydrogen bond | Average (±30) |
|---|---|---|---|---|---|
| C□ C | 153 pm | C - C | 154 pm | O-H --- O-H | 280 pm |
| C – N | 133 pm | C - N | 148 pm | N-H --- O=C | 290 pm |
| N - C□ | 146 pm | C - O | 143 pm | O-H --- O=C | 280 pm |

**Secondary structure**

highly regular sub-structures (alpha helix and strands of beta pleated sheet), which are locally defined, meaning that there can be many different secondary motifs present in one single protein molecule.

Secondary structure is the spatial arrangement of the polypeptide ignoring the confirmation of the side chains. So secondary structure is local ordered structure brought about by hydrogen bonding mainly within the peptide backbone. The most common secondary structure elements in proteins are the alpha helix and the beta sheet.

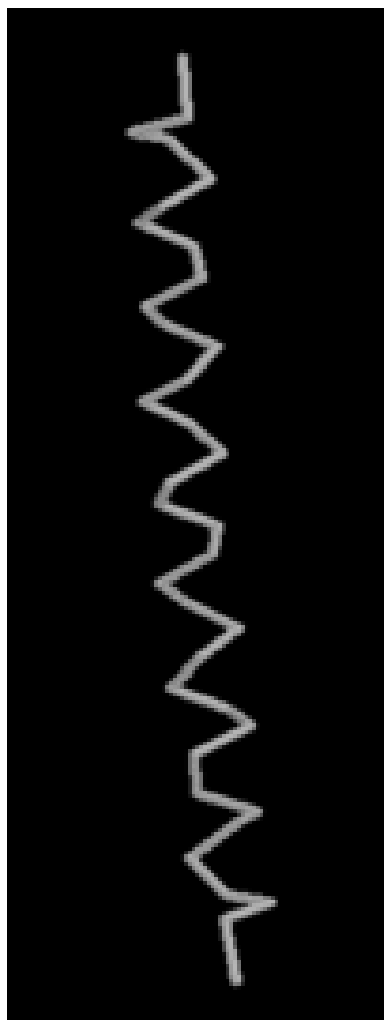

**left**

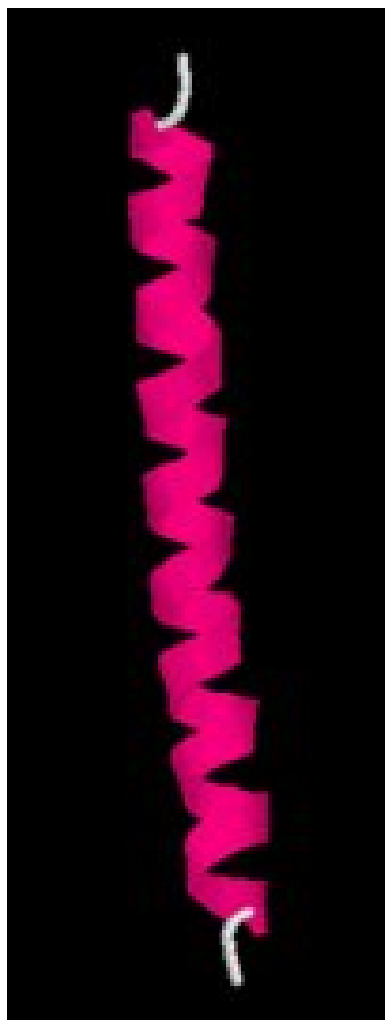

**right**

**Left: Ca atom trace. Right: Secondary structure cartoon ("ribbon")**

By building models of peptides using known information about bond lengths and angles, the first elements of secondary structure, the alpha helix and the beta sheet. Each of these two secondary structure elements have a regular geometry, meaning they are constrained to specific values of the dihedral angles ψ and φ.

**Tertiary structure**

three-dimensional structure of a single protein molecule; a spatial arrangement of the secondary structures. It also describes the completely folded and compacted polypeptide chain.

Tertiary structure is the three dimensional structure of the entire polypeptide it can be said the global folding of a single polypeptide chain. Hydrophobic effect is one of the major driving forces in determining the tertiary structure of globular proteins. The polypeptide chain folds in such a way that the side chains of the nonpolar amino acids are hidden within the structure and the side chains of the polar residues are exposed on the outer surface. Hydrogen bonding plays an important role in stabilizing tertiary structure. In some proteins disulfide bonds between cysteine residues are involve in stabilizing the tertiary structure.

The tertiary structure encompasses all the noncovalent interactions that are not considered secondary structure.

**Quaternary structure**

complex of several protein molecules or polypeptide chains, usually called protein subunits in this context, which function as part of the larger assembly or protein complex.

Quaternary structure refers to the three dimensional structure of proteins that are composed of two or more polypeptide chains. The spatial arrangement of these subunits is the quaternary structure of the protein. The forces that hold sub-units together are the same weak bonds as those that stabilize the tertiary structure of proteins like van der Waals, hydrogen bonds and salt bridges. The contact region between sub-units resembles the interior of a protein. The sub-units may be identical or non-identical.

The quaternary structure is not required for all proteins to be functional; many proteins may have only secondary or tertiary structure.

## 2.3 PROTEIN STRUCTURE DETERMINATION

Around 90% of the protein structures available in the Protein Data Bank have been determined by X-ray crystallography. This method allows one to measure the 3D density distribution of electrons in the protein (in the crystallized state) and thereby infer the 3D coordinates of all the atoms to be determined to a certain resolution. Roughly 9% of the known protein structures have been obtained by Nuclear Magnetic Resonance techniques, which can also be used to determine secondary structure. Note that aspects of the secondary structure as whole can be determined via other biochemical techniques such as circular dichroism or dual polarisation interferometry. Secondary structure can also be predicted with a high degree of accuracy (see next section). Cryo-electron microscopy has recently become a means of determining protein structures to high resolution (less than 5 angstroms or 0.5 nanometer) and is anticipated to increase in power as a tool for high resolution work in the next decade. This technique is still a valuable resource for researchers working with very large protein complexes such as virus coat proteins and amyloid fibers.

A rough guide to the resolution of protein structures

| Resolution (Å) | Meaning |
|---|---|
| >4.0 | Individual coordinates meaningless |
| 3.0 - 4.0 | Fold possibly correct, but errors are very likely. Many sidechains placed with wrong rotamer. |
| 2.5 - 3.0 | Fold likely correct except that some surface loops might be mismodelled. Several long, thin sidechains (lys, glu, gln, etc) and small sidechains (ser, val, thr, etc) likely to have wrong rotamers. |
| 2.0 - 2.5 | As 2.5 - 3.0, but number of sidechains in wrong rotamer is considerably less. Many small errors can normally be detected. Fold normally correct and number of errors in surface loops is small. Water molecules and small ligands become visible. |
| 1.5 - 2.0 | Few residues have wrong rotamer. Many small errors can normally be detected. Folds are extremely rarely incorrect, even in surface loops. |
| 0.5 - 1.5 | In general, structures have almost no errors at this resolution. Rotamer libraries and geometry studies are made from these structures. |

## 2.3 THE DNA, RNA & PROTEINS

DNA or otherwise called deoxyribonucleic acid is the building block of the life. It contains the information the cell requires to synthesize protein and to replicate itself, to be short it is the storage repository for the information that is required for any cell to function. Watson-Crick has discovered the current-structure of DNA in 1953.The famous double-helix structure of DNA has its own significance. There are basically four nucleotide bases, which make up the DNA. Adenine (A), Guanine (G), Thymine (T) and Cytosine(C). A DNA sequence looks something like this "ATTGCTGAAGGTGCGG". DNA is measured according to the number of base pairs it consists of, usually in kBp or mBp(Kilo/Mega base pairs). Each base has its complementary base, which means in the double helical structure of DNA, A will have T as its complimentary and similarly G will have C. DNA molecules are incredibly long. If all the DNA bases of the human genome were typed as A, C, T and G, the 3 billion letters would fill 4,000 books of 500 pages each! The DNA is broken down into bits and is tightly wound into coils, which are called chromosomes; human beings have 23 pairs of chromosomes. These chromosomes are further broken down into smaller pieces of code called Genes. The 23 pairs of chromosomes consist of about 70,000 genes and every gene has its own function. As I have mentioned earlier, DNA is made up of four nucleotide bases, finding out the arrangement of the bases is called DNA sequencing, there are various methods for sequencing a DNA, it is usually carried out by a machine or by running the DNA sample over a gel otherwise called gel electrophoresis. A typical sequence would look like this "ATTTGCTGACCTG".

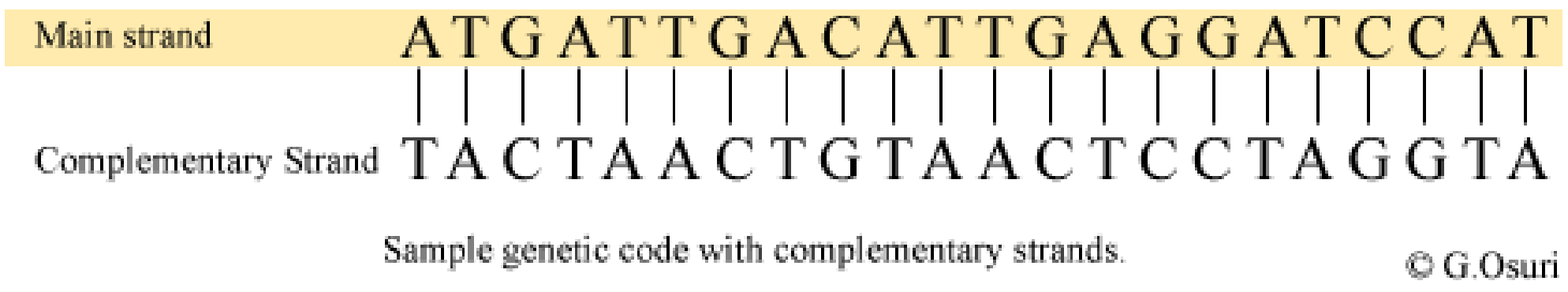

**Sample genetic code with complementary strands.**

Determining the gene's functionality and position of the gene in the chromosome is called gene mapping. Recent developments show that scientists are mapping every gene in the human body. They named their project Human Genome Project (HGP), which involves careful study of all the 70,000 genes in human body. Whew! That's some thing unimaginable. When there is a change in the genetic code it is called mutation.

The significance of a DNA is very high. The gene's sequence is like language that instructs cell to manufacture a particular protein. An intermediate language, encoded in the sequence of Ribonucleic Acid (RNA), translates a gene's message into a protein's amino acid sequence. It is the protein that determines the trait. This is called central dogma of life.

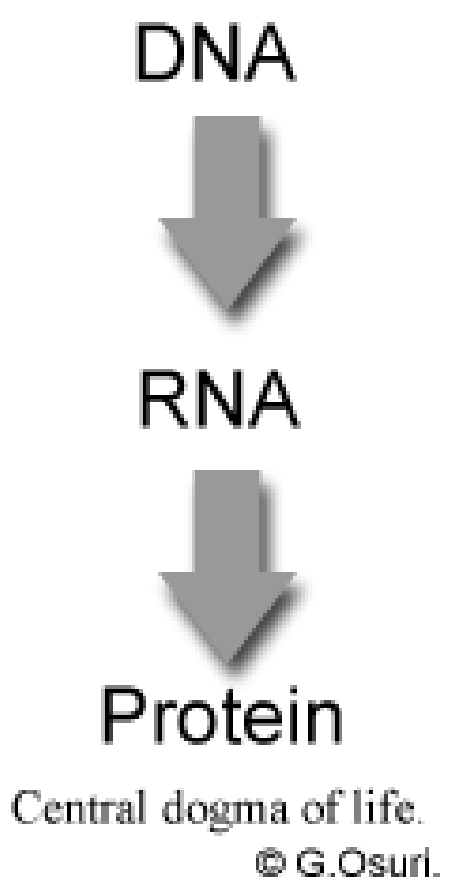

**Central dogma of life.**

*Notes: Genes are DNA sequences instruct cells to produce particular proteins, which in turn determine traits. Chromosomes are strings of genes. Mutations are changes in gene's DNA sequence.*

RNA is somewhat similar to DNA; they both are nucleic acids of nitrogen-containing bases joined by sugar-phosphate backbone. How ever structural and functional differences distinguish RNA from DNA. Structurally, RNA is a single-stranded where as DNA is double stranded. DNA has Thymine, where as RNA has Uracil. RNA nucleotides include sugar ribose, rather than the Deoxyribose that is part of DNA. Functionally, DNA maintains the protein-encoding information, whereas RNA uses the information to enable the cell to synthesize the particular protein.

| RNA | DNA |
|---|---|
| Single-Stranded | Double-Stranded |
| Has Uracil as a base | Has Thymine as a base |
| Ribose as the sugar | Deoxyribose as the sugar |
| Uses protein-encoding information | Maintains protein-encoding information |

**Differences between DNA and RNA**

*Notes: DNA stores the genetic information, where as RNA uses the information to help the cell produces the protein.*

## 2.3.1 TRANSCRIPTION, TRANSLATION

As I have mentioned earlier DNA forms the backbone for our lives, you should understand how it actually functions and helps the cell to function.

Transcription is a process of making an RNA strand from a DNA template, and the RNA molecule that is made is called transcript. In the synthesis of proteins, there are actually three types of RNA that participate and play different roles:

a. Messenger RNA(mRNA), which carries the genetic information from DNA and is used as a template for protein synthesis.
b. Ribosomal RNA(rRNA), which is a major constituent of the cellular particles called ribosomes on which protein synthesis actually takes place.
c. A set of transfer RNA(tRNA) molecules, each of which incorporates a particular amino acid subunit into the growing protein when it recognizes a specific group of three adjacent bases in the mRNA.

DNA maintains genetic information in the nucleus. RNA takes that information into the cytoplasm, where the cell uses it to construct specific proteins, RNA synthesis is transcription; protein synthesis is translation.
RNA differs from DNA in that it is single stranded, contains Uracil instead of Thymine and ribose instead of deoxyribose, and has different functions. The central dogma depicts RNA as a messenger between gene and protein, but does not adequately describe RNA's other function.
Transcription is highly controlled and complex. In Prokaryotes, genes are expressed as required, and in multicellular organisms, specialized cell types express subsets of gene. Transcription factors recognize sequences near a gene and bind sequentially, creating a binding transcription. Transcription proceeds as RNAP inserts complementary RNA bases opposite the coding strand of DNA. Antisense RNA blocks gene expression.
Messenger RNA transmits information in a gene to cellular structures that build proteins. Each three mRNA bases in a row forms a codon that specifies a particular amino acid. Ribosomal RNA and proteins form ribosomes, which physically support the other participants in protein synthesis and help catalyze formation of bonds betweens amino acids.

In eukaryotes, RNA is often altered before it is active. Messenger RNA gains a cap of modified nucleotides and a poly A tail. Introns are transcribed and cut out, and exons are reattached by ribozymes. RNA editing introduced bases changes that alter the protein product in different cell types.
The genetic code is triplet, non-overlapping, continuous, universal, and degenerate. As translation begins, mRNA, tRNA with bound amino acids, ribosomes, energy molecules and protein factos assemble. The mRNA leader sequence binds to rRNA in the small subunit of a ribosome, and the first codon attracts a tRNA bearing methionine. Next, as the chain elongates, the large ribosomal subunit attaches and the appropriate anticodon parts of tRNA molecules form peptide bonds, a polypeptide grows. At a stop codon, protein synthesis ceases. Protein folding begins as translation proceeds, with enzymes and chaperone proteins assisting the amino acid chain in assuming its final functional form. Translation is efficient and economical, as RNA, ribosomes, enzymes, and key proteins are recycled.

A much detailed explanation could be found in the book 'Molecular Biology' by David Freifelder. I suggest you read this book if you interested in detailed knowledge about transcription and translation.

**Amino acids**

Nitrogen-containing amino acids are essential for life and they are the building blocks of proteins. Amino acids, as ancient and ubiquitous molecules, have been co-opted by evolution for a variety of purposes in living systems. The importance in reading this section is limited to those who wants to visualize the structures.

**Amino acid Structure:**

There are basically 20 standard amino acids having different structures in their side chains(R groups) . The common amino acids are known as a-amino acids because they have a primary amino group(-NH2) and a carboxylic acid group(-COOH) as substitutes of the a carbon atoms. Proline is an exception because it has a secondary amino group (-NH-), for uniformity it is also treated as alpha-amino acid.

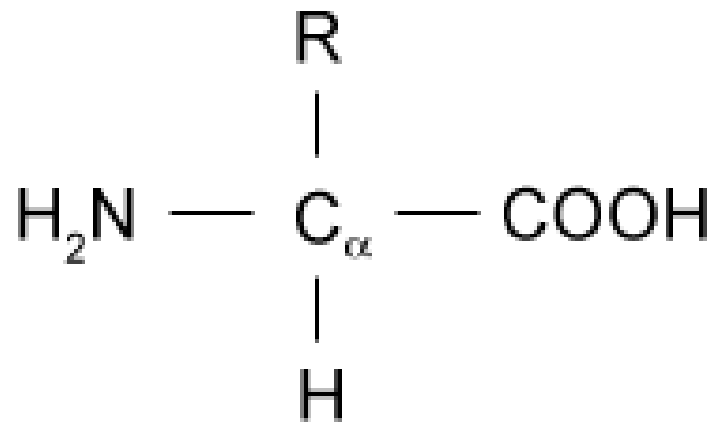

**General structure of a- amino acid.**

**General properties:**

The amino and carboxylic acid groups of amino acids readily ionize. At a pH(~7.4), the amino groups are protonated and the carboxyl acid groups are in their conjugate base(carboxylate) form, this shows that an amino acid that can act as an Acid and also a base. Amino acids can bear charged groups of opposite polarity, hence they are know as zwitterions or dipolar ions. The ionic property of the side chains influences the physical and chemical property of free amino acids and amino acids in proteins.

**Peptide bonds:**

Elimination of water (condensation) can polarize amino acids to form long chains. The resulting CO-NH linkage, an amide linkage, is known as peptide bond. Polymers composed of two, three, a few(3 - 10), and many amino acid units are known, respectively, as dipeptides, tripeptides, oligopeptides, and polypeptides, commonly they are called "peptides".

**Classification:**

There are basically three major classifications for amino acids (1) those with nonpolar R group, (2) those with uncharged polar R groups, and (3) those with charged polar R group.
Non-polar amino acid side chains have a variety of shapes and sizes, there are basically nine acids under this classification. Glycine has the smallest possible side chain, an H atom. Alanine, valine, leucine, and isoleucine have aliphatic hydrocarbon side chains ranging in size from a methyl group for alanine to isomeric butyl groups for leucine and isoleucine. Methionine has a thiol ether side chain that resembles an n-butyl group in many of its physical properties(C and A have nearly equal electronegativities, and S is about the size of a methylene group). Proline has a cyclic pyrrolidine side group. Phenylalanine(with its phenyl moiety) and typtophan(with its indole group) conaint aromatic side groups, which are characterized by bulk as well as nonpolarity.

Uncharged Polar side chains have Hydroxyl, Amide, or Thiol Groups
There are six amino acids under this. Serine and threonine have hydroxylic R side chains of different sizes. Asparagine and glutamine have amide-bearing side chains of different sizes. Tyrosine has a phenolic group and is aromatic. Cystein is very unique among all 20 amino acid because it has a thiol group that form a disulfide bond with other Cystein through oxidation.
Charged Polar side chains, they are positively or negatively charged. Five amino acids contribute to this type. The side chains are positively charged; they are lysine, which has a butylammonium side chain, arginine, which has a guanidine group, and histidine, which bears an imidazolium moiety.

**Nomenclature:**

The three letter and single letter abbreviations are given, most of them are taken from the first three letters and pronounced and written for eg. The symbol Glx in dicates Glu or Gln, care should be taken while writing these notations. The one-letter notations are used while comparing sequences of several similar proteins. With this we end our discussion about amino acids and you don't have to worry about them until visualizing of a protein comes into picture. During translation the genetic code in the form of RNA is translated into a protein sequence.

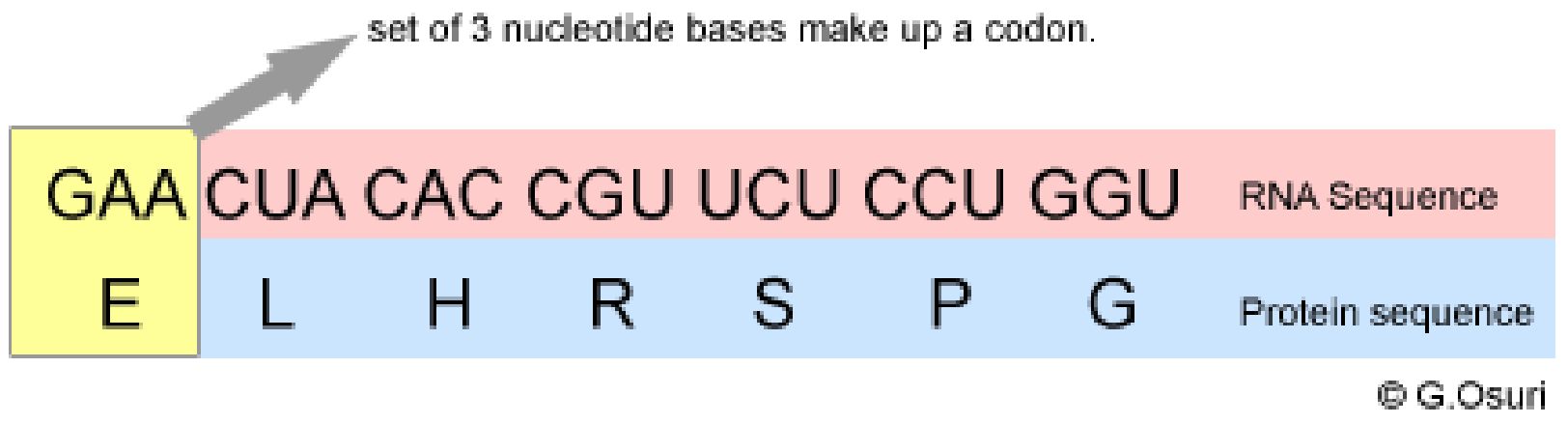

**synthesis of protein sequence from a mRNA**

## 2.3.2 CENTRAL DOGMA OF PROTEIN SYNTHESIS

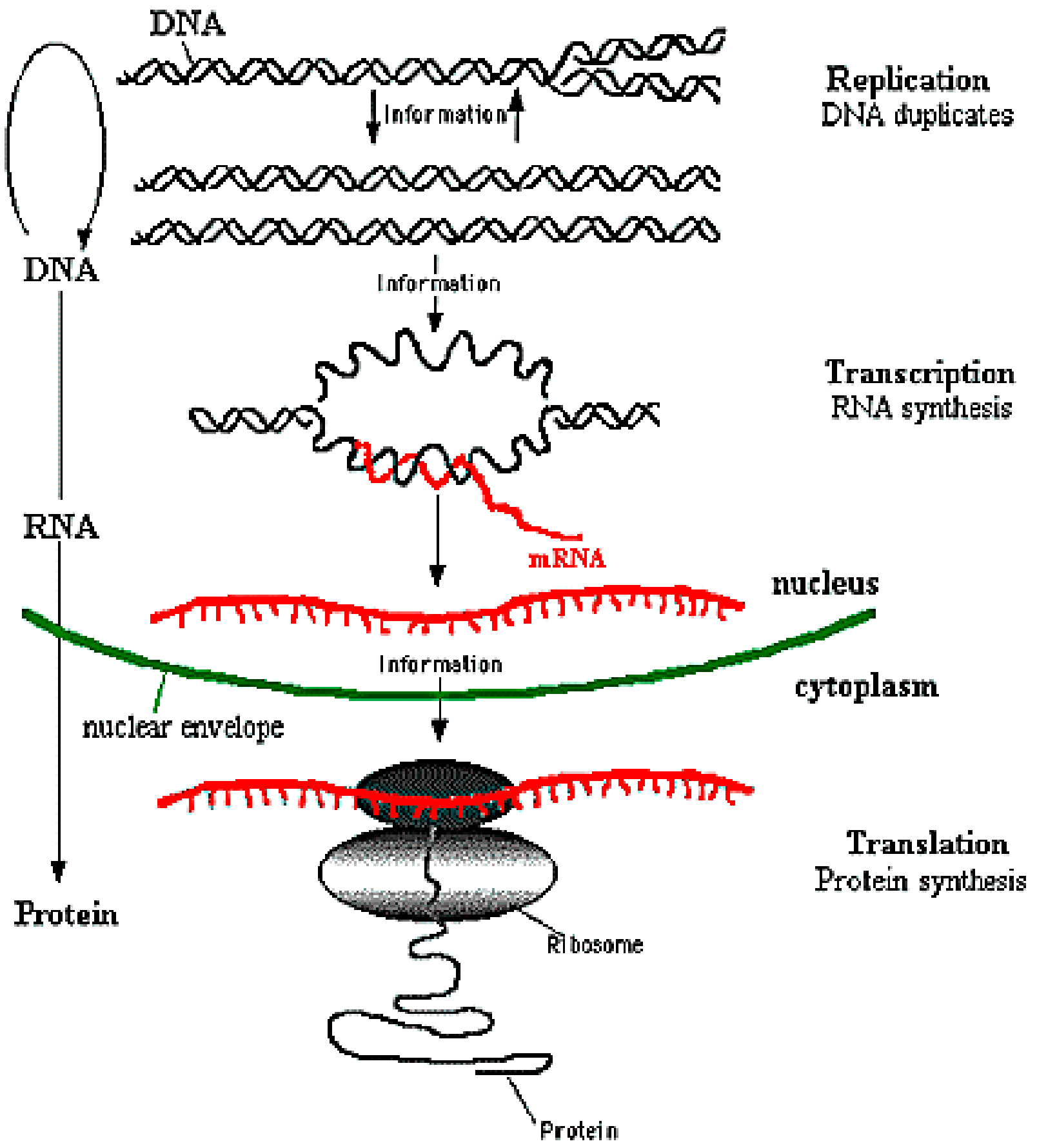

**The Central Dogma of Molecular Biology**

The DNA contain the information for protein in the form of nucleotide sequences. This information is copied by mRNA. The mRNA synthesis from DNA is called **transcription**.
The nucleotide sequence of mRNA determines the sequence of amino acids in a protein. The assembling of amino acids as per the sequence of nucleotides of mRNA is called **translation**. This is the most accepted fact and hence the central dogma of protein synthesis.

Proteins constitute the major part by dry weight of an actively growing cell. They are widely distributed in living matter. All enzymes are proteins. Proteins are built up from about 20 amino acids which constitute the basic building blocks. In proteins the amino acids are linked up by peptide bonds to from long chains called polypeptides.

The sequence of amino acids has a bearing on the properties of a protein, and is characteristic for a particular protein. The basic mechanism of protein synthesis is that DNA makes RNA, which in turn makes protein. The central dogma of protein synthesis is expressed as follows:

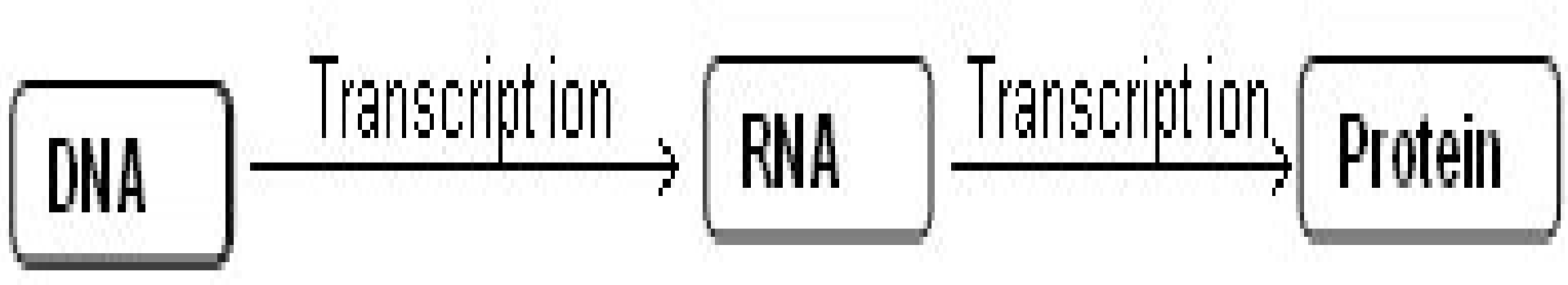

Central dogma of protein synthesis.

## 2.4 PROTEIN STRUCTURE DATABASES

In biology, a protein structure database is a database that is modeled around the various experimentally determined protein structures. The aim of most protein structure databases is to organize and annotate the protein structures, providing the biological community access to the experimental data in a useful way.

**Bioinformatics**

Bioinformatics is the application of computer science to the field of molecular biology. The term bioinformatics was coined by Paulien Hogeweg in 1979 for the study of informatic processes in biotic systems. Its primary use since at least the late 1980s has been in genomics and genetics, particularly in those areas of genomics involving large-scale DNA sequencing. Common activities in bioinformatics include mapping and analyzing DNA and protein sequences, aligning different DNA and protein sequences to compare them and creating and viewing 3-D models of protein structures.
The primary goal of bioinformatics is to increase our understanding of biological processes.

Important sub-disciplines within bioinformatics and computational biology include:

- the development and implementation of tools that enable efficient access to, and use and management of, various types of information.
- the development of new algorithms (mathematical formulas) and statistics with which to assess relationships among members of large data sets, such as methods to locate a gene within a sequence, predict protein structure and/or function, and cluster protein sequences into families of related sequences.

**Prediction of protein structure**

Protein structure prediction is another important application of bioinformatics. The amino acid sequence of a protein, the so-called primary structure, can be easily determined from the sequence on the gene that codes for it.

### Genotype-phenotype distinction

The genotype-phenotype distinction is drawn in genetics. "Genotype" is an organism's full hereditary information, even if not expressed. "Phenotype" is an organism's actual observed properties, such as morphology, development, or behavior. This distinction is fundamental in the study of inheritance of traits and their evolution.

The genotype represents its exact genetic makeup — the particular set of genes it possesses. Two organisms whose genes differ at even one locus (position in their genome) are said to have different genotypes.

The mapping of a set of genotypes to a set of phenotypes is sometimes referred to as the genotype-phenotype map.

Even two organisms with identical genotypes normally differ in their phenotypes. One experiences this in everyday life with monozygous (i.e. identical) twins. Identical twins share the same genotype, since their genomes are identical; but they never have the same phenotype, although their phenotypes may be very similar.

### Phenotype

A phenotype is any observable characteristic or trait of an organism: such as its morphology, development, biochemical or physiological properties, or behavior. Phenotypes result from the expression of an organism's genes as well as the influence of environmental factors and possible interactions between the two.

The interaction between genotype and phenotype has often been conceptualized by the following relationship:

**genotype + environment → phenotype**

A slightly more nuanced version of the relationships is:

**genotype + environment + random-variation → phenotype**

**Phenotype** This is the "outward, physical manifestation" of the organism. These are the physical parts, the sum of the atoms, molecules, macromolecules, cells, structures, metabolism, energy utilization, tissues, organs, reflexes and behaviors; anything that is part of the observable structure, function or behavior of a living organism.

**Genotype** This is the "internally coded, inheritable information" carried by all living organisms. This stored information is used as a "blueprint" or set of instructions for building and maintaining a living creature.

The Genetic Code is stored on one of the two strands of a DNA molecules as a linear, non-overlapping sequence of the nitrogenous bases Adenine (A), Guanine (G), Cytosine (C) and Thymine (T). These are the "alphabet" of letters that are used to write the "code words".

The genetic code consists of a sequence of three letter "words" (sometimes called 'triplets', sometimes called 'codons'), written one after another along the length of the DNA strand.

Each code word is a unique combination of three letters (like the ones shown above) that will eventually be interpreted as a single amino acid in a polypeptide chain. There are 64 code words possible from an 'alphabet' of four letters.

One of these code words, the 'start signal' begins all the sequences that code for amino acid chains. Three of these code words act as 'stop signals' that indicate that the message is over. All the other sequences code for specific amino acids.

Some amino acids are only coded for by a single 'word', while some others are coded for by up to four 'words'. The genetic code is redundant.

## The Protein Data Bank

The Protein Data Bank (PDB) was established in 1971 as the central archive of all experimentally determined protein structure data. Today the PDB is maintained by an international consortia collectively known as the Worldwide

Protein Data Bank (wwPDB). The mission of the wwPDB is to maintain a single archive of macromolecular structural data that is freely and publicly available to the global community.

## Protein Structure Databases

Because the PDB releases data into the public domain, the data has been used in various other protein structure databases.

Examples of protein structure databases include (in alphabetical order);

**Database of Macromolecular Movements**

describes the motions that occur in proteins and other macromolecules, particularly using movies

**JenaLib**

the Jena Library of Biological Macromolecules is aimed at a better dissemination of information on three-dimensional biopolymer structures with an emphasis on visualization and analysis.

**MODBASE**

a database of three-dimensional protein models calculated by comparative modeling

**MSD**

the Macromolecular Structure Database (MSD) the European project for the collection, management and distribution of data about macromolecular structures, derived in part from the PDB

**OCA**

a browser-database for protein structure/function - The OCA integrates information
from KEGG, OMIM, PDBselect, Pfam, PubMed, SCOP, SwissProt and others.

**OPM**

provides spatial positions of protein three-dimensional structures with respect to the lipid bilayer.

**PDB Lite**

derived from OCA, PDB Lite was provided to make it as easy as possible to find and view a macromolecule within the PDB

**PDBsum**

provides an overview macromolecular structures in the PDB, giving schematic diagrams of the molecules in each structure and of the interactions between them

**PDBTM**

the Protein Data Bank of Transmembrane Proteins a selection of the PDB.

**PDBWiki**

a community annotated knowledge base of biological molecular structures

**Proteopedia**

the collaborative, 3D encyclopedia of proteins and other molecules. A wiki that contains a page for every entry in the PDB (>50,000 pages), with a Jmol view that highlights functional sites and ligands. Offers an easy-to-use scene-authoring tool so you don't have to learn Jmol script language to create customized molecular scenes. Custom scenes are easily attached to "green links" in descriptive text that display those scenes in Jmol.

**SCOP**

the Structural Classification of Proteins a detailed and comprehensive description of the structural and evolutionary relationships between all proteins whose structure is known.

**SWISS-MODEL Repository**

a database of annotated protein models calculated by homology modeling

**TOPSAN**

the Open Protein Structure Annotation Network a wiki designed to collect, share and distribute information about protein three-dimensional structures.

\

# METHODOLOGY USED & SPECIFICATIONS

# 3.1 HIDDEN MARKOV MODEL

## 3.1.1 INTRODUCTION TO HIDDEN MARKOV MODEL

A hidden Markov Model (HMM) is a statistical model where the system being modeled is assumed to be a Markov process with unknown parameters, and the challenge is to determine the hidden parameters from the observed parameters. The extracted model parameters can then be used to perform further analysis, for example for prediction of patterns of a system.
In a regular Markov model, the state is directly visible to the observer, and therefore the state transition probabilities are the only parameters of concern. Whereas in hidden Markov model, the state is not directly visible, but variables influenced by the state are visible. Each state has a probability distribution over the possible output tokens. Therefore the sequence of tokens generated by an HMM gives some information about the sequence of states.

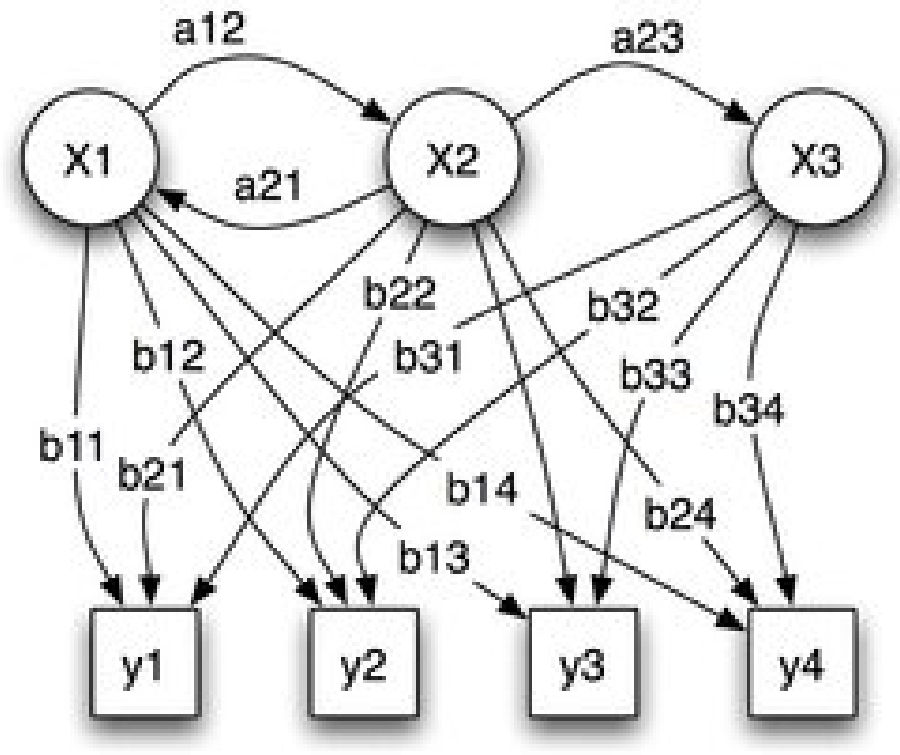

Probabilistic parameters of a hidden Markov model (example)
x — states
y — possible observations
a — state transition probabilities
b — output probabilities

**A hidden Markov model (HMM)** is a statistical model in which the system being modeled is assumed to be a Markov process with unobserved state. An HMM can be considered as the simplest dynamic Bayesian network.

In a regular Markov model, the state is directly visible to the observer, and therefore the state transition probabilities are the only parameters. In a hidden Markov model, the state is not directly visible, but output dependent on the state is visible. Each state has a probability distribution over the possible output tokens. Therefore the sequence of tokens generated by a HMM gives some information about the sequence of states.

Hidden Markov models are especially known for their application in temporal pattern recognition such as speech, handwriting, gesture recognition, part-of-speech tagging, musical score following, partial discharges and bioinformatics.

**Architecture of a hidden Markov model**

The diagram below shows the general architecture of an instantiated HMM. Each oval shape represents a random variable that can adopt any of a number of values. The random variable x(t) is the hidden state at time t (with the model from the above diagram, x(t) ∈ { x1, x2, x3 }). The random variable y(t) is the observation at time t (y(t) ∈ { y1, y2, y3, y4 }). The arrows in the diagram (often called a trellis diagram) denote conditional dependencies.

From the diagram, it is clear that the conditional probability distribution of the hidden variable x(t) at time t, given the values of the hidden variable x at all times, depends only on the value of the hidden variable x(t − 1): the values at time t − 2 and before have no influence. This is called the Markov property. Similarly, the value of the observed variable y(t) only depends on the value of the hidden variable x(t) (both at time t).

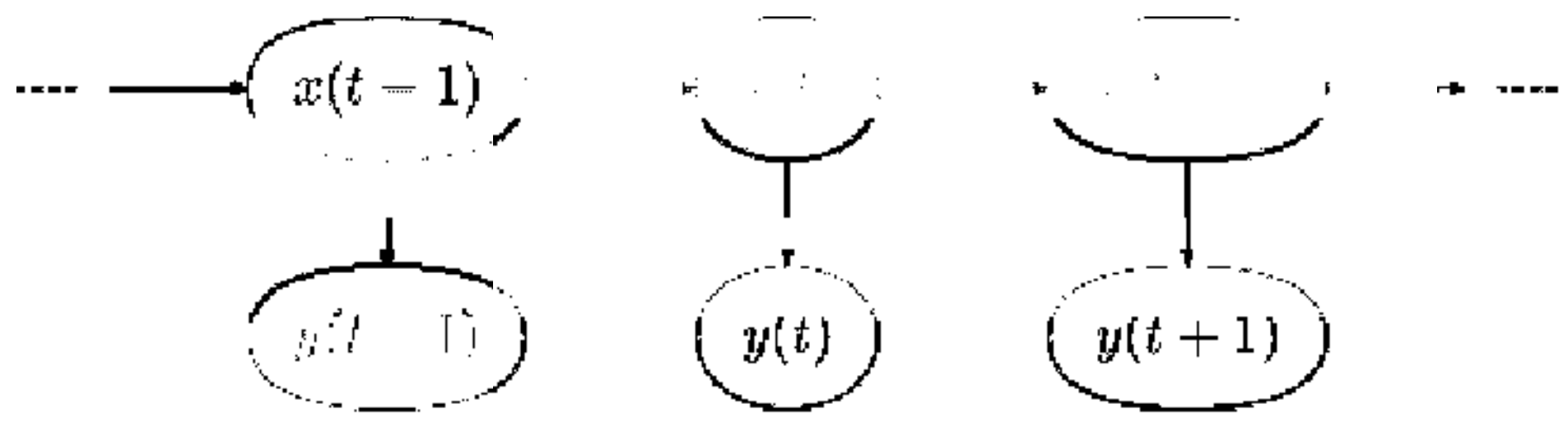

**Probability of an observed sequence**

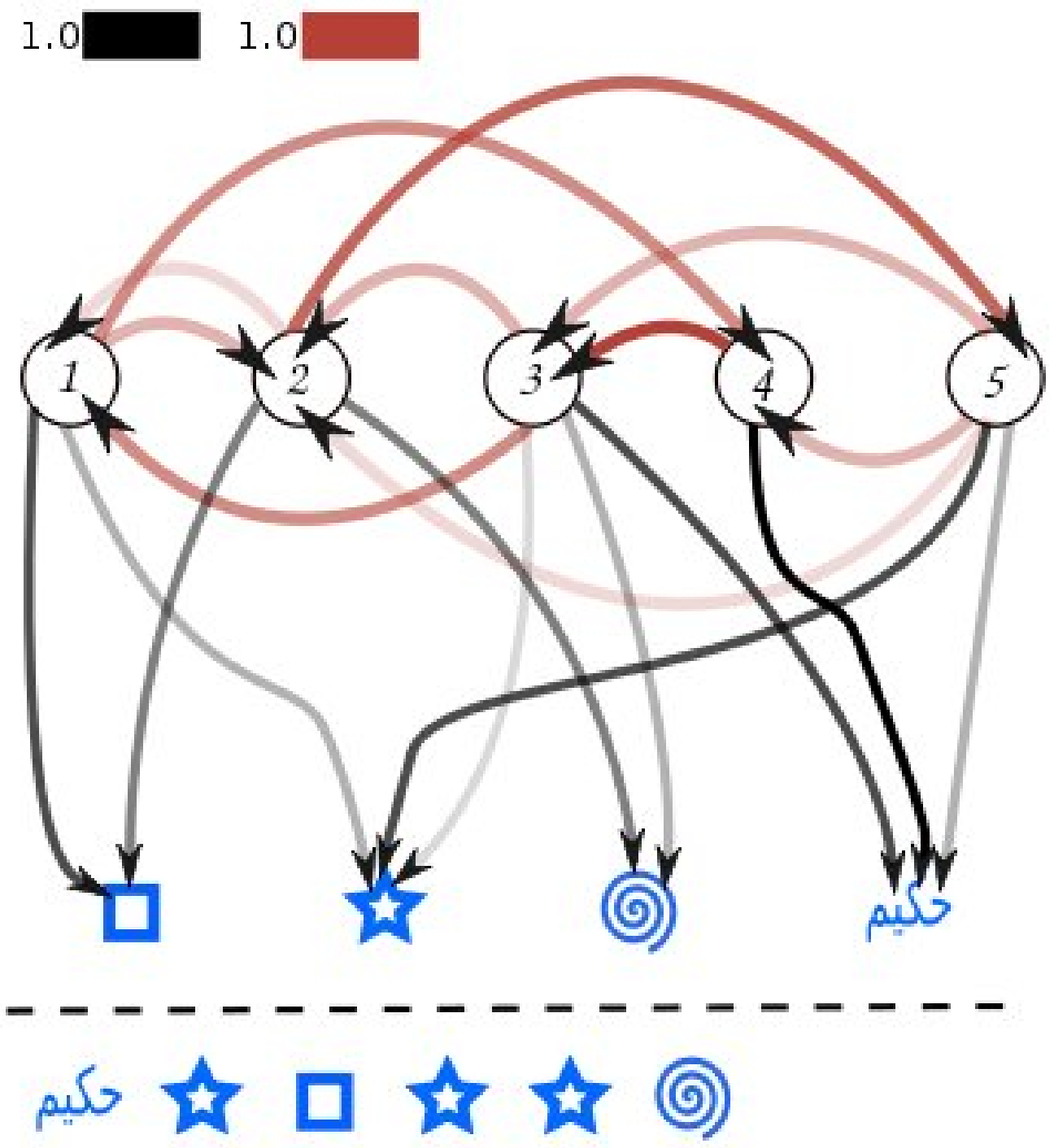

The observation sequence above can be produced by the following state sequences.
5 3 2 5 3 2
4 3 2 5 3 2
3 1 2 5 3 2

Transition and observation probabilities are indicated by the line opacity.

The probability of observing a sequence

$$Y = y(0), y(1), \dots, y(L-1)$$

of length L is given by

$$P(Y) = \sum_{X} P(Y \mid X) P(X),$$

where the sum runs over all possible hidden-node sequences

$$X = x(0), x(1), \dots, x(L-1).$$

Brute-force calculation of P(Y) is intractable for most real-life problems, as the number of possible hidden node sequences is typically extremely high and scales exponentially with the length of the sequence. The calculation can however be sped up enormously using the Forward-backward algorithm.

## 3.1.2 ELEMENTS OF AN HMM

An HMM is characterized by the following:

1. N, the number of states in the model. Although the states are hidden, for many practical applications there is often some physical significance attached to the states or to sets of states of the model. Generally the states are interconnected in such a way that any state can be reached from any other state. We denote the individual states as S = {$S_l$, $S_2$,. . . $S_N$}, and the state at time t as qt.

2. M, the number of distinct observation symbols per state, i.e., the discrete alphabet size. The observation symbols correspond to the physical output of the system being modeled. We denote the individual symbols as V = {$v_l$, $v_2$, . . .,$v_M$}

3. The state transition probability distribution A = { $a_{ij}$ } where

   $$a_{ij} = p[q_{t+l} = Sj|qt = Si], 1 <= i, j <= N.$$

   For the special case where any state can reach any other state in a single step, we have $a_{ij}$ > 0 for all i, j. For other types of HMMs, we would have $a_{ij}$ = 0 for one or more (i,j) pairs.

4. The observation symbol probability distribution in state j, B = { $b_{j(k)}$}, where

   $$b_{j(k)} = P[v_k \text{ at } t|q_t = S_j], 1 <= j <= N$$

5. The initial state distribution π = { $\pi_i$ } where

   $$\pi_i = p[q_l = S_i], 1 <=i <= N.$$

Given appropriate values of N, M, A, B, and π , the HMM can be used as a generator to give an observation sequence O=$O_1$,$O_2$,....$O_T$.
(where each observation $O_T$ is one of the symbols from V, and T is the number of observations in the sequence) as follows:

1. Choose an initial state q1 = Si according to the initial state distribution π.
2. Set t = 1.
3. Choose $O_T$ = $v_k$ according to the symbol probability distribution in state $S_i$, i.e., $b_i$ (k).

4. Transit to a new state $q_{t+1} = S_j$ according to the state transition probability distribution for state $S_i$, i.e., $a_{ij}$
5. Set t = t + 1; return to step 3) if t < T; otherwise terminate the procedure.

The above procedure can be used as both a generator of observations, and as a model for how a given observation sequence was generated by an appropriate HMM.
It can be seen that a complete specification of an HMM requires specification of two model parameters (N and M), specification of observation symbols, and the specification of the three probability measures A, B, and π. For convenience, we use the compact notation

$$\boldsymbol{\lambda = (A, B, \pi)}$$

to indicate the complete parameter set of the model.

**Notational conventions**

T = length of the sequence of observations (training set)
N = number of states (we either know or guess this number)
M = number of possible observations (from the training set)
Omega_X = {q_1,...q_N} (finite set of possible states)
Omega_O = {v_1,...,v_M} (finite set of possible observations)
X_t random variable denoting the state at time t (state variable)
O_t random variable denoting the observation at time t (output variable)
sigma = o_1,...,o_T (sequence of actual observations)

**Distributional parameters**

A = {a_ij} s.t. a_ij = Pr(X_t+1 = q_j |X_t = q_i) (transition probabilities)
B = {b_i} s.t. b_i(k) = Pr(O_t = v_k | X_t = q_i t) (observation probabilities)
pi = {pi_i} s.t. pi_i = Pr(X_0 = q_i) (initial state distribution)

**Definitions**

A **hidden Markov model** (HMM) is a five-tuple (Omega_X,Omega_O,A,B,pi). Let lambda = {A,B,pi} denote the parameters for a given HMM with fixed Omega_X and Omega_O.

**Problems**

1. Find Pr(sigma|lambda): the probability of the observations given the model.
2. Find the most likely state trajectory given the model and observations.
3. Adjust lambda = {A,B,pi} to maximize Pr(sigma|lambda).

**Motivation**

A discrete-time, discrete-space dynamical system governed by a Markov chain emits a sequence of observable outputs: one output (observation) for each state in a trajectory of such states. From the observable sequence of outputs, infer the most likely dynamical system. The result is a model for the underlying process. Alternatively, given a sequence of outputs, infer the most likely sequence of states. We might also use the model to predict the next observation or more generally a continuation of the sequence of observations.

Hidden Markov models are used in speech recognition. Suppose that we have a set W of words and a separate training set for each word. Build an HMM for each word using the associated training set. Let lambda_w denote the HMM parameters associated with the word w. When presented with a sequence of observations sigma, choose the word with the most likely model, i.e.,

w* = arg max_{w in W} Pr(sigma|lambda_w)

## Forward-Backward Algorithm

### Preliminaries

Define the alpha values as follows,
    alpha_t(i) = Pr(O_1=o_1,...,O_t=o_t, X_t = q_i | lambda)
Note that
    alpha_T(i) = Pr(O_1=o_1,...,O_T=o_T, X_T = q_i | lambda)
        = Pr(sigma, X_T = q_i | lambda)
The alpha values enable us to solve Problem 1 since, marginalizing, we obtain
    Pr(sigma|lambda) = sum_i=1^N Pr(o_1,...,o_T, X_T = q_i | lambda)
        = sum_i=1^N alpha_T(i)

Define the beta values as follows,

    beta_t(i) = Pr(O_t+1=o_t+1,...,O_T=o_T | X_t = q_i, lambda)
We will need the beta values later in the Baum-Welch algorithm.

### Algorithmic Details

1. Compute the forward (alpha) values:
    a. alpha_1(i) = pi_i b_i(o_1)
    b. alpha_t+1(j) = [sum_i=1^N alpha_t(i) a_ij] b_j(o_t+1)
2. Computing the backward (beta) values:
    a. beta_T(i) = 1
    b. beta_t(i) = sum_j=1^N a_ij b_j(o_t+1) beta_t+1(j)

## Viterbi Algorithm

### Intuition

Compute the most likely trajectory starting with the empty output sequence; use this result to compute the most likely trajectory with an output sequence of length one; recurse until you have the most likely trajectory for the entire sequence of outputs.

### Algorithmic Details

1. Initialization:

For 1 <= i <= N,

a. delta_1(i) = pi b_i(o_1)

b. Phi_1(i) = 0

2. Recursion:

For 2 <= t <= T, 1 <= j <= N,

a. delta_t(j) = max_i [delta_t-1(i)a_ij]b_j(o_t)

b. Phi_t(j) = argmax_i [delta_t-1(i)a_ij]

3. Termination:

a. p* = max_i [delta_T(i)]

b. i*_T = argmax_i [delta_T(i)]

4. Reconstruction:

For t = t-1,t-2,...,1,

i*_t = Phi_t+1(i*_t+1)

The resulting trajectory, i*_1,...,i*_T, solves Problem 2.

### 3.1.3 THE BASIC PROBLEMS FOR HMM

There are three basic problems of interest that must be solved for the model to be useful in real world applications. These problems are the following:

**Problem 1:** Given the observation sequence O=O1,O2,....OT and a model λ = (A, B, π), how to efficiently compute P(O|λ), the probability of the observation sequence, given the model.

**Problem 2:** Given the observation sequence O=O1,O2,....OT and the model λ how to choose a corresponding state sequence Q = q1 q2 . . . qT which is optimal in some meaningful sense.

**Problem 3:** How to adjust the model parameters λ = (A, B, π), to maximize P (O|λ).

Problem 1 is the evaluation problem, namely given a model and a sequence of observations, how do we compute the probability that the observed sequence was produced by the model. We can also view the problem as one of scoring how well a given model matches a given observation sequence. The latter viewpoint is extremely useful. For example, if we consider the case in which we are trying to choose among several competing models, the solution to Problem 1 allows us to choose the model which best matches the observations.

Problem 2 is the one in which we attempt to uncover the hidden part of the model, i.e., to find the “correct” state sequence.

Problem 3 is the one in which we attempt to optimize the model parameters so as to best describe how a given observation sequence comes about. The observation sequence used to adjust the model parameters is called a training sequence since it is used to “train” the HMM. The training problem is the crucial one for most applications of HMMs, since it allows us to optimally adapt model parameters to observed training data-i.e., to create best models for real phenomena.

## 3.1.4 SOLUTIONS TO THE BASIC PROBLEMS OF HMM

### A. Solution to Problem 1.

The Forward-Backward Procedure is used to solve this problem.
Consider the forward variable $\alpha_{t(i)}$ defined as

$$\alpha_{t(i)} = P(O1, O2 \ldots O_t, q_t = S_i|\lambda)$$

i.e., the probability of the partial observation sequence, O1 ,O2 . . . Ot,(until time t) and state Si at time t ,given the model λ. We can solve for $\alpha_{t(i)}$ inductively, as follows:

1) Initialization:

$$\alpha_{1(i)} = \pi_i . b_{i(O1)}, \; 1 <= i <= N.$$

2) Induction:

$$\alpha_{(t+1)(j)} = [\sum_{i=1}^{N} \alpha_t(i). a_{ij}] . b_{j(Ot+1)}, \; 1 <= t <= T-1$$
$$1 <= j <= N.$$

3) Termination:

$$P(Q|\lambda) = [\sum_{i=1}^{N} \alpha_{T(i)}]$$

Step(I) initializes the forward probabilities as the joint probability of state Si and initial observation O1. The induction step, which is the heart of the forward calculation, is illustrated in fallowing figure.

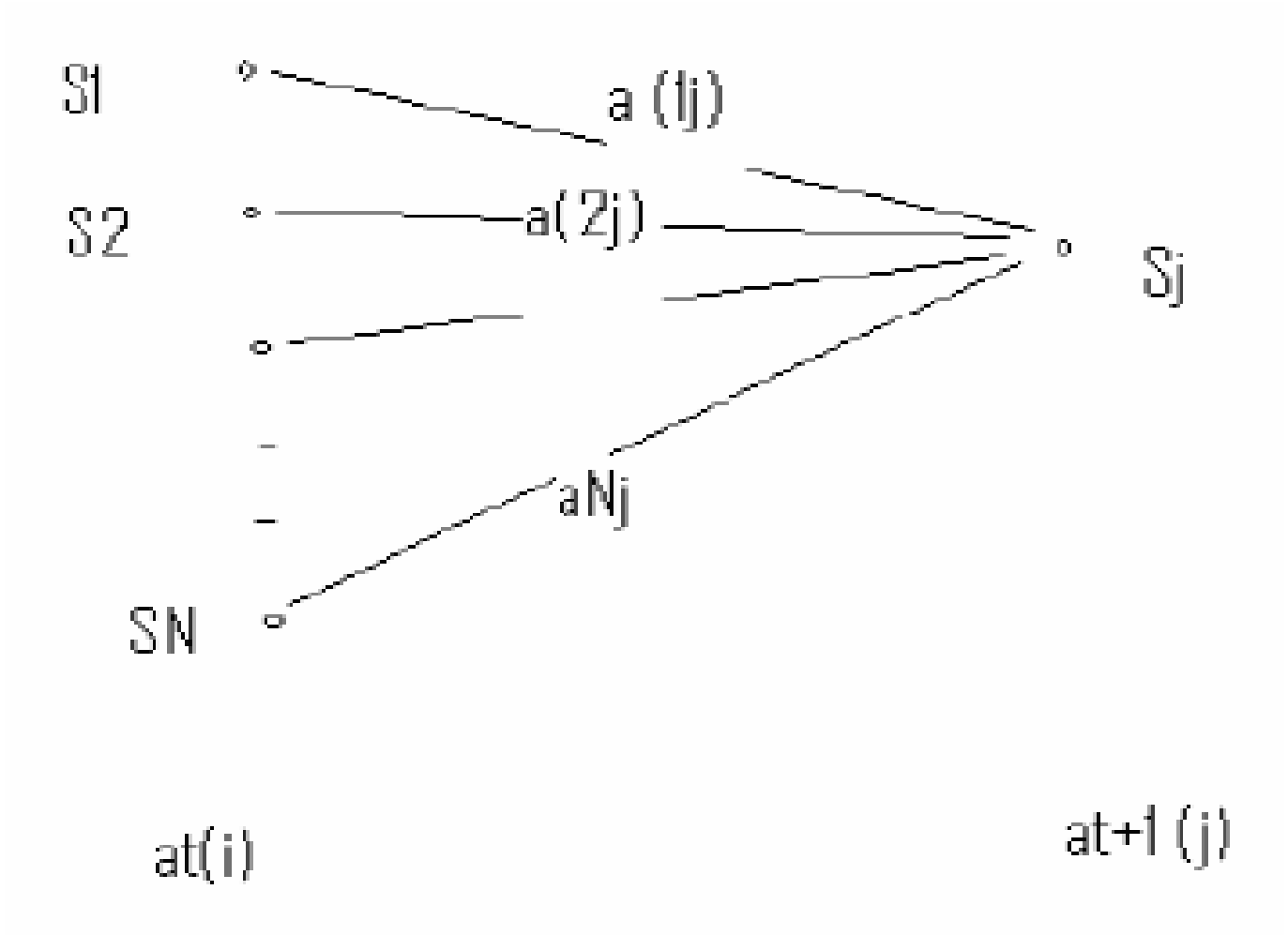

**Figure (1)** Illustration of the sequence of operations required for the computation of the forward variable $\alpha_{t+1(j)}$ [8].

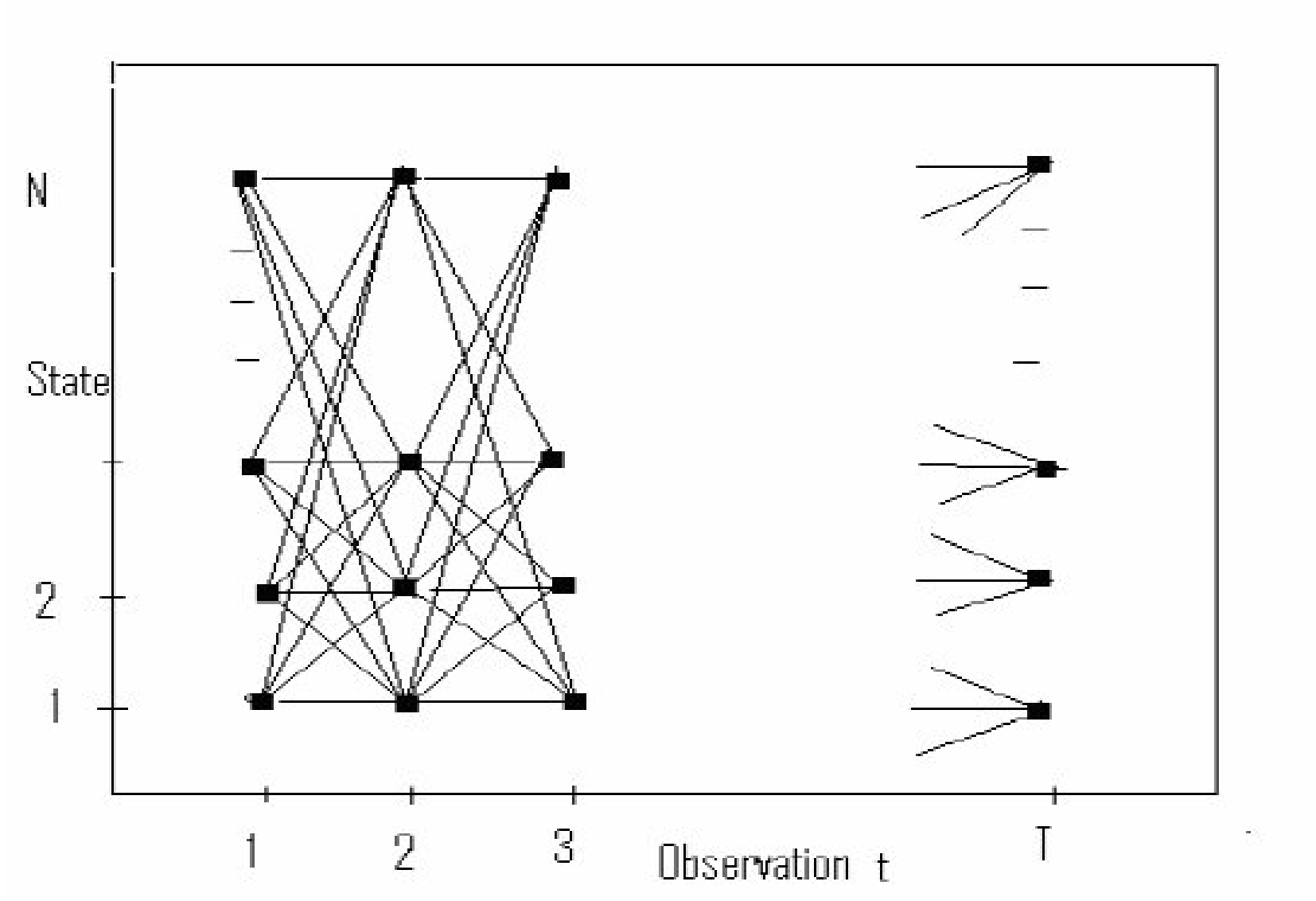

**Figure (2)** Implementation of the computation of $\alpha_{t\ (i)}$ in terms of a lattice of observations t, and states i [8].

Above figure shows how state $S_j$ can be reached at time t + 1 from the N possible states, $S_i$, 1 <= i <= N, at time t. Since $\alpha_{t(i)}$ is the probability of the joint event that $O_1$ , $O_2$ . . . $O_t$, are observed, and the state at time t is Si, the product $\alpha_{t(i)} . a_{ij}$ is then the probability of the joint event that are observed, and state Sj is reached at time t + 1 via state Si at time t. Summing this product over all the N possible states Si, 1 <= i <= N at time t results in the probability of slat time t + 1 with all the accompanying previous partial observations. Once this is done and Sj is known, it is easy to see that $\alpha_{t+l(j)}$ is obtained by accounting for observation Ot+1 in state j, i.e., by multiplying the summed quantity by the probability $b_j(O_{t+l})$. The computation of $\alpha_{(t+1)(j)}$ is performed for all states j, 1 <= j <= N, for a given t; the computation is then iterated for t = 1,2, . . . , T - 1. Finally, step 3) gives the desired calculation of P(O|λ) as the sum of the terminal forward variables $\alpha_{T(i)}$. This is the case since, by definition,

$$\alpha_{T(i)} = P(O_1 , O_2 . . . O_T, q_T = S_i \mid \lambda)$$

and hence P(O|λ) is just the sum of the $\alpha_{T(i)}$'s.

In a similar manner, we can consider a backward variable βt(i) defined as

$$\beta_{t(i)} = P(O_{t+1}, O_{t+2} . . . O_T \mid q_t = S_i, \lambda )$$

i.e., the probability of the partial observation sequence from t + 1 to the end, given state Si at time t and the model λ, again we can solve for βt(i) inductively, as follows:

1) Initialization:

$$\beta_{T(i)} = 1, 1 <= i <= N.$$

2) Induction:

$$\beta_{t(i)} = \sum_{i=1}^{N} a_{ij} . b_j(O_{t+1})\ \beta_{t+1(j)} , t = T-1, T-2, ..., 1$$

$$1 <= j <= N.$$

The initialization step (1) arbitrarily defines $\beta_{T(i)}$ to be 1 for all i. Step(2), shows that in order to have been in state $S_i$ at time t, and to account for the observation sequence from time t + 1 on, we have to consider all possible states $S_j$ at time t + 1, accounting for the transition from $S_i$ to $S_j$ (the $a_{ij}$ term), as well as the observation in state j (the $b_{j(Ot+1)}$ term), and then account for the remaining partial observation sequence from state j (the $\beta_{t+1(j)}$ term).

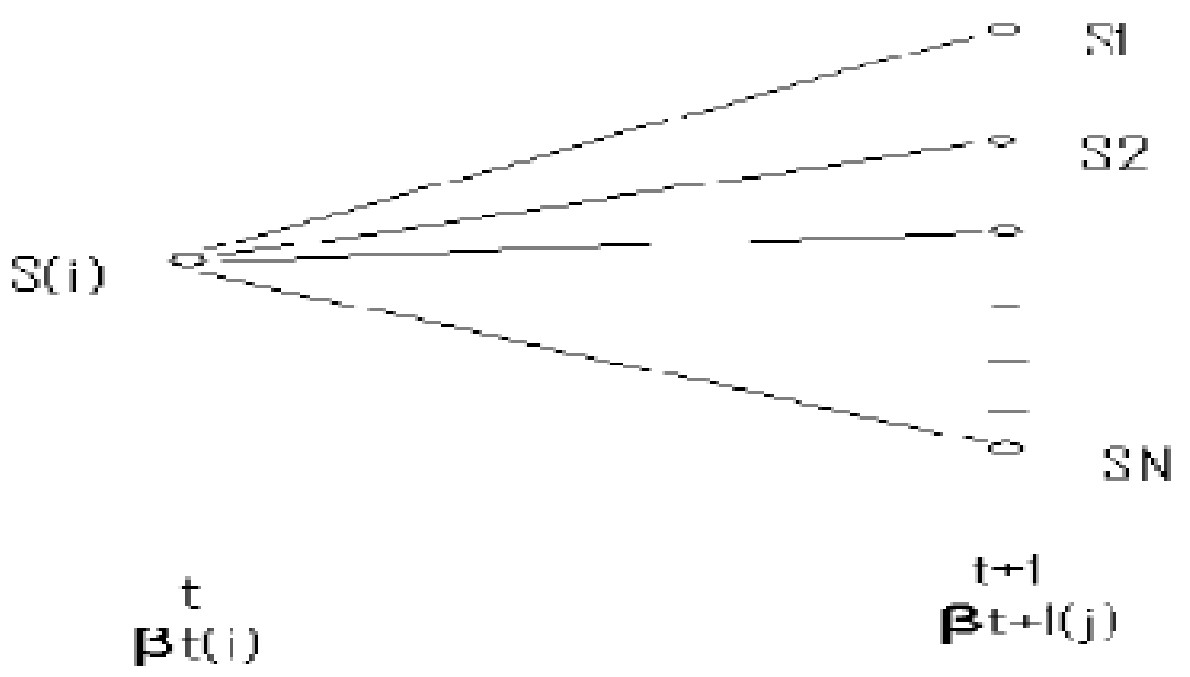

Figure (3) Illustration of the sequence of operations required for the computation of the backward variable $\beta_t(i)$ [8].

**B. Solution to Problem 2.**

There are several possible ways of finding the "optimal" state sequence associated with the given observation sequence. The difficulty lies with the definition of the optimal state sequence; i.e., there are several possible optimality criteria. For example, one possible optimality criterion is to choose the states qt which are individually most likely. This optimality criterion maximizes the expected number of correct individual states. To implement this solution, we define the variable

$\gamma_{t(i)} = P(qt = S_i | O, \lambda)$

i.e., the probability of being in state Si at time t, given the observation sequence O, and the model λ . Above equation can be expressed simply in terms of the forward-backward variables, i.e.,

$$\gamma_{t(i)} = \alpha_{t(i)} . \beta_{t(i)} / P(O|\lambda) = \alpha_{t(i)} . \beta_{t(i)} / \sum_{i=1}^{N} \alpha_{t(i)} . \beta_{t(i)} .$$

since $\alpha_{t(i)}$ accounts for the partial observation sequence $O_1, O_2, \ldots O_t$, and state Si at t, while $\beta_{t(i)}$ accounts for the remainder of the observation sequence $O_{t+1}, O_{t+2} \ldots O_T$, given state Si at t. The normalization factor

$$P(O|\lambda) = \sum_{i=1}^{N} \alpha_{t(i)} . \beta_{t(i)}$$

makes $\gamma_{t(i)}$ a probability measure so that

$$\sum_{i=1}^{N} \gamma_{t(i)} = 1$$

Using $\gamma_{t(i)}$ ,we can solve for the individually most likely state qt at time t, as

$$q_t = \underset{1 <= i <= N}{\text{argmax}} [\gamma_{t(i)}] , 1 <= t <= T$$

Although above equation maximizes the expected number of correct states (by choosing the most likely state for each t), there could be some problems with the resulting state sequence. For example, when the HMM has state transitions which have zero probability ($a_{ij}$ = 0 for some i and j ), the"optima1" state sequence may, in fact, not even be a valid state sequence. This is due to the fact that the above solution simply determines the most likely state at every instant, without regard to the probability of occurrence of sequences of states. tools to explore the relationship of protein sequence and structure.

**C. Solution to problem 3.**

The third and the most difficult problem of HMMs are to determine a method to adjust the model parameters (A, B, π) to maximize the probability of the observation sequence given the model. There is no known way to analytically solve for the model which maximizes the probability of the observation sequence. In fact, given any finite observation sequence as training data, there is no optimal way of estimating the model parameters. We can, however, choose λ = (A, B, π) such that P (O|λ) is locally maximized using an iterative procedure such as the Baum-Welch method (or equivalently the EM (expectation-modification) method), or using gradient techniques.

### 3.1.5 FORWARD AND BACKWARD PROBABILITIES FOR A PROFILE HMM

As with general HMMs, the main problem is to assign meaningful values to the transition and emission probabilities to a profile HMM. It is possible to use the Baum-Welch algorithm for training the model probabilities, but it first has to be shown how to compute the forward and backward probabilities needed for the algorithm.

Given a string $X = (x_1, \dots, x_m)$ we define:

- The forward probabilities:

$$f_j^M(i) = P(x_1, \dots, x_i \text{ ending at } M_j)$$

$$f_j^I(i) = P(x_1, \dots, x_i \text{ ending at } I_j)$$

$$f_j^D(i) = P(x_1, \dots, x_i \text{ ending at } D_j)$$

- 

- The backward probabilities:

$$b_j^M(i) = P(x_{i+1}, \dots, x_m \text{ beginning from } M_j)$$

$$b_j^I(i) = P(x_{i+1}, \dots, x_m \text{ beginning from } I_j)$$

$$b_j^D(i) = P(x_{i+1}, \dots, x_m \text{ beginning from } D_j)$$

- 

**Computing the Forward Probabilities**:

1.
    Initialization:

$f_{begin}(0) = 1$

2.
    Recursion:

$$\begin{aligned} f_j^M(i) = e_{M_j}(x_i) \cdot [&f_{j-1}^M(i-1) \cdot a_{M_{j-1},M_j} + \\ &f_{j-1}^I(i-1) \cdot a_{I_{j-1},M_j} + \\ &f_{j-1}^D(i-1) \cdot a_{D_{j-1},M_j}] \end{aligned}$$

$$f_j^I(i) = e_{I_j}(x_i) \cdot [f_j^M(i-1) \cdot a_{M_j,I_j} + f_j^I(i-1) \cdot a_{I_j,I_j} + f_j^D(i-1) \cdot a_{D_j,I_j}]$$

$$f_j^D(i) = f_{j-1}^M(i) \cdot a_{M_{j-1},D_j} + f_{j-1}^I(i) \cdot a_{I_{j-1},D_j} + f_{j-1}^D(i) \cdot a_{D_{j-1},D_j}$$

**Computing the Backward Probabilities**:

1.
Initialization:

$b^M{}_L(m) = a_{ML,end}$

$b^I{}_L(m) = a_{IL,end}$

$b^D{}_L(m) = a_{DL,end}$

2.
Recursion:

$$b_j^M(i) = b_{j+1}^M(i+1) \cdot a_{M_j,M_{j+1}} \cdot e_{M_{j+1}}(x_{i+1}) + b_j^I(i+1) \cdot a_{M_j,I_j} \cdot e_{I_j}(x_{i+1}) + b_{j+1}^D(i) \cdot a_{M_j,D_{j+1}}$$

$$b_j^I(i) = b_{j+1}^M(i+1) \cdot a_{I_j,M_{j+1}} \cdot e_{M_{j+1}}(x_{i+1}) + b_j^I(i+1) \cdot a_{I_j,I_j} \cdot e_{I_j}(x_{i+1}) + b_{j+1}^D(i) \cdot a_{I_j,D_{j+1}}$$

$$b_j^D(i) = b_{j+1}^M(i+1) \cdot a_{D_j,M_{j+1}} \cdot e_{M_{j+1}}(x_{i+1}) + b_j^I(i+1) \cdot a_{D_j,I_j} \cdot e_{I_j}(x_{i+1}) + b_{j+1}^D(i) \cdot a_{D_j,D_{j+1}}$$

The forward and backward variables can then be combined to re-estimate emission and transition probability parameters as follows:

**Baum-Welch re-estimation equations fo profile HMMs**:

1. Expected emission counts from sequence X:

$$E_{M_k}(a) = \frac{1}{P(X)} \sum_{i|x_i=a} f_k^M(i) b_k^M(i)$$

$$E_{I_k}(a) = \frac{1}{P(X)} \sum_{i|x_i=a} f_k^I(i) b_k^I(i)$$

2. Expected transition counts from sequence x:

$$A_{X_k M_{k+1}} = \frac{1}{P(X)} \sum_i f_k^X(i) a_{X_k M_{k+1}} e_{M_{k+1}}(x_{i+1}) b_{k+1}^M(i+1)$$

$$A_{X_k I_k} = \frac{1}{P(X)} \sum_i f_k^X(i) a_{X_k I_k} e_{I_k}(x_{i+1}) b_k^I(i+1)$$

$$A_{X_k D_{k+1}} = \frac{1}{P(X)} \sum_i f_k^X(i) a_{X_k D_{k+1}} b_{k+1}^D(i)$$

### 3.1.6 APPLICATIONS OF HMMs

HMM’s have applications in the many areas of computational biology like gene finding and prediction, Protein- Profile HMMs and Prediction of protein secondary structure.

**Gene finding and prediction**

The gene-prediction HMMs can be used to predict the structure of the gene. Our objective is to find the coding and non-coding regions of an unlabeled string of DNA nucleotides.
The motivation behind this is to assist in the annotation of genomic data produced by genome sequencing methods and to gain insight into the mechanisms involved in transcription, splicing and other processes
A string of DNA nucleotides containing a gene will have separate regions, introns (non-coding regions within a gene) and exons (coding regions). These regions are separated by functional sites start and stop codons and splice sites (acceptors and donors). In the process of transcription, only the exons are left to form the protein sequence.

Many problems in biological sequence analysis have a grammatical structure . HMMs are very useful in modeling grammar. The input to such a HMM is the genomic DNA sequence and the output, in the simplest case is a parse tree of exons and introns on the DNA sequence.

**Protein- Profile HMMs**

Protein structural similarities make it possible to create a statistical model of a protein family which is called a **profile**. The idea is, given a single amino acid target sequence of unknown structure, we want to infer the structure of the resulting protein. The profile HMM is built by analyzing the distribution of amino-acids in a training set of related proteins. This HMM in a natural way can model positional dependant gap penalties. Profile HMM’s can also be used for the following purpose:

**Scoring a sequence**
We can calculate the probability of a sequence given a profile by simply multiplying emission and transition probabilities along the path.

**Classifying sequences in a database**
Given a HMM for a protein family and some unknown sequences, we are trying to find a path through the model where the new sequence fits in or we are tying to ‘align’ the sequence to the model. Alignment to the model is an assignment of states to each residue in the sequence. There are many such alignments and the Viterbi’s algorithm is used to give the probability of the sequence for that alignment.

**Creating multiple sequence alignment**
HMMs can be used to automatically create a multiple alignment from a group of unaligned sequences. By taking a close look at the alignment, we can see the history of evolution. One great advantage of HMMs is that they can be estimated from sequences, without having to align the sequences first. The sequences used to estimate or train the model are called the training sequences, and any reserved sequences used to evaluate the model are called the test sequences.

The model estimation is done with the forward-backward algorithm. It is an iterative algorithm that maximizes the likelihood of the training sequences.

**HMM for protein secondary structure**

While applying HMM to predict protein secondary structures from protein sequences, researchers commonly take the amino acid sequence as observed and seek to recover the hidden, secondary protein structure from this sequence.

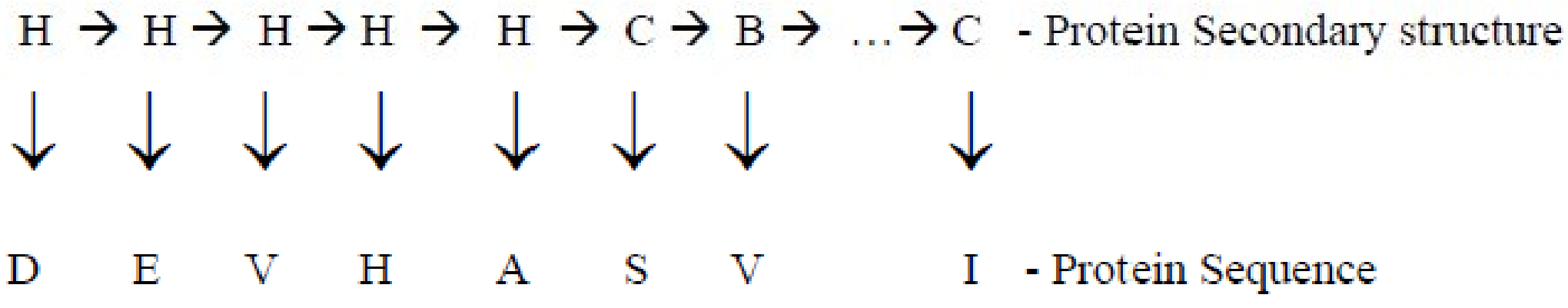

**Figure (4)** Emission of protein sequence (observed) from protein secondary structure (hidden) [10]

Figure (4) is translating the structure prediction problem into a hidden Markov model. The hidden process (upper line) is the succession of secondary structures: H for a α-helix, B for a β-strand, and C for a coil. The observed sequence, D, E, V, H, A, S, V, I, is the amino acid sequence (lower line). Horizontal arrows symbolize the hidden process's first-order dependence. Vertical arrows indicate the dependence between the observed sequence and hidden process. Successive amino acids are independent given the hidden process.
tools to explore the relationship of protein sequence and structure

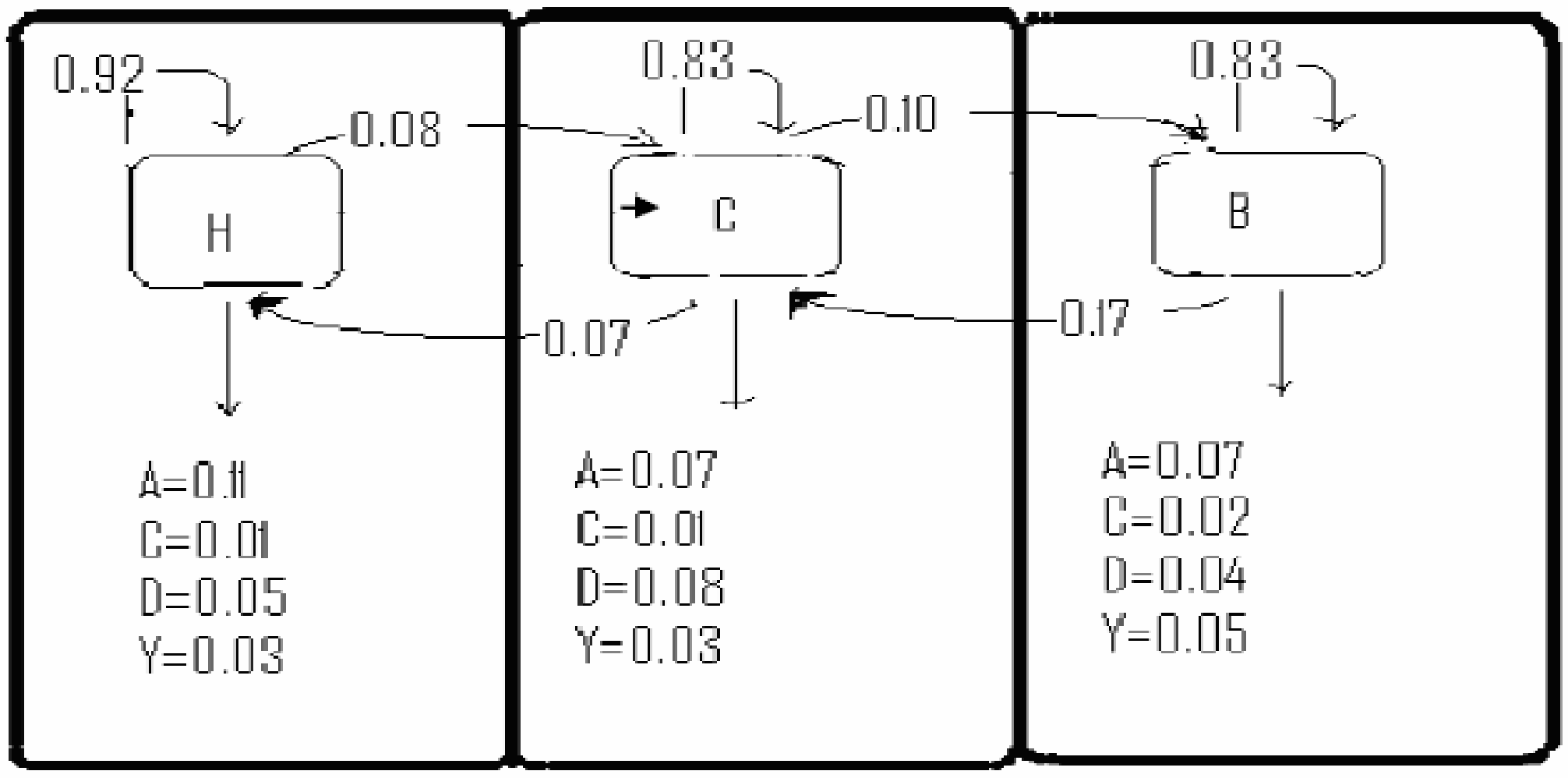

**Figure (5)** Three-state HMM for secondary-structure prediction [10]

Figure (5) shows a three-state HMM for prediction of secondary structures H (helix), C (coil) and B (beta sheet) from their emission relationship with the amino acids: A, C, D, and Y. The numbers next to each amino acids represent their emission probabilities. The numbers associated with state transitions are given with arrows.

### 3.1.7 ADVANTAGES OF HMMs

HMM's can accommodate variable-length sequence, because most biological data has variable-length properties, machine learning techniques which require a fixed-length input, such as neural networks or support vector machines, are less successful in biological sequence analysis.
HMM's allow position dependant gap penalties. HMM's treat insertions and deletions is a statistical manner that is dependent on position.

## 3.1.8 LIMITATIONS OF HMMs

HMM is a linear Model so they are unable to capture higher order correlations among amino-acids. In HMM we take markov Chain assumption of independent events. It means probabilities of states are supposed to be independent which is not true of biology.
In the training problem, we need to watch out for local maxima and so model may not converge to a truly optimal parameter set for a given training set. Secondly, Since the model is only as good as your training set, this may lead to over-fitting.

## 3.1.9 OPEN AREAS FOR RESEARCH IN HMMS IN BIOLOGY

1. Integration of structural information into profile HMMs: Despite the almost obvious application of using structural information on a member protein family when one exists to better the parameterization of the HMM, this has been extremely hard to achieve in practice.

2. Model architecture: The architectures of HMMs have largely been chosen to be the simplest architectures that can fit the observed data. Is this the best architecture to use? Can one use protein structure knowledge to make better architecture decisions, or, in limited regions, to learn the architecture directly from the data? Will these implied architectures have implications for our structural understanding?

3. Biological mechanism: In gene prediction, the HMM's may be getting close to replicating the same sort of accuracy as the biological machine (the HMM's have the additional task of finding the gene in the genomic DNA context, which is not handled by the biological machine that processes the RNA). What constraints does our statistical model place on the biological mechanism- in particular, can we consider a biological mechanism that could use the same information as the HMM.

# 3.2 ARTIFICIAL NEURAL NETWORK

## 3.2.1 INTRODUCTION TO ARTIFICIAL NEURAL NETWORK

The relationship between an amino acid sequence and the structure of the protein it forms is currently unknown. Researchers do not understand the folding process which causes this transformation and have termed this the protein folding problem. An artificial neural network (ANN) approach may be successful in solving the problem by implementing an ANN to predict protein structure from the amino acid sequence .

The protein folding problem articulates the challenge of determining how a protein is formed from its primary structure. Proteins are synthesized within the body. They begin as a chain of amino acids. This chain then kinks and folds as complex interactions take place between the molecules. Within seconds, the process is complete and a protein structure has formed. Only a single structure will form from a given amino acid sequence. Thus, the original sequence uniquely determines the resulting structure. How this happens remains unknown to researchers. The rules governing folding remain a mystery. Because we do not understand how protein folding occurs, we must determine a posteriori the protein’s structural feature. This information is often much harder to get than the amino acid sequence. The difficulties encountered in determining protein structure have stifled progress in protein engineering and gene therapy research. Our inability to predict protein structure is often viewed as the last major hurdle of molecular biology. With an algorithm for predicting protein folding, we would be in a much better position to harness the benefits of the Human Genome Project and hasten the arrival of intelligent drug design.
As the protein folding problem is such a pressing issue, it has received enormous attention from researchers. Most research has focused on deterministic methods. That is, an algorithm gives the rules for turning an input sequence into an output structure. The research lies in figuring out the rules to be followed.

One strategy for predicting structure is to look at overall trends and deduce from them a general set of rules. For example, we may observe a particular structure and note that it always comes from a class of amino acid sequences. Thus we may deduce that the class of sequences always gives rise to the observed structure. These observations are maintained in a database. When a new sequence is encountered, it is checked against the database for homologies. These homologies then give clues about the final structure.
The other strategy is to try combining various laws of chemistry and physics to predict the folding mechanism. Folding occurs as new chemical bonds are formed and others are broken. At the molecular level, quantum mechanical effects also become important. Thus, folding must occur according to the principles of physics and chemistry. The problem with this strategy is the risk of oversimplifying the folding process. There is still much about quantum effects that we do not understand. Perhaps this is why no algorithm has yet been discovered that successfully predicts structure. An alternative way to predict folding is to use an ANN. Such computer programming paradigms are esteemed for their ability to learn on their own the implicit relationship between input and output data. The network will figure out for itself what rules to follow.
Artificial neural networks draw their inspiration from biological neural networks 16, 17, 18, 19]. The basic computing unit of the brain is the neuron. The neuron gathers inputs through its many dendrites and sends them to the soma.

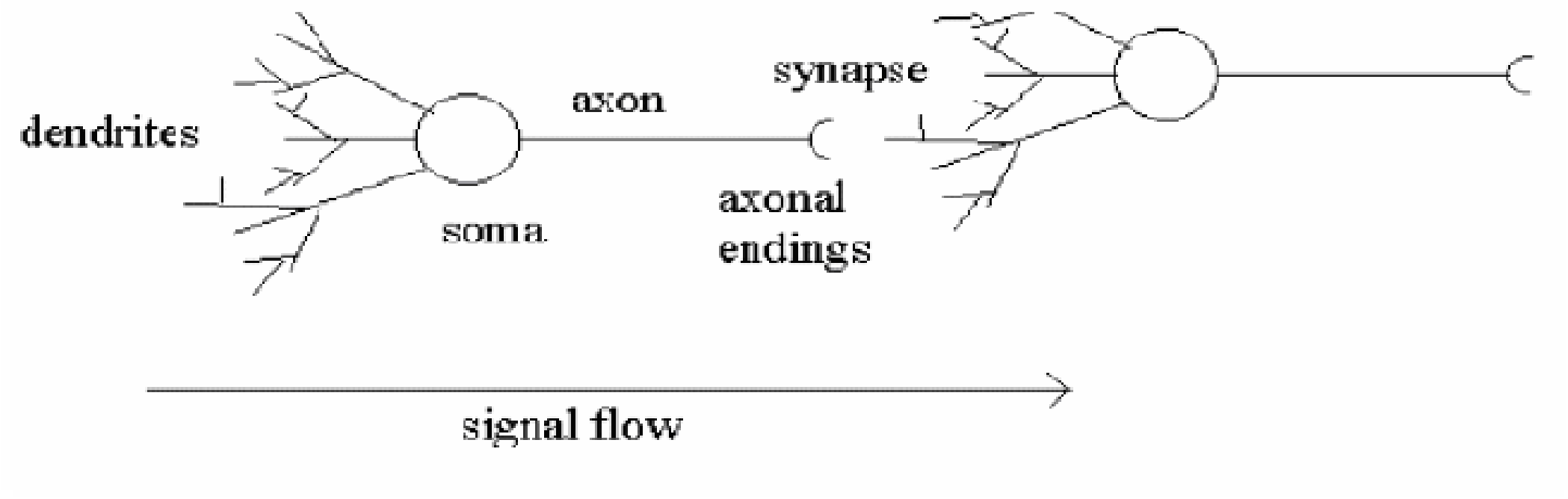

**Figure (6)** Schematic representation of biological neuron: The dendrites receive the input signal. The soma processes the signal with a nonlinear function. The axon transmits the signal to the axonal endings. The signal is transmitted to the subsequent neuron via the synapse [55]

A nonlinear function is then applied to the signals within the soma. In essence, the soma sums the dendritic inputs together and evaluates the result according to a threshold function, if the result is larger than the threshold an action potential fires. This action potential then propagates away from the soma, down the axon. The output signal leaves the neuron through the axonal endings. This simplified model of the neuron serves as the basis for the artificial neuron.

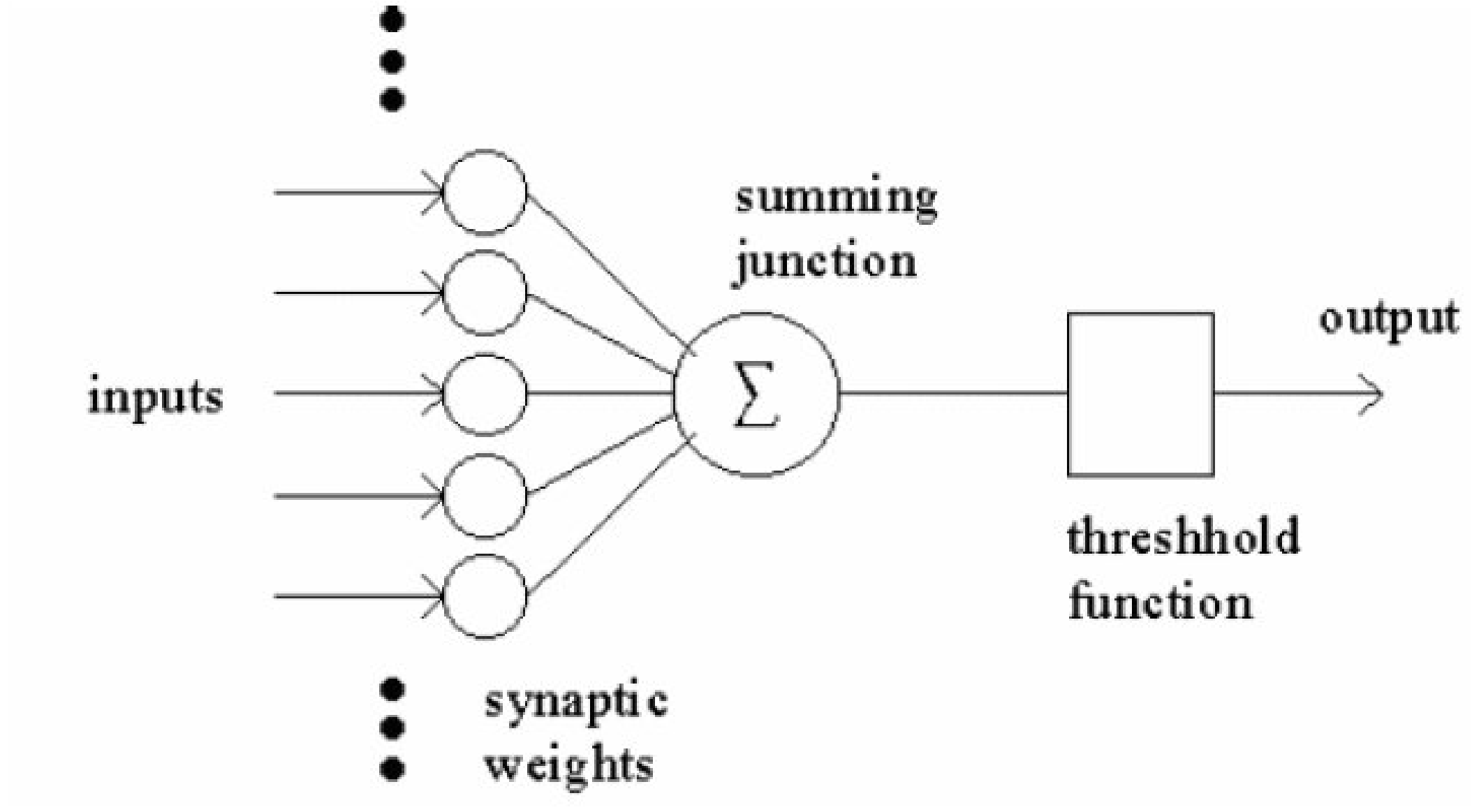

**Figure (7)** Schematic representation of biological neuron: The artificial neuron preserves the computational function of the biological neuron [55]

The brain is composed of networks of these biological neurons. The output of one neuron serves as input to one or more other neurons. However, the axon of one neuron is not directly connected to the dendrite of another. Rather a gap, known as the synapse, exists between them. There are many factors within the synapse which affect how the signal from the output neuron is transmitted to the input neuron. So while the connection between neurons determines which neurons influence other neurons, the nature of the synapse determines the extent of this influence. The nature of the synapse is constantly changing with each experience the synapse adapts so connections are either strengthened or weakened. Thus, the synapse is believed to be the primary source of learning. The state of the synapse encapsulates our acquired knowledge and controls our information processing. In an ANN, the synapse is modeled by a weight which indicates how much influence one artificial neuron has on another to which it is connected. Using this scheme the artificial neuron performs a weighted sum of the inputs, compares this to a threshold and outputs the result to other neurons down the line.

For ANNs, there are various architectures in which neurons can be arranged to specify their connections. The neurons are first clumped into layers. The interlayer connections are primarily divided into two groups: feedforward and feedback. A feedforward network has only unidirectional connections. In this way, signals propagate forward from the input layer to the output layer. In feedback networks, a particular layer may be connected to the next layer or any of the previous layers. This arrangement leads to a feedback loop for the signals. We must design the architecture of the network. The number of neurons in each layer must be decided; as well inter- and intra-layer connections must be hardwired. While the connections are hardwired, the weights between neurons can be changed by the network. This changing of weights causes the network to learn a solution to a problem. The optimal weights are determined by the network through training. There are various learning paradigms which guide this process. In a supervised learning scheme, the network is presented with an input and the correct output. The network compares its prediction with the true result and adjusts its weights accordingly. If the network is not presented with the desired output it must learn to organize its weights without being able to measure its predictive success and minimize its error. In such an unsupervised scheme, neurons compete for the opportunity to update their weights, resulting in self-organization.

For the case of protein structural prediction we are primarily concerned with supervised learning . The network is presented with an amino acid sequence and the corresponding structure. The ANN updates its weights as it minimizes the error between its prediction and the true structure. Learning takes place with each new case as the network is trained to predict the protein structure from the amino acid sequence. On its own, the network discovers the underlying rules governing folding. The extent of
learning is tested by supplying the network with a previously unseen sequence and comparing its prediction with the known structure.

To solve the protein folding problem is to know the rules governing folding. A viable approach to the problem must work towards this end. The criteria by which one can measure the capability of their approach to reach this goal are as follows:

**Firstly**, a solution to the protein folding problem must accurately predict protein structure from an amino acid sequence. This is a challenging problem because of the huge number of forces at work. The electrostatic forces of repulsion and attraction, chemical bonding and quantum effects must all be combined. Thus to solve the problem requires immense computational power and the ability to conceptualize and account for all these factors.

**Secondly**, the approach must be credible. It must provide a solution that agrees with the basic principles of chemistry and physics, is applicable to all proteins (not just a class), and is independent of those factors not scientifically relevant to the outcome. Essentially, we are asking that the solution comes as a scientific theory: compatible with other truths, broad in scope, and bounded in independent variables.

We can use the above criteria to evaluate if it is even possible to use ANNs to solve the protein folding problem. In what follows, we use these criteria to examine this possibility and to compare ANNs and deterministic algorithms as methods for approaching our problem. To begin this evaluation it is important to note that we have not yet been successful in our efforts to solve the protein folding problem. This suggests that the scope of the problem may be too large for the human brain to conceive (at least at this point in time). In all our efforts to simplify the problem, we disregard critical influences of the folding process.

The result is an inaccurate prediction and an unsuccessful algorithm. On the other hand, this is precisely the kind of task at which ANNs excel. ANNs are very effective at combining multiple non-linear factors, such as those affecting folding. They exceed human computing power for detecting trends embedded deep within the data. This exceptional ability to conceptualize allows them to predict the resulting structure with much more accuracy. The inadequacy of ANNs is their inherent dependency on their training set. One neural net, having been trained on certain data, may produce a prediction quite different from that obtained from another neural net, trained on different data. This poses problems not only for accuracy but also to the credibility of ANNs as tools for scientific research. This is currently the largest drawback for using ANNs. Deterministic approaches can avoid this problem with credibility. However, we should not hasten to dismiss ANNs based on the credibility issue. New research has shown that it is possible to extract useful information from the synaptic weights of the network. If we were able to extract the rules from the network, in this or a similar manner, then we could generalize them to make them compatible with other networks. Once formulated as a deterministic algorithm, the issues of credibility would be obsolete. Thus, with more research in this area, ANNs could be used to conceive a deterministic algorithm for solving the protein folding problem. Speed at which a program executes is determined by the specific design rather than the solution methodology. For deterministic programs, execution speed is proportional to the complexity of the algorithm. Improved accuracy usually comes with increased complexity. Thus, any hope for an accurate deterministic program would be very slow. With ANNs, speed is dependent on architecture. The more layers we have, the more time we need. As well, feedback is much slower than feed forward. An ideal implementation would increase speed while retaining accuracy.
Based on the previous evaluation, the ANN approach is a feasible alternative for addressing the protein folding problem. Although the ANN approach is not without its faults, it remains a strong contender for solving the protein folding problem. Keeping this in mind, we should dedicate ourselves to improving the accuracy and credibility of ANNs. To improve accuracy, we can experiment with various architectures and training sets. As well, we should try to give the network more biological information. Information such as size, electrical properties, chemical reactivity and hydrogen bonding capacity of the amino acids involved may be quite useful to the network.

To address the credibility issue, we must figure out how to extract information (in the form of rules) from the network. These rules should then be reformulated to be compatible with other networks, applicable to all sequences and compliant with the truths of chemistry and physics. This is a focal point for further research and should be thoroughly explored.

## 3.2.2 INTRODUCTION TO FEED-FORWARD NETS

Feed-forward nets are the most well-known and widely-used class of neural network. The popularity of feed-forward networks derives from the fact that they have been applied successfully to a wide range of information processing tasks in such diverse fields as speech recognition, financial prediction, image compression, medical diagnosis and protein structure prediction; new applications are being discovered all the time.

In common with all neural networks, feed-forward networks are trained, rather than programmed, to carry out the chosen information processing tasks. Training a feed-forward net involves adjusting the network so that it is able to produce a specific output for each of a given set of input patterns. Since the desired inputs are known in advance, training a feed-forward net is an example of supervised learning.

### 3.2.3 THE FEED-FORWARD ARCHITECTURE

Feed-forward networks have a characteristic layered architecture, with each layer comprising one or more simple processing units called artificial neurons or nodes. Each node is connected to one or more other nodes by real-valued weights (parameters), but not to nodes in the same layer. All feedforward nets have an input layer and an output layer. A net with only an input and an output layer is called a single layer net or single layer perceptron (not a two-layer net, because the input layer - which presents a given input pattern to the net but serves no computational function - is not counted).
Feed-forward nets are generally implemented with an additional node - called the bias unit - in all layers except the output layer. Typically the output of each bias unit is 1.0 for all patterns in the data set. The output *y* of each (non-input) node in the network (for a given pattern *p*) is simply the weighted sum of its inputs, i.e.

$$a_{i,p} = \sum_{j} w_{ij} y_{j,p}$$

$$y_{i,p} = \int a_{i,p}$$

The squashing function *f(x)*, which is required to be both monotonic and differentiable, is typically the sigmoid or logistic function, given by

$$\int(x) = \frac{1}{1+e^{-x}}$$

### 3.2.4 TRAINING A FEED-FORWARD NET

Feed-forward nets are trained using a set of patterns known as the training set for which the desired outputs are known in advance - a process known in the neural network literature as supervised learning. Every pattern must have the same number of elements as the net has input nodes, and every target the same number of elements as the net has output nodes. Taken together, a training pattern and its associated target are known as a training pair. Prior to training, the network weights are initialised to small random values. A training algorithm is then used to progressively reduce the total network error by iteratively adjusting the weights. The best-known and simplest training algorithm for feed-forward networks is backpropagation, based on the venerable classical optimisation method steepest descent.

When training a neural network, it is important not to lose sight of the underlying purpose, which is *not* to learn the training set to the highest degree of accuracy. Rather, the aim is to generate a network that is good at classifying patterns similar to, but not identical to, patterns in the training set - i.e. a network that has the ability to generalise. A considerable amount of neural network research has been concerned with specifying the conditions necessary to generate a network that will generalise well. A variety of factors have been identified, including:

1. *The number of training patterns versus the number of network weights.* If the number of network weights is too large compared with the number of training patterns, there is a risk of *over-fitting*. One 'rule of thumb' asserts that there should be at least 20 times as many patterns at network weights.

2. *The number of hidden nodes.* In case of too few hidden nodes the network will be unable to learn a given tasks and in case of too many its generalisation will be poor. In fact, the 'basic' secondary structure prediction task can be learned by a single layer perceptron, i.e. by a feed-forward net with zero hidden nodes.

3. *The number of training iterations*. In case of too few training iterations network will be unable to extract important features from the training set; and in case of too many net will begin to learn the details of the training set to the detriment of its ability to abstract general features - a process known as over-training.

There are seven target structural classes given in DSSP (Dictionary of Protein Secondary Structure ) which include G,H,I,T,E,B,S [ G is 3-turn helix (3_10 helix) minimum length 3 residues, H is 4-turn helix (alpha helix) minimum length 4 residues, I is 5-turn helix (pi helix) minimum length 5 residues, T is hydrogen bonded turn (3, 4 or 5 turn), E is beta sheet in parallel and/or anti-parallel sheet conformation (extended strand) minimum length 2 residues, B is residue in isolated beta-bridge (single pair beta-sheet hydrogen bond formation). S is bend (the only non-hydrogen-bond based assignment)]. In DSSP residues which are not in any of the above conformations is designated as ' ' (space).
Both the residues and target classes are encoded in binary format [for example Alanine (A): 0 0 0 0 1 Helix (H): 0 0 1 etc.]. Thus each pattern presented to the network comprises *n**5 inputs for a window of size *n*. The advantage of this sparse encoding scheme is that it does not introduce an artificial ordering; each amino acid and secondary structure type is given equal weight. The main disadvantage is that it entails a large number of network parameters.

## 3.2.5 MEASURING PERFORMANCE

The most popular statistical measure of performance is simply the percentage of correctly classified residues, known as *Q3*.The main problem with *Q3* as a measure is that it fails to penalise the network predictions (e.g. non-helix residues predicted to be helix) or under-predictions (e.g. helix residues predicted to be non-helix).

## 3.2.6 DRAWBACKS WITH THE BASIC APPROACH

Predictions are based on a limited local context (the window size). No account is taken of non-local factors, yet there is considerable evidence that long-range interactions constrain the formation of protein secondary structure. Predictions are based on a limited amount of biological information. For example, the network is not presented with any evolutionary information or with information about the physico-chemical properties of proteins. No account is taken of what we know about the principles underlying protein structure, e.g. knowledge about what constitutes a 'protein-like' prediction. The predictions are uncorrelated, i.e. each prediction is made in isolation, taking no account of the predictions for neighbouring residues.

# SYSTEM ANALYSIS & APPROACHES

**Dataset for study:**

For our study we have collected protein sequence-structure data set from the Jpred distribution list by the Barton Group at University of Dundee. We have selected 507 data from the database available as free distribution with above group. All sequences in this set have been compared pairwise, and are non redundant to a 5 SD cut-off. For training and testing we have taken 407 and 100 protein sequence-structure pair respectively.

## 4.1 APPROACH FOR HIDDEN MARKOV MODEL

**General approach to build the HMM based first model:**

The work has been divided in two models. First model is related with prediction of protein secondary structures from protein sequences and have taken Protein sequence as observed state and Protein secondary structure as hidden state. Second model is related with prediction of protein sequence from protein secondary structures and have taken protein secondary structure as observed state and protein sequence as hidden state.
It has been assumed that N, the number of states in the model. Although the states are hidden, for many practical applications there is often some physical significance attached to the states or to sets of states of the model. Generally the states are interconnected in such a way that any state can be reached from any other state. We denote the individual state as $S = \{S_l, S_{2,} \ldots.. S_N\}$, and the state at time t as $q_t$.

For **first model we have done the following steps for the protein sequence-structure pairs**.

**Step i) Calculation of initial probability for hidden state.**

As transition is occurring among the protein secondary structures (hidden states) so initial probability, $\pi_i$ for i-th protein secondary structures will be the ratio of " how many times that particular state occurred at first position in the training dataset(m)" to " how many times any state occurred at first position i.e., total number of training protein sequence structure pairs(M) ".

$$\pi_i = \frac{m}{M}$$

where $1 \leq i \leq N$.

**Step ii) Calculation of transition probability for training dataset.**

Here transition is occurring among the hidden states (protein secondary structure). As we are using DSSP structural symbols, the transition matrix for training dataset will be of 8 × 8 matrix.

Transition probability from one state to a particular state will be the ratio of "how many times transition occur from one state, ***i*** to that particular state, ***j*** (n)" to "total transition from one state to any other states (N)"

$$a_{ij} = p(q_t + 1 = S_j \mid q_t = S_i) = \frac{n}{N}$$

for $1 \leq \{i, j\} \leq N$ where N =8 for our case.

**Step iii) Calculation of emission probability for training dataset.**

Emission is occurring when transition occur among hidden states. As any hidden state can emit any one of the twenty amino acids, so the emission matrix will be of 8 × 20 matrix.

Emission probability of an observed state by a particular hidden state will be the ratio of " how many times that observed state, ***k*** is emitted by that particular hidden state, ***j***" (r) to " total number of emission of any observed state by that particular hidden state. ***j***" (R).

$$b_j(k) = p(v_k \text{ at } t \mid q_t = S_j) = \frac{r}{R}$$

for $1 \leq j \leq 8$ and $1 \leq k \leq 20$.

**Step iv) Calculation of forward, backward and forward-backward variables:**

Using transition probability and emission probability we have calculated the forward variable ($\alpha_i$) and backward variable ($\beta_i$) for i-th hidden state within a hidden sequence of state 8]. Then we have calculated the forward-backward variable ($\gamma_i$) for i-th hidden state using following formula:

$$\pi_i = \frac{m}{M}$$

**Step v) Calculation of prediction efficiency:**

For comparison of predicted and actual protein secondary structure for the test data, we calculated the number of matches for the predicted protein secondary structures and calculated the efficiency for three hidden states ($Q_3$ score) for each test data as

$$\mathbf{Q_3 = (N_{match} / N_{total}) \text{ X } 100}$$

Total efficiency for test dataset is calculated as the mean of the $Q_3$ scores.

**Second model: Prediction of protein sequences by protein secondary structures**

Similarly we designed a model for protein sequence prediction from protein secondary structures. In this model protein secondary structures and sequences are taken as observed and hidden states respectively.

Efficiencies of both the model were compared to explore and compare the conservation of protein sequence and structure.

## 4.2 APPROACH FOR ARTIFICIAL NEURAL NETWORK

**General approach to build the ANN based first model:**

**Step1) Representation of sequence and secondary structures into binary format:**

We converted sequence and secondary structures into a binary number format. To represent 20 amino acids at least 5 binary digits are needed. For example, if the number of residue in a sequence is 10, it will create a 10 by 5 matrix of binary digits whose each row signifies each residue.

To represent eight secondary structure elements at least 3 binary digits are needed. For example, if the number of conformation in a secondary structure sequence is 10, it will create a 10 by 3 matrix of binary digits whose each row signifies each conformation ('GHITEBSU').Binary format of sequence and secondary structure were used as input and output respectively to train the feed forward net.

**Step 2) Training feed-forward network to get adjusted weights:**

We implemented weight-updating following feed-forward ANN architecture and using sigmoidal function y = 1/ (1+exp (-a)); where a is the net at the output nodes of the network Weight adjustment will be following 200 iterations per epoch i.e., per residue position. For example, for a sequence of size 40 it runs 40x200 times.

**Step 3) Prediction of secondary structure:**

Once the adjusted weights will be obtained, adjusted weights and 13 residue window is used to get 3 digit binary numbers. These three digit binary number will give the predicted structure.

**Step 4) Calculation of Q3score:**

Calculation of the Q3score (i.e., the percentage of matched structural motifs) given test structure and resultant structure using the following formula

$$Q_3 = (N_{match} / N_{total}) \times 100$$

**General approach to build the ANN based second model:**

In the second model binary format of the secondary structure is used as input and sequence is used as output. Using the adjusted weights given by feed forward net test sequence is predicted after this calculation of Q3score will be done.

Once we got the efficiency results for artificial neural network, the conclusions made on the basis hidden markov model is cross checked if both the model are supporting for the same conclusions.

# SYSTEM STRUCTURE & REQUIREMENTS

## 5.1 DEVELPOMENT TOOLS USED

**MATLAB** stands for "Matrix Laboratory" and is a numerical computing environment and fourth-generation programming language. Developed by The MathWorks, MATLAB allows matrix manipulations, plotting of functions and data, implementation of algorithms, creation of user interfaces, and interfacing with programs written in other languages, including C, C++, and Fortran.

Although MATLAB is intended primarily for numerical computing, an optional toolbox uses the MuPAD symbolic engine, allowing access to symbolic computing capabilities. An additional package, Simulink, adds graphical multi-domain simulation and Model-Based Design for dynamic and embedded systems.

In 2004, MathWorks claimed that MATLAB was used by more than one million people across the industry and the academic world. MATLAB users come from various backgrounds of engineering, science, and economics. Among these users are Massachusetts Institute of Technology, RWTH Aachen University, ABB Group, Boeing, Ford Motor, Halliburton, Lockheed Martin, Motorola, NASA, Novartis, Pfizer, Philips, Toyota, and UniCredit Bank.

MATLAB was created in the late 1970s by Cleve Moler, then chairman of the computer science department at the University of New Mexico. He designed it to give his students access to LINPACK and EISPACK without having to learn Fortran. It soon spread to other universities and found a strong audience within the applied mathematics community. Jack Little, an engineer, was exposed to it during a visit Moler made to Stanford University in 1983. Recognizing its commercial potential, he joined with Moler and Steve Bangert. They rewrote MATLAB in C and founded The MathWorks in 1984 to continue its development. These rewritten libraries were known as JACKPAC. In 2000, MATLAB was rewritten to use a newer set of libraries for matrix manipulation, LAPACK.

MATLAB was first adopted by control design engineers, Little's specialty, but quickly spread to many other domains. It is now also used in education, in particular the teaching of linear algebra and numerical analysis, and is popular amongst scientists involved with image processing.

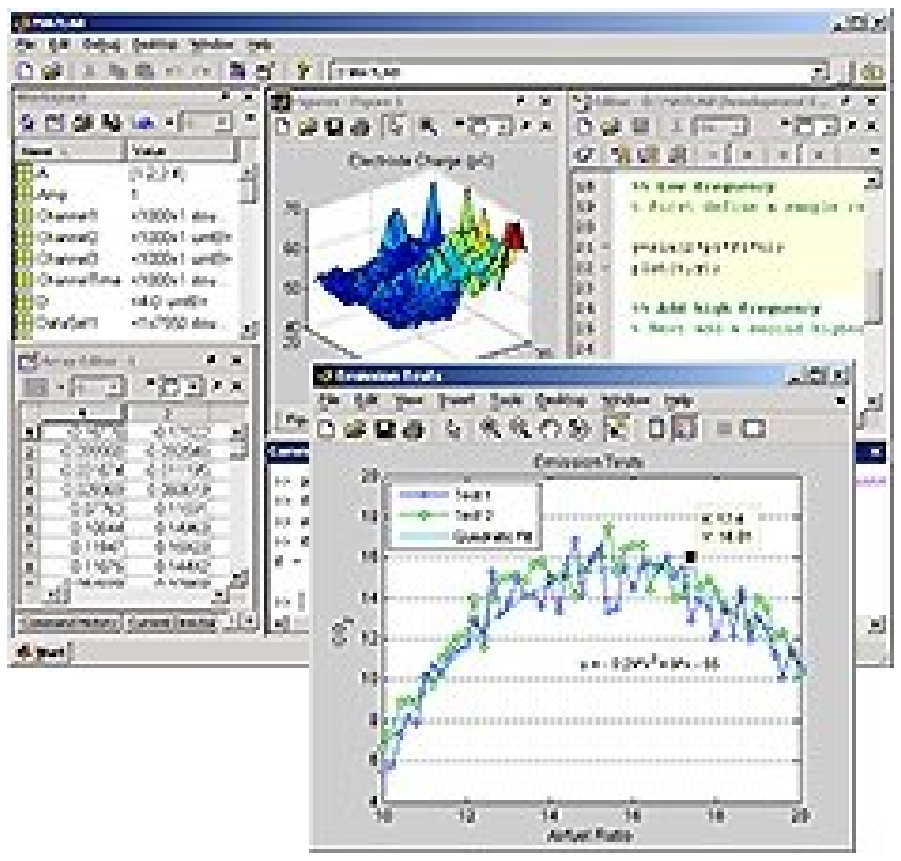

## MATLAB Versions

Versions of MATLAB are available for almost all major computing platforms. Our material was produced and tested on the version designed for the Microsoft Windows environment. The vast majority of it should work with other versions, but no guarantees can be offered.

Of particular interest are the *Student Versions* of MATLAB. Prices are generally below $100. These systems include most of the features of the language, but no matrix can have more than 8,192 elements, with either the number of rows or columns limited to 32. For many applications this proves to be of no consequence. At the very least, one can use a student version to experiment with the language.

The Student Editions are sold as books with disks enclosed. They are published by Prentice-Hall and can be ordered through bookstores.

In addition to the MATLAB system itself, Mathworks offers sets of *Toolboxes*, containing MATLAB functions for solving a number of important types of problems. Of particular interest to us is the *optimization toolbox*, which will be discussed in a later section.

## Matrices as Fundamental Objects

MATLAB is one of a few languages in which each variable is a matrix (broadly construed) and "knows" how big it is. Moreover, the fundamental operators (e.g. addition, multiplication) are programmed to deal with matrices when required. And the MATLAB environment handles much of the bothersome housekeeping that makes all this possible. Since so many of the procedures required for Macro-Investment Analysis involve matrices, MATLAB proves to be an extremely efficient language for both communication and implementation.

## Limitations

MATLAB is a proprietary product of The MathWorks, so users are subject to vendor lock-in. Although MATLAB Builder can deploy MATLAB functions as library files which can be used with .NET or Java application building environment, future development will still be tied to the MATLAB language. MATLAB, like Fortran, Visual Basic and Ada, uses parentheses, e.g. y = f(x), for both indexing into an array and calling a function. Although this syntax can facilitate a switch between a procedure and a lookup table, both of which correspond to mathematical functions, a careful reading of the code may be required to establish the intent.
Mathematical matrix functions generally accept an optional argument to specify a direction, while others, like plot, do not, and so require additional checks. There are other cases where MATLAB's interpretation of code may not be consistently what the user intended[citation needed] (e.g. how spaces are handled inside brackets as separators where it makes sense but not where it doesn't, or backslash escape sequences which are interpreted by some functions like fprintf but not directly by the language parser because it wouldn't be convenient for Windows directories). What might be considered as a convenience for commands typed interactively where the user can check that MATLAB does what the user wants may be less supportive of the need to construct reusable code.
Array indexing is one-based which is the common convention for matrices in mathematics, but does not accommodate any indexing convention of sequences that have zero or negative indices. For instance, in MATLAB the DFT (or FFT) is defined with the DC component at index 1 instead of index 0, which is not consistent with the standard definition of the DFT in any literature. This one-based indexing convention is hard coded into MATLAB, making it difficult for a user to define their own zero-based or negative-indexed arrays to concisely model an idea having non-positive indices. A workaround can be constructed to create an ancillary frequency labeling array with element values equal to the index less 1 that would be used instead of the index, or the MATLAB programmer can remember to subtract 1 from the index obtained from functions that return an index such as find(), min(), max().

MATLAB built-in datatypes are always passed by value. Therefore, all input parameters to a function are usually copied (later MATLAB releases introduced lazy copy where if an input parameters are not being altered, a copy is not being made, however this has certain restrictions). Instances of user-defined classes are also copied and passed by value as the default, however user-defined classes can be made to exhibit reference behavior by inheriting from the abstract handle class. Variables of such classes store a reference to the instance. An alternative to using references is global variables, however MATLAB JIT does not support globals, as well as structures, therefore the code performance will degrade.
The MATLAB editor does not have code completion, references searches, or refactoring. Since MATLAB is weakly typed, these tools would be hard, if not impossible, to provide in future releases since it is unknown which methods of an object can be called without knowing its type. Without code completion, there is a tendency to use short cryptic names for variables, function and methods, making code hard to read. Without code completion or reference searches it can take a long time to text search for what functions and methods can be called in a given context. Without refactoring, it is time consuming and error prone to change variable, method, or function names, or to modify an API.
Productivity of a development team working on a large software project in MATLAB will likely be several times slower and result in a higher defect rate than development in a language and IDE with type checking, code completion, reference search, refactoring tools, and unit testing support.

It is very difficult to employ agile programming practices, such as Extreme Programming, in MATLAB due to the lack of refactoring tools.

MATLAB lacks any native support for GPU acceleration such as OpenCL, Microsoft's DirectCompute, or nVidia's CUDA.

## 5.2 ALGORITM USED

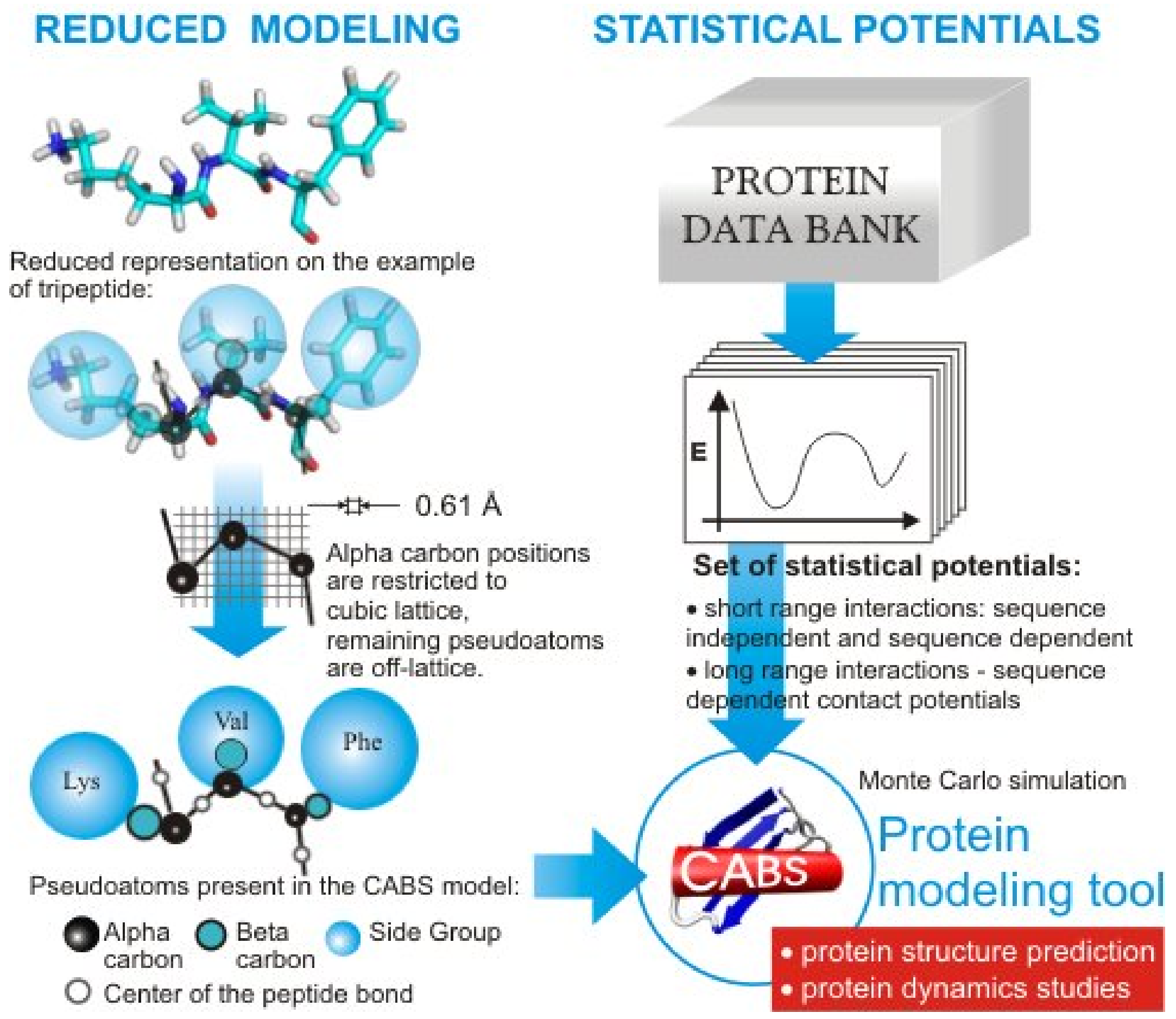

An algorithm has been developed to improve the success rate in the prediction of the secondary structure of proteins by taking into account the predicted class of the proteins. This method has been called the 'double prediction method' and consists of a first prediction of the secondary structure from a new algorithm which uses parameters of the type described by Chou and Fasman, and the prediction of the dass of the proteins from their amino acid composition. These two independent predictions allow one to optimize the parameters calculated over the secondary structure database to provide the final prediction of secondary structure. This method has been tested on 59 proteins in the database (i.e. 10 322 residues) and yields 72% success in class prediction, 61.3% of residues correctly predicted for three states (helix, sheet and coil) and a good agreement between observed and predicted contents in secondary structure.

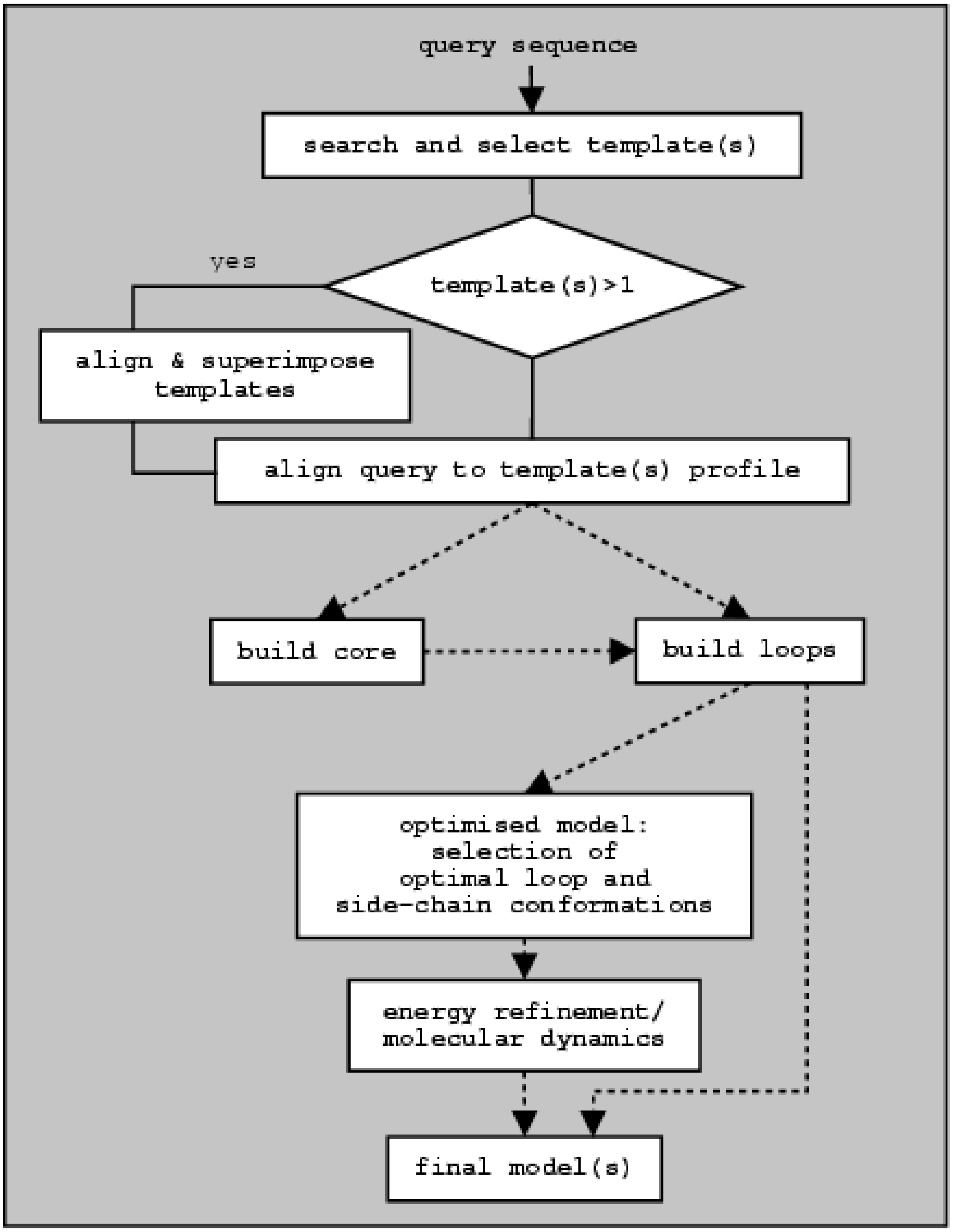
query sequence
search and select template(s)
template(s)>1
yes
align & superimpose templates
align query to template(s) profile
build core
build loops
optimised model: selection of optimal loop and side-chain conformations
energy refinement/ molecular dynamics
final model(s)

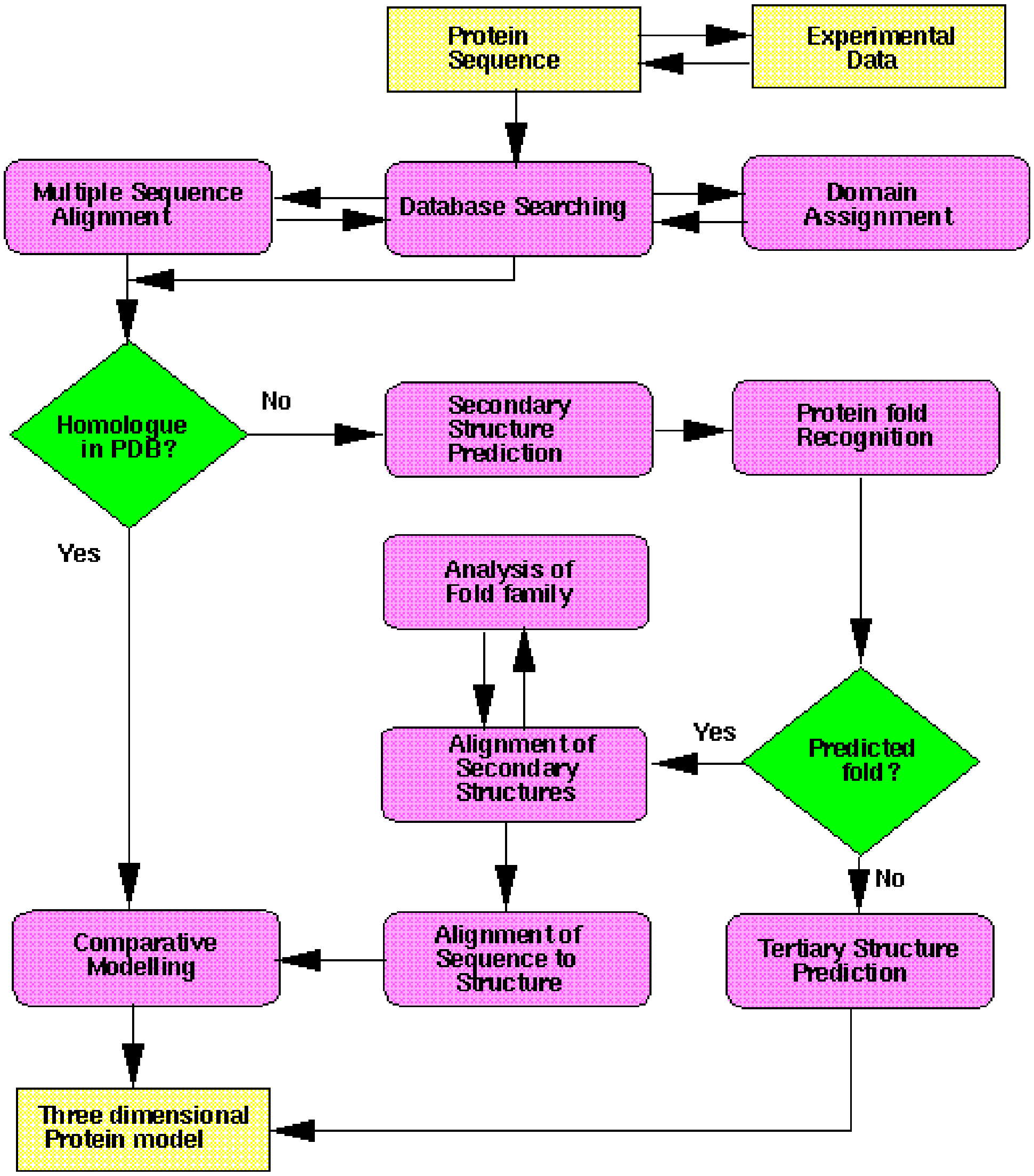
Protein Sequence
Experimental Data
Multiple Sequence Alignment
Database Searching
Domain Assignment
Homologue in PDB?
No
Yes
Secondary Structure Prediction
Protein fold Recognition
Analysis of Fold family
Alignment of Secondary Structures
Yes
Predicted fold ?
No
Comparative Modelling
Alignment of Sequence to Structure
Tertiary Structure Prediction
Three dimensional Protein model

## 5.3 DATA FLOW

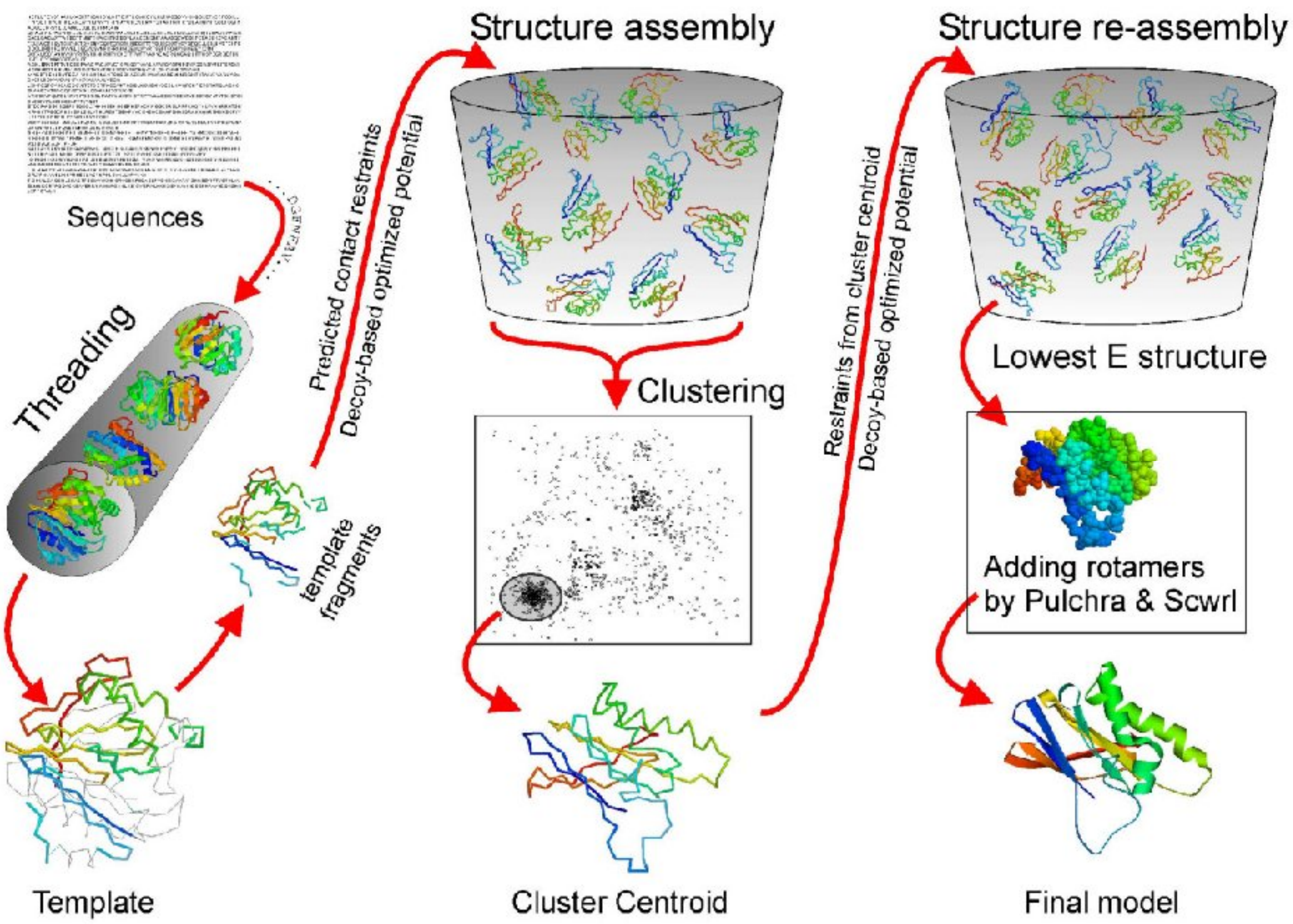

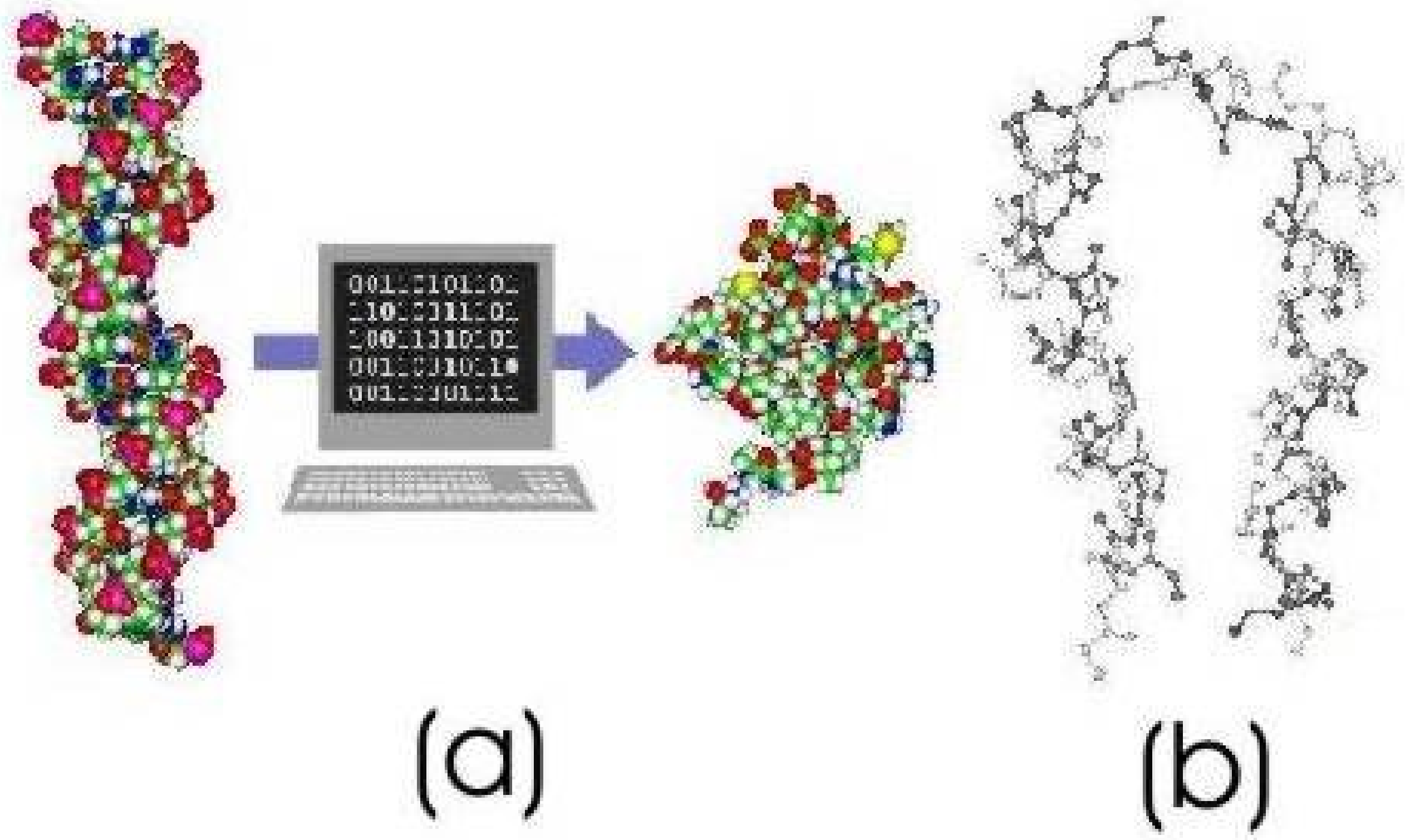

(a) **Shows the MATLAB module and;**
(b) **Shows the final model of protein structure predicted.**

### 5.3.1 DFD of the Model

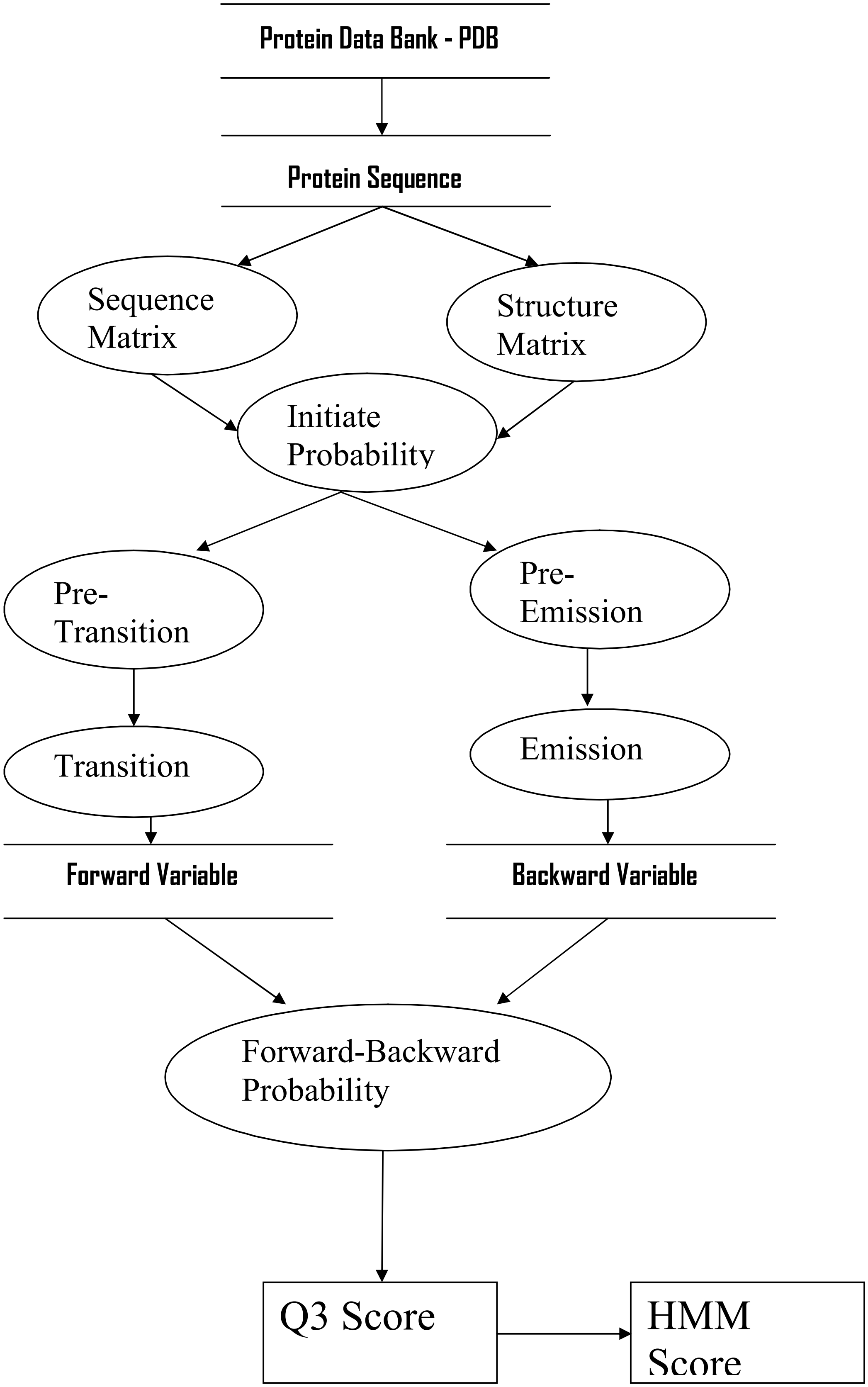

# DESIGN & DOCUMENTATION

## 6.1 DESIGN OBJECTIVE

The Specification (i.e. the “outside” view) of a program should obviously be as free as possible of aspects imposed by “how” the program will work (i.e. the “inside” view). It is seldom a document from which coding can directly be done. So design fills gap between specification and coding; taking the specifications, deciding how the program will be organized, and the methods it will use, in sufficient detail as to be directly codeable.

The design needs to be

- Correct and complete
- Understandable
- At the right level
- Maintainable, and to facilitate maintenance of the produced code

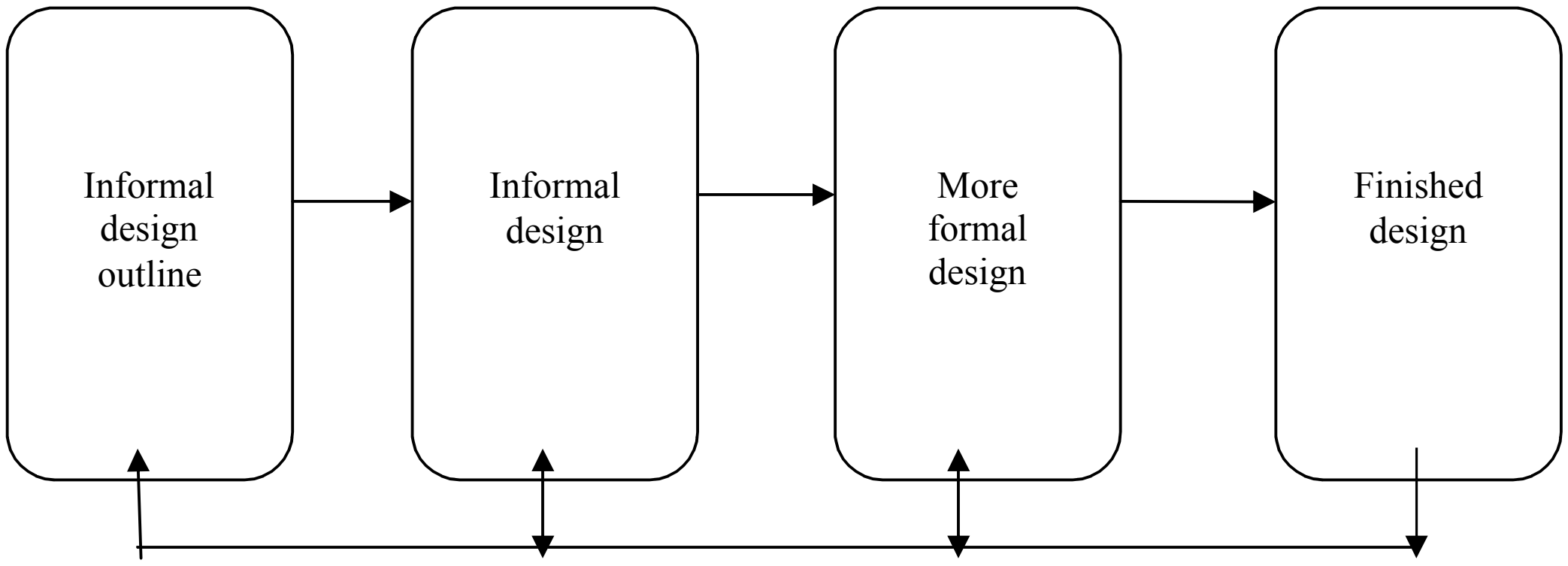

**The transformation of an informal design to a detail design**

## 6.2 SYSTEMIZED BLOCK DIAGRAM OF SYSTEM

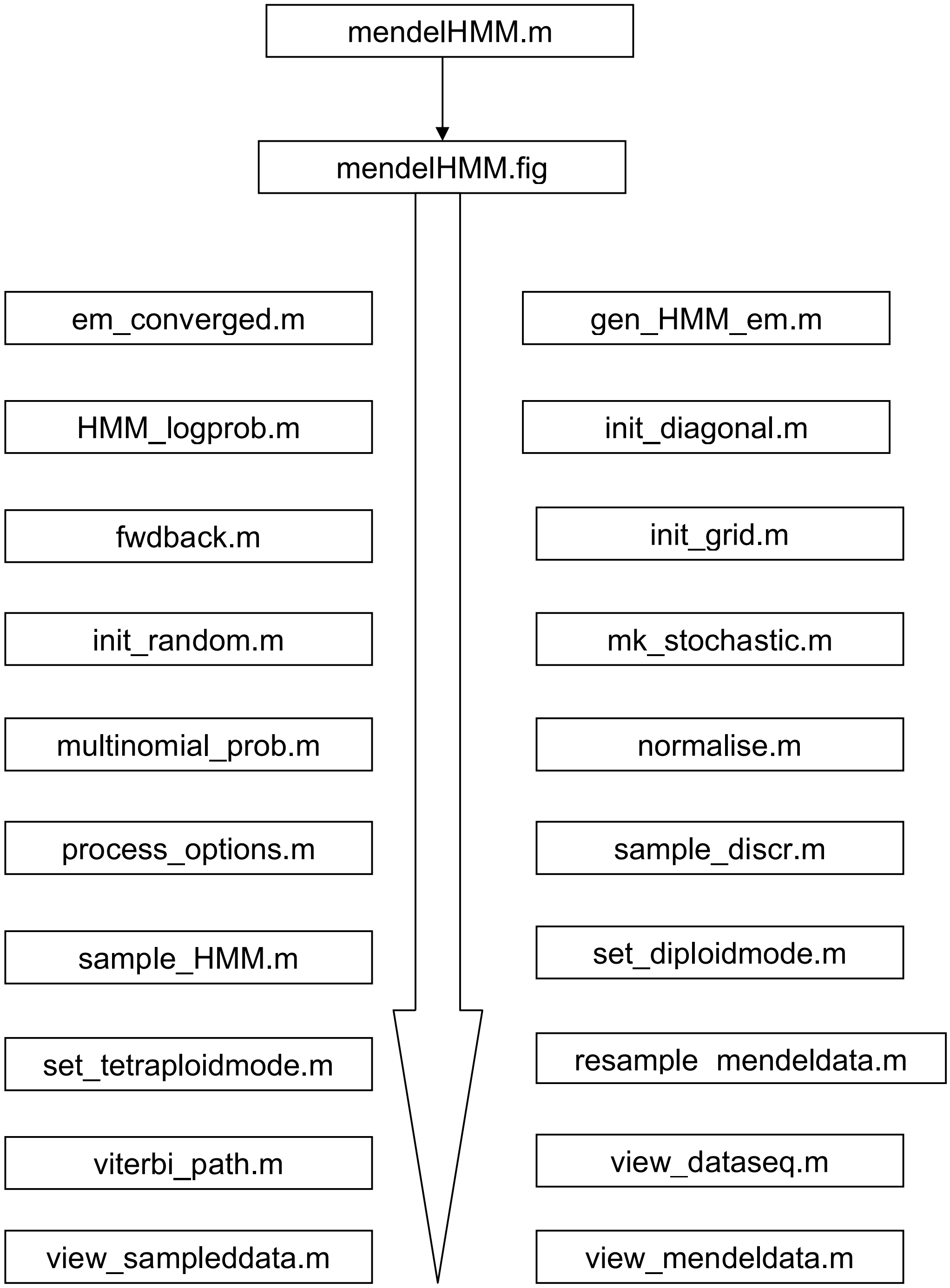

## 6.3 CODING

### <mendelHMM.m>

```
function varargout = mendelHMM(varargin)

gui_Singleton = 1;
gui_State = struct('gui_Name',       mfilename, ...
                   'gui_Singleton',  gui_Singleton, ...
                   'gui_OpeningFcn', @mendelHMM_OpeningFcn, ...
                   'gui_OutputFcn',  @mendelHMM_OutputFcn, ...
                   'gui_LayoutFcn',  [] , ...
                   'gui_Callback',   []);
if nargin & isstr(varargin{1})
    gui_State.gui_Callback = str2func(varargin{1});
end

if nargout
    [varargout{1:nargout}] = gui_mainfcn(gui_State, varargin{:});
else
    gui_mainfcn(gui_State, varargin{:});
end
% End initialization code - DO NOT EDIT


% --- Executes just before mendelHMM is made visible.
function mendelHMM_OpeningFcn(hObject, eventdata, handles, varargin)
% This function has no output args, see OutputFcn.
% hObject    handle to figure
% eventdata  reserved - to be defined in a future version of MATLAB
% handles    structure with handles and user data (see GUIDATA)
% varargin   command line arguments to mendelHMM (see VARARGIN)

% Choose default command line output for mendelHMM
handles.output = hObject;

% Update handles structure
guidata(hObject, handles);

% UIWAIT makes mendelHMM wait for user response (see UIRESUME)
% uiwait(handles.figure1);


% --- Outputs from this function are returned to the command line.
function varargout = mendelHMM_OutputFcn(hObject, eventdata, handles)
% varargout  cell array for returning output args (see VARARGOUT);
% hObject    handle to figure
% eventdata  reserved - to be defined in a future version of MATLAB
% handles    structure with handles and user data (see GUIDATA)

% Get default command line output from handles structure
varargout{1} = handles.output;


% --- Executes on button press in pushparametersview.
function pushparametersview_Callback(hObject, eventdata, handles)
% hObject    handle to pushparametersview (see GCBO)
```

```
% eventdata  reserved - to be defined in a future version of MATLAB
% handles    structure with handles and user data (see GUIDATA)

global prior2 transmat2 obsmat2
% Must converge parameters to strings:
strtrans=num2str(transmat2, '%7.3f')
strobs=num2str(obsmat2, '%7.3f')

% Estimated parameters are shown on a new figure where text objects are
% created:
figure('Name',['mendelHMM EM-estimated model: ']), clf
handl_transmattxt = uicontrol('Style', 'text', 'String', 'Transition
matrix P = ',...
    'Position', [20 400 150 30] );
handl_transmatview = uicontrol('Style', 'text', 'String', strtrans,...
     'Position', [250 250 350 200]);
handl_obsmattxt = uicontrol('Style', 'text', 'String', 'Obs. matrix E =
',...
    'Position', [20 100 150 30] );
handl_obsmatview = uicontrol('Style', 'text', 'String', strobs,...
    'Position', [250 10 350 200] );



% --- Executes on button press in pushestimate.
function pushestimate_Callback(hObject, eventdata, handles)
% hObject    handle to pushestimate (see GCBO)
% eventdata  reserved - to be defined in a future version of MATLAB
% handles    structure with handles and user data (see GUIDATA)

global prior1 transmat1 obsmat1 dataE LL prior2 transmat2 obsmat2

% extracting selected initialization metod:
method=get(handles.popinitializationNum,'Value');
% extracting best loglik TEXT from GUI:
startLL=get(handles.textinitialbestlikelihoodNum,'String');

maxround=5; %-----may be changed here!!!!!!----
round=1; % when initialization method is random (=2), we want many
guessrounds:
while (round == 1) | ((method == 2) & (round <=maxround))
    % First part of test is to assure that initialization is ready
(transmat1-variables):
    if (round ==1 & str2num(startLL) == -Inf) | round >=2
        round
        runinitialization(method, handles);
    end
    [LL, priorEM, transmatEM, obsmatEM] = gen_HMM_em(dataE, prior1,
transmat1, obsmat1, handles, 'max_iter', 15);
    % use model to compute log likelihood
    loglik = HMM_logprob(dataE, priorEM, transmatEM, obsmatEM)
    % log lik is slightly different than LL(end), since it is computed
after the final M step
    set(handles.textlikelihoodNum,'string',loglik) % updated likelihood
presented at GUI
    best_llstr=get(handles.textbestlikelihoodNum,'String'); % extracting
best loglik TEXT from GUI
    if loglik > str2num(best_llstr)
        set(handles.textbestlikelihoodNum,'string',loglik) % updated
best score at GUI
```

```
        % update parameters for best score:
        prior2=priorEM;
        transmat2=transmatEM;
        obsmat2=obsmatEM;
    end
    round=round+1;
end %while

% LL
transmatEM
obsmatEM

% --- Executes on button press in pushshample.
function pushshample_Callback(hObject, eventdata, handles)
% hObject    handle to pushshample (see GCBO)
% eventdata  reserved - to be defined in a future version of MATLAB
% handles    structure with handles and user data (see GUIDATA)

% We want to produce new set of training data:
global prior transmat obsmat hidX dataE

set(handles.popusedataNum,'Value',1) %present 'No' instead of MendelData
at GUI
updateGUI(handles); % The GUI and the basic matrices in the HMM is
uppdated
numbex=str2num(get(handles.editnumNum,'String')) %first extracting
number of experiments
T=str2num(get(handles.editlengthNum,'String')) %then extracting length
of each sequence
[hidX,dataE] = sample_HMM(prior, transmat, obsmat, T, numbex)


% --- Executes on button press in pushlikelihoodview.
function pushlikelihoodview_Callback(hObject, eventdata, handles)
% hObject    handle to pushlikelihoodview (see GCBO)
% eventdata  reserved - to be defined in a future version of MATLAB
% handles    structure with handles and user data (see GUIDATA)

global LL
figure('Name',['mendelHMM: likelihood in EM training ']), clf
plot(LL,'*');
xlabel('iteration')
ylabel('loglikelihood')
title('EM training of HMM');


% --- Executes on button press in pushviewdata.
function pushviewdata_Callback(hObject, eventdata, handles)
% hObject    handle to pushviewdata (see GCBO)
% eventdata  reserved - to be defined in a future version of MATLAB
% handles    structure with handles and user data (see GUIDATA)

view_dataseq %display graphical view of sequences


% --- Executes on button press in pushmendeldataview.
function pushmendeldataview_Callback(hObject, eventdata, handles)
% hObject    handle to pushmendeldataview (see GCBO)
```

```
% eventdata  reserved - to be defined in a future version of MATLAB
% handles    structure with handles and user data (see GUIDATA)

experiment=get(handles.popusedataNum,'Value'); %extracting selected
value from object

switch experiment
    case 1 % Value used when "No" is select, e.g. model sampled dataset
        view_sampleddata;
        % helpdlg('Select Mendel data first','Dataset Selection');
    otherwise
        view_mendeldata(experiment-1); %display graphical view of mendel
exp. 1..8
end %switch


% --- Executes on mouse press over figure background.
function figure1_ButtonDownFcn(hObject, eventdata, handles)
% hObject    handle to figure1 (see GCBO)
% eventdata  reserved - to be defined in a future version of MATLAB
% handles    structure with handles and user data (see GUIDATA)


% --- Executes during object creation, after setting all properties.
function popgenesNum_CreateFcn(hObject, eventdata, handles)
% hObject    handle to popgenesNum (see GCBO)
% eventdata  reserved - to be defined in a future version of MATLAB
% handles    empty - handles not created until after all CreateFcns
called

% Hint: popupmenu controls usually have a white background on Windows.
%       See ISPC and COMPUTER.
if ispc
    set(hObject,'BackgroundColor','white');
else

set(hObject,'BackgroundColor',get(0,'defaultUicontrolBackgroundColor'));
end

% --- Executes on selection change in popgenesNum.
function popgenesNum_Callback(hObject, eventdata, handles)
% hObject    handle to popgenesNum (see GCBO)
% eventdata  reserved - to be defined in a future version of MATLAB
% handles    structure with handles and user data (see GUIDATA)

% Hints: contents = get(hObject,'String') returns popgenesNum contents
as cell array
%        contents{get(hObject,'Value')} returns selected item from
popgenesNum

updateGUI(handles); % The GUI and the basic matrices in the HMM is
uppdated
set(handles.popusedataNum,'Value',1) %present 'No' instead of MendelData
at GUI


% --- Executes during object creation, after setting all properties.
function poppploidyNum_CreateFcn(hObject, eventdata, handles)
```

```
% hObject    handle to poppl oidyNum (see GCBO)
% eventdata  reserved - to be defined in a future version of MATLAB
% handles    empty - handles not created until after all CreateFcns
called

% Hint: popupmenu controls usually have a white background on Windows.
%       See ISPC and COMPUTER.
if ispc
    set(hObject,'BackgroundColor','white');
else

set(hObject,'BackgroundColor',get(0,'defaultUicontrolBackgroundColor'));
end


% --- Executes on selection change in poppl oidyNum.
function poppl oidyNum_Callback(hObject, eventdata, handles)
% hObject    handle to poppl oidyNum (see GCBO)
% eventdata  reserved - to be defined in a future version of MATLAB
% handles    structure with handles and user data (see GUIDATA)

updateGUI(handles); % The GUI and the basic matrices in the HMM is
uppdated

% --------------------------------------------------------------------
function textpenotypesNum_Callback(hObject, eventdata, handles)
% hObject    handle to textpenotypesNum (see GCBO)
% eventdata  reserved - to be defined in a future version of MATLAB
% handles    structure with handles and user data (see GUIDATA)


% --- Executes during object creation, after setting all properties.
function popusedataNum_CreateFcn(hObject, eventdata, handles)
% hObject    handle to popusedataNum (see GCBO)
% eventdata  reserved - to be defined in a future version of MATLAB
% handles    empty - handles not created until after all CreateFcns
called

% Hint: popupmenu controls usually have a white background on Windows.
%       See ISPC and COMPUTER.
if ispc
    set(hObject,'BackgroundColor','white');
else

set(hObject,'BackgroundColor',get(0,'defaultUicontrolBackgroundColor'));
end


% --- Executes on selection change in popusedataNum.
function popusedataNum_Callback(hObject, eventdata, handles)
% hObject    handle to popusedataNum (see GCBO)
% eventdata  reserved - to be defined in a future version of MATLAB
% handles    structure with handles and user data (see GUIDATA)

% Hints: contents = get(hObject,'String') returns popusedataNum contents
as cell array
%        contents{get(hObject,'Value')} returns selected item from
popusedataNum
```

```
global dataE;

experiment=get(hObject,'Value'); %extracting selected value from object
if experiment >= 2 & experiment <= 10
    % clear dataE, hidX; % prepare to read a new data set
    dataE=resample_mendeldata(experiment-1) % No experiment has Value
number 1
    [n,m]= size(dataE);
    set(handles.editnumNum,'string',n); % new samplenumber presented at
GUI
    set(handles.editlengthNum,'string',m); % new sample length presented
at GUI
    switch experiment
        case 9 %This case is experiment 8 with 2 genes
            set(handles.popgenesNum,'Value',2); % new gene value
presented at GUI
            updateGUI(handles); % The GUI and the basic matrices in the
HMM is uppdated
        case 10 %This case is experiment 9 with 3 genes
            set(handles.popgenesNum,'Value',3); % new gene value
presented at GUI
            updateGUI(handles); % The GUI and the basic matrices in the
HMM is uppdated
        otherwise %This case is experiment 1-7 (1 gene)
            set(handles.popgenesNum,'Value',1); % new gene value
presented at GUI
            updateGUI(handles); % The GUI and the basic matrices in the
HMM is uppdated
    end %switch
end %if

% --- Executes during object creation, after setting all properties.
function editnumNum_CreateFcn(hObject, eventdata, handles)
% hObject    handle to editnumNum (see GCBO)
% eventdata  reserved - to be defined in a future version of MATLAB
% handles    empty - handles not created until after all CreateFcns
called

% Hint: edit controls usually have a white background on Windows.
%       See ISPC and COMPUTER.
if ispc
    set(hObject,'BackgroundColor','white');
else

set(hObject,'BackgroundColor',get(0,'defaultUicontrolBackgroundColor'));
end

function editnumNum_Callback(hObject, eventdata, handles)
% hObject    handle to editnumNum (see GCBO)
% eventdata  reserved - to be defined in a future version of MATLAB
% handles    structure with handles and user data (see GUIDATA)

% Hints: get(hObject,'String') returns contents of editnumNum as text
%        str2double(get(hObject,'String')) returns contents of
editnumNum as a double
updateGUI(handles); % The GUI and the basic matrices in the HMM is
uppdated
set(handles.popusedataNum,'Value',1) %present 'No' instead of MendelData
at GUI
```

```
% --- Executes during object creation, after setting all properties.
function editlengthNum_CreateFcn(hObject, eventdata, handles)
% hObject    handle to editlengthNum (see GCBO)
% eventdata  reserved - to be defined in a future version of MATLAB
% handles    empty - handles not created until after all CreateFcns
called

% Hint: edit controls usually have a white background on Windows.
%       See ISPC and COMPUTER.
if ispc
    set(hObject,'BackgroundColor','white');
else

set(hObject,'BackgroundColor',get(0,'defaultUicontrolBackgroundColor'));
end


function editlengthNum_Callback(hObject, eventdata, handles)
% hObject    handle to editlengthNum (see GCBO)
% eventdata  reserved - to be defined in a future version of MATLAB
% handles    structure with handles and user data (see GUIDATA)

% Hints: get(hObject,'String') returns contents of editlengthNum as text
%        str2double(get(hObject,'String')) returns contents of
editlengthNum as a double

updateGUI(handles); % The GUI and the basic matrices in the HMM is
uppdated
set(handles.popusedataNum,'Value',1) %present 'No' instead of MendelData
at GUI


% --- Executes during object creation, after setting all properties.
function popinitializationNum_CreateFcn(hObject, eventdata, handles)
% hObject    handle to popinitializationNum (see GCBO)
% eventdata  reserved - to be defined in a future version of MATLAB
% handles    empty - handles not created until after all CreateFcns
called

% Hint: popupmenu controls usually have a white background on Windows.
%       See ISPC and COMPUTER.
if ispc
    set(hObject,'BackgroundColor','white');
else

set(hObject,'BackgroundColor',get(0,'defaultUicontrolBackgroundColor'));
end

 % --- Executes on selection change in popinitializationNum.
function popinitializationNum_Callback(hObject, eventdata, handles)
% hObject    handle to popinitializationNum (see GCBO)
% eventdata  reserved - to be defined in a future version of MATLAB
% handles    structure with handles and user data (see GUIDATA)

% Hints: contents = get(hObject,'String') returns popinitializationNum
contents as cell array
```

```
%        contents{get(hObject,'Value')} returns selected item from
popinitializationNum

method=get(hObject,'Value'); %extracting selected value from object
runinitialization(method, handles);

% --------------------------------
%function runinitialization(method, handles)
% User defined function to initialize the parameters.

E=str2num(get(handles.textpenotypesNum,'String')); % extracting number
of observables
X=str2num(get(handles.textstatesNum,'String')); % extracting number of
states
% genesNo=get(handles.popgenesNum,'Value'); %first extracting number of
genes
% initbest_ll=str2num(get(handles.textbestlikelihoodNum,'String')); %
extracting best loglik

% Now we can initialize our model:
switch method
    case 1 % dominant diagonal method
        dominance=0.67 % ------ may be changed here!!!!
        loglik_init = init_diagonal (X, E, dominance)
    case 2 % random parameter method
        loglik_init = init_random (X, E)
end %switch
set(handles.textinitialbestlikelihoodNum,'string',loglik_init) % update
initial likelihood presented at GUI

% ---------------------------------
function updateGUI(handles)
% User defined function executed when a new dataset is selected for use.
% The number of phenotypes and states is derived from the selections and
% uppdated on the GUI. The basic matrices in the HMM is also adjusted.

genes=get(handles.popgenesNum,'Value'); %extracting number of genes
val=get(handles.popploidyNum,'Value'); %extracting ploidy selected value
from object
% (for one gene the real ploidy level, p, will be p=2*val)
% The number of genotypes will be p + 1. (combinatorics: choosing p out
of 2
% with replacement and disregarding order)
% The total number of states = number of genotypes ^ number of genes:
% states=(2*val +1)^genes
% We then uppdate the number of states and of phenotypes:
set(handles.textstatesNum,'string',(2*val+1)^genes) % updated states
presented at GUI
set(handles.textpenotypesNum,'string',2^genes) % updated phenotypes
presented at GUI

% We adjust the previous likelihoods:
set(handles.textinitialbestlikelihoodNum,'string',-Inf) % updated
likelihood presented at GUI
set(handles.textlikelihoodNum,'string',-Inf) % updated at GUI
set(handles.textbestlikelihoodNum,'string',-Inf) % update at GUI

% We must also adjust the basic matrices in the HMM model:
switch val
```

```
    case 1
        set_diploidmode
    case 2
        set_tetraploidmode
end %switch
```

**<fwdback.m>**

```
function [alpha, beta, gamma, loglik, xi, gamma2] =
fwdback(init_state_distrib, transmat, obslik, ...
                            varargin)
% FWDBACK Compute the posterior probs. in an HMM using the forwards
backwards algo.
%
% [alpha, beta, gamma, loglik, xi, gamma2] = fwdback(init_state_distrib,
transmat, obslik, ...)
%
% From the MIT-toolbox by Kevin Murphy, 2003.
%
% Notation:
% Y(t) = observation, Q(t) = hidden state, M(t) = mixture variable (for
MOG outputs)
% A(t) = discrete input (action) (for POMDP models)
%
% INPUT:
% init_state_distrib(i) = Pr(Q(1) = i)
% transmat(i,j) = Pr(Q(t) = j | Q(t-1)=i)
%  or transmat{a}(i,j) = Pr(Q(t) = j | Q(t-1)=i, A(t-1)=a) if there are
discrete inputs
% obslik(i,t) = Pr(Y(t)| Q(t)=i)
%   (Compute obslik using eval_pdf_xxx on your data sequence first.)
%
% Optional parameters may be passed as 'param_name', param_value pairs.
% Parameter names are shown below; default values in [] - if none,
argument is mandatory.
%
% For HMMs with MOG outputs: if you want to compute gamma2, you must
specify
% 'obslik2' - obslik(i,j,t) = Pr(Y(t)| Q(t)=i,M(t)=j)  []
% 'mixmat' - mixmat(i,j) = Pr(M(t) = j | Q(t)=i)  []
%
% For HMMs with discrete inputs:
% 'act' - act(t) = action performed at step t
%
% Optional arguments:
% 'fwd_only' - if 1, only do a forwards pass and set beta=[], gamma2=[]
[0]
% 'scaled' - if 1,  normalize alphas and betas to prevent underflow [1]
% 'maximize' - if 1, use max-product instead of sum-product [0]
%
% OUTPUTS:
% alpha(i,t) = p(Q(t)=i | y(1:t)) (or p(Q(t)=i, y(1:t)) if scaled=0)
% beta(i,t) = p(y(t+1:T) | Q(t)=i)*p(y(t+1:T)|y(1:t)) (or p(y(t+1:T) |
Q(t)=i) if scaled=0)
% gamma(i,t) = p(Q(t)=i | y(1:T))
% loglik = log p(y(1:T))
% xi(i,j,t-1)  = p(Q(t-1)=i, Q(t)=j | y(1:T))
% gamma2(j,k,t) = p(Q(t)=j, M(t)=k | y(1:T)) (only for MOG  outputs)
%
% If fwd_only = 1, these become
```

```
% alpha(i,t) = p(Q(t)=i | y(1:t))
% beta = []
% gamma(i,t) = p(Q(t)=i | y(1:t))
% xi(i,j,t-1)  = p(Q(t-1)=i, Q(t)=j | y(1:t))
% gamma2 = []
%
% Note: we only compute xi if it is requested as a return argument,
since it can be very large.
% Similarly, we only compute gamma2 on request (and if using MOG
outputs).
%
% Examples:
%
% [alpha, beta, gamma, loglik] = fwdback(pi, A,
eval_pdf_cond_multinomial(sequence, B));
%
% [B, B2] = eval_pdf_cond_mixgauss(data, mu, Sigma, mixmat);
% [alpha, beta, gamma, loglik, xi, gamma2] = fwdback(pi, A, B,
'obslik2', B2, 'mixmat', mixmat);


if nargout >= 5, compute_xi = 1; else compute_xi = 0; end
if nargout >= 6, compute_gamma2 = 1; else compute_gamma2 = 0; end

[obslik2, mixmat, fwd_only, scaled, act, maximize, compute_xi,
compute_gamma2] = ...
    process_options(varargin, ...
    'obslik2', [], 'mixmat', [], ...
    'fwd_only', 0, 'scaled', 1, 'act', [], 'maximize', 0, ...
            'compute_xi', compute_xi, 'compute_gamma2', compute_gamma2);


[Q T] = size(obslik);

if isempty(obslik2)
  compute_gamma2 = 0;
end

if isempty(act)
  act = ones(1,T);
  transmat = { transmat } ;
end

scale = ones(1,T);

% scale(t) = Pr(O(t) | O(1:t-1)) = 1/c(t) as defined by Rabiner (1989).
% Hence prod_t scale(t) = Pr(O(1)) Pr(O(2)|O(1)) Pr(O(3) | O(1:2)) ... =
Pr(O(1), ... ,O(T))
% or log P = sum_t log scale(t).
% Rabiner suggests multiplying beta(t) by scale(t), but we can instead
% normalise beta(t) - the constants will cancel when we compute gamma.

loglik = 0;

alpha = zeros(Q,T);
gamma = zeros(Q,T);
if compute_xi
  xi = zeros(Q,Q,T-1);
else
  xi = [];
end
```

```
%%%%%%%%% Forwards %%%%%%%%%%

t = 1;
alpha(:,1) = init_state_distrib(:) .* obslik(:,t);
if scaled
  [alpha(:,t), scale(t)] = normalise(alpha(:,t));
end
for t=2:T
  %trans = transmat(:,:,act(t-1))';
  trans = transmat{act(t-1)};
  if maximize
    m = max_mult(trans', alpha(:,t-1));
    %A = repmat(alpha(:,t-1), [1 Q]);
    %m = max(trans .* A, [], 1);
  else
    m = trans' * alpha(:,t-1);
  end
  alpha(:,t) = m(:) .* obslik(:,t);
  if scaled
    [alpha(:,t), scale(t)] = normalise(alpha(:,t));
  end
  if compute_xi & fwd_only  % useful for online EM
    xi(:,:,t-1) = normalise((alpha(:,t-1) * obslik(:,t)') .* trans);
  end
end
if scaled
  if any(scale)==0 % or 1?
    loglik = -inf;
  else
    loglik = sum(log(scale));
  end
else
  loglik = log(sum(alpha(:,T)));
end

if fwd_only
  gamma = alpha;
  beta = [];
  gamma2 = [];
  return;
end


%%%%%%%%% Backwards %%%%%%%%%%

beta = zeros(Q,T);
if compute_gamma2
  M = size(mixmat, 2);
  gamma2 = zeros(Q,M,T);
else
  gamma2 = [];
end

beta(:,T) = ones(Q,1);
gamma(:,T) = normalise(alpha(:,T) .* beta(:,T));
t=T;
if compute_gamma2
  denom = obslik(:,t) + (obslik(:,t)==0); % replace 0s with 1s before
dividing
```

```
  gamma2(:,:,t) = obslik2(:,:,t) .* mixmat .* repmat(gamma(:,t), [1 M])
./ repmat(denom, [1 M]);
  %gamma2(:,:,t) = normalise(obslik2(:,:,t) .* mixmat .*
repmat(gamma(:,t), [1 M])); % wrong!
end
for t=T-1:-1:1
  b = beta(:,t+1) .* obslik(:,t+1);
  %trans = transmat(:,:,act(t));
  trans = transmat{act(t)};
  if maximize
    B = repmat(b(:)', Q, 1);
    beta(:,t) = max(trans .* B, [], 2);
  else
    beta(:,t) = trans * b;
  end
  if scaled
    beta(:,t) = normalise(beta(:,t));
  end
  gamma(:,t) = normalise(alpha(:,t) .* beta(:,t));
  if compute_xi
    xi(:,:,t) = normalise((trans .* (alpha(:,t) * b')));
  end
  if compute_gamma2
    denom = obslik(:,t) + (obslik(:,t)==0); % replace 0s with 1s before
dividing
    gamma2(:,:,t) = obslik2(:,:,t) .* mixmat .* repmat(gamma(:,t), [1
M]) ./ repmat(denom, [1 M]);
    %gamma2(:,:,t) = normalise(obslik2(:,:,t) .* mixmat .*
repmat(gamma(:,t), [1 M]));
  end
end


% We now explain the equation for gamma2
% Let zt=y(1:t-1,t+1:T) be all observations except y(t)
% gamma2(Q,M,t) = P(Qt,Mt|yt,zt) = P(yt|Qt,Mt,zt) P(Qt,Mt|zt) / P(yt|zt)
%               = P(yt|Qt,Mt) P(Mt|Qt) P(Qt|zt) / P(yt|zt)
% Now gamma(Q,t) = P(Qt|yt,zt) = P(yt|Qt) P(Qt|zt) / P(yt|zt)
% hence
% P(Qt,Mt|yt,zt) = P(yt|Qt,Mt) P(Mt|Qt) [P(Qt|yt,zt) P(yt|zt) /
P(yt|Qt)] / P(yt|zt)
%               = P(yt|Qt,Mt) P(Mt|Qt) P(Qt|yt,zt) / P(yt|Qt)
%
```

## <gen_HMM_em>

```
function [LL, prior, transmat, obsmat] = gen_HMM_em(data, prior,
transmat, obsmat, handles, varargin)
% gen_HMM Find the ML/MAP parameters of an HMM with discrete outputs
using EM.
% [ll_trace, prior, transmat, obsmat] = learn_dhmm(data, prior0,
transmat0, obsmat0, ...)
%
% From the MIT-toolbox by Kevin Murphy, 1998-2003.
%
% Notation: Q(t) = hidden state, Y(t) = observation
%
% INPUTS:
% data{ex} or data(ex,:) if all sequences have the same length
```

```
% prior(i)
% transmat(i,j)
% obsmat(i,o)
%
% Optional parameters may be passed as 'param_name', param_value pairs.
% Parameter names are shown below; default values in [] - if none,
argument is mandatory.
%
% 'max_iter' - max number of EM iterations [10]
% 'thresh' - convergence threshold [1e-4]
% 'verbose' - if 1, print out loglik at every iteration [1]
% 'obs_prior_weight' - weight to apply to uniform dirichlet prior on
observation matrix [0]
%
% To clamp some of the parameters, so learning does not change them:
% 'adj_prior' - if 0, do not change prior [1]
% 'adj_trans' - if 0, do not change transmat [1]
% 'adj_obs' - if 0, do not change obsmat [1]


[max_iter, thresh, verbose, obs_prior_weight, adj_prior, adj_trans,
adj_obs] = ...
    process_options(varargin, 'max_iter', 10, 'thresh', 1e-4, 'verbose',
1, ...
            'obs_prior_weight', 0, 'adj_prior', 1, 'adj_trans', 1,
'adj_obs', 1);

previous_loglik = -inf;
loglik = 0;
converged = 0;
num_iter = 1;
LL = [];

if ~iscell(data)
  data = num2cell(data, 2); % each row gets its own cell
end

while (num_iter <= max_iter) & ~converged
  % E step
  [loglik, exp_num_trans, exp_num_visits1, exp_num_emit] = ...
      compute_ess_dhmm(prior, transmat, obsmat, data, obs_prior_weight);

  % M step
  if adj_prior
    prior = normalise(exp_num_visits1);
  end
  if adj_trans & ~isempty(exp_num_trans)
    transmat = mk_stochastic(exp_num_trans);
  end
  if adj_obs
    obsmat = mk_stochastic(exp_num_emit);
  end

  if verbose, fprintf(1, 'iteration %d, loglik = %f\n', num_iter,
loglik); end
  set(handles.textlikelihoodNum,'string',loglik); % updated likelihood
presented at GUI
  num_iter =  num_iter + 1;
  converged = em_converged(loglik, previous_loglik, thresh);
  previous_loglik = loglik;
  LL = [LL loglik];
```

```
end

%%%%%%%%%%%%%%%%%%%%%%%

function [loglik, exp_num_trans, exp_num_visits1, exp_num_emit,
exp_num_visitsT] = ...
    compute_ess_dhmm(startprob, transmat, obsmat, data, dirichlet)
% COMPUTE_ESS_DHMM Compute the Expected Sufficient Statistics for an HMM
with discrete outputs
% function [loglik, exp_num_trans, exp_num_visits1, exp_num_emit,
exp_num_visitsT] = ...
%    compute_ess_dhmm(startprob, transmat, obsmat, data, dirichlet)
%
% INPUTS:
% startprob(i)
% transmat(i,j)
% obsmat(i,o)
% data{seq}(t)
% dirichlet - weighting term for uniform dirichlet prior on expected
emissions
%
% OUTPUTS:
% exp_num_trans(i,j) = sum_l sum_{t=2}^T Pr(X(t-1) = i, X(t) = j|
Obs(l))
% exp_num_visits1(i) = sum_l Pr(X(1)=i | Obs(l))
% exp_num_visitsT(i) = sum_l Pr(X(T)=i | Obs(l))
% exp_num_emit(i,o) = sum_l sum_{t=1}^T Pr(X(t) = i, O(t)=o| Obs(l))
% where Obs(l) = O_1 .. O_T for sequence l.


numex = length(data);
[S O] = size(obsmat);
exp_num_trans = zeros(S,S);
exp_num_visits1 = zeros(S,1);
exp_num_visitsT = zeros(S,1);
exp_num_emit = dirichlet*ones(S,O);
loglik = 0;

for ex=1:numex
  obs = data{ex};
  T = length(obs);
  %obslik = eval_pdf_cond_multinomial(obs, obsmat);
  obslik = multinomial_prob(obs, obsmat);
  [alpha, beta, gamma, current_ll, xi] = fwdback(startprob, transmat,
obslik);

  loglik = loglik +  current_ll;
  exp_num_trans = exp_num_trans + sum(xi,3);
  exp_num_visits1 = exp_num_visits1 + gamma(:,1);
  exp_num_visitsT = exp_num_visitsT + gamma(:,T);
  % loop over whichever is shorter
  if T < O
    for t=1:T
      o = obs(t);
      exp_num_emit(:,o) = exp_num_emit(:,o) + gamma(:,t);
    end
  else
    for o=1:O
      ndx = find(obs==o);
      if ~isempty(ndx)
    exp_num_emit(:,o) = exp_num_emit(:,o) + sum(gamma(:, ndx), 2);
```

```
      end
    end
  end
end
```

**<sample_HMM.m>**

```
function [hiddenX, obsE] = sample_HMM(initial_prob, transmatP, obsmatE,
T, numbex)
% Sampling of data for a given HMM structure.

% sample_HMM produces a realization of a Hidden Markov Model with
transition probability
% matrix transmatP, and observation emission matrix obsmatE.
% The vector initial_prob is used as the initial state, and there are
% generated numbex sequenses both of the hidden chain and the related
oberved sequence.
% Each of the numbex row of hiddenX and obsE is of length T.

% From the mendelHMM-toolbox by Steinar Thorvaldsen, 2004.
% http://www.math.uit.no/bi/hmm/

% [hiddenX, obsE] = sample_HMM(initial_prob, transmatP, obsmatE, T,
numbex)

% Outputs: hiddenX  =   Matrix of sequence of dim (mumbex x T) of
numbers between 1 and n,
%                       where n is the number of states or the dimension
of
%                       transmatP
%          obsE =       Matrix of sequences of dim (numbex x T) of
numbers
%                       between 1 and m, where m is given by the dim of
obsmatE.

% Inputs:  initial_prob = Initial state vector.
%          transmatP =    Transition probability matrix
%          obsmatE =      Observation emission matrix
%          T  =           The length of time series
%          numbex =       The number of chains
%

hiddenX = sample_MC(initial_prob, transmatP, T, numbex);
obsE = zeros(numbex, T);

% We may now use the general function  sample_discr  to compute the
related
% observation for each state:
for i=1:numbex
  h = hiddenX(i,1);
  obsE(i,1) = sample_discr(obsmatE(h,:));
  for j=2:T
    h = hiddenX(i,j);
    obsE(i,j) = sample_discr(obsmatE(h,:));
  end
end
```

## 6.4 OUTPUT

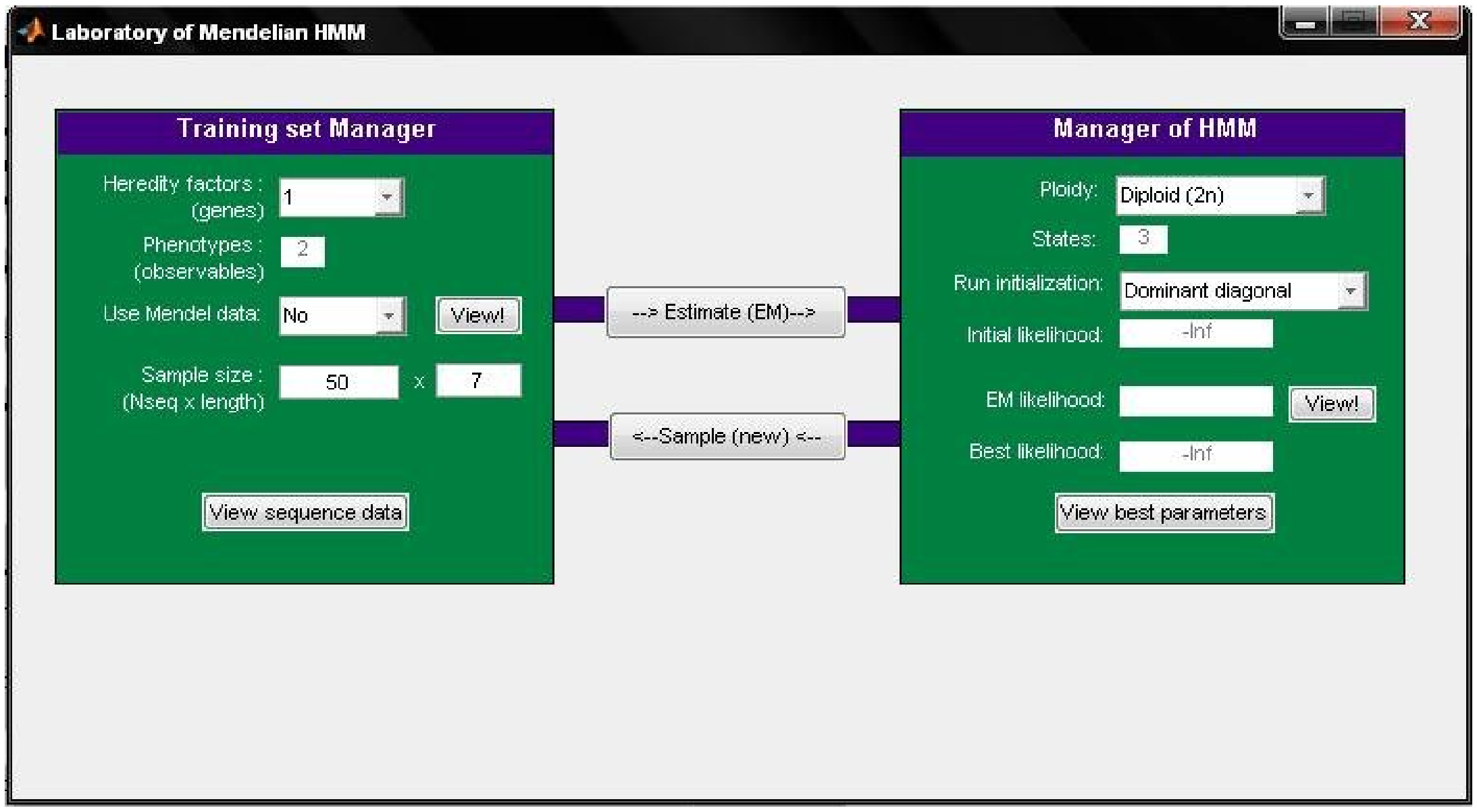

**HOME page mendelHMM**

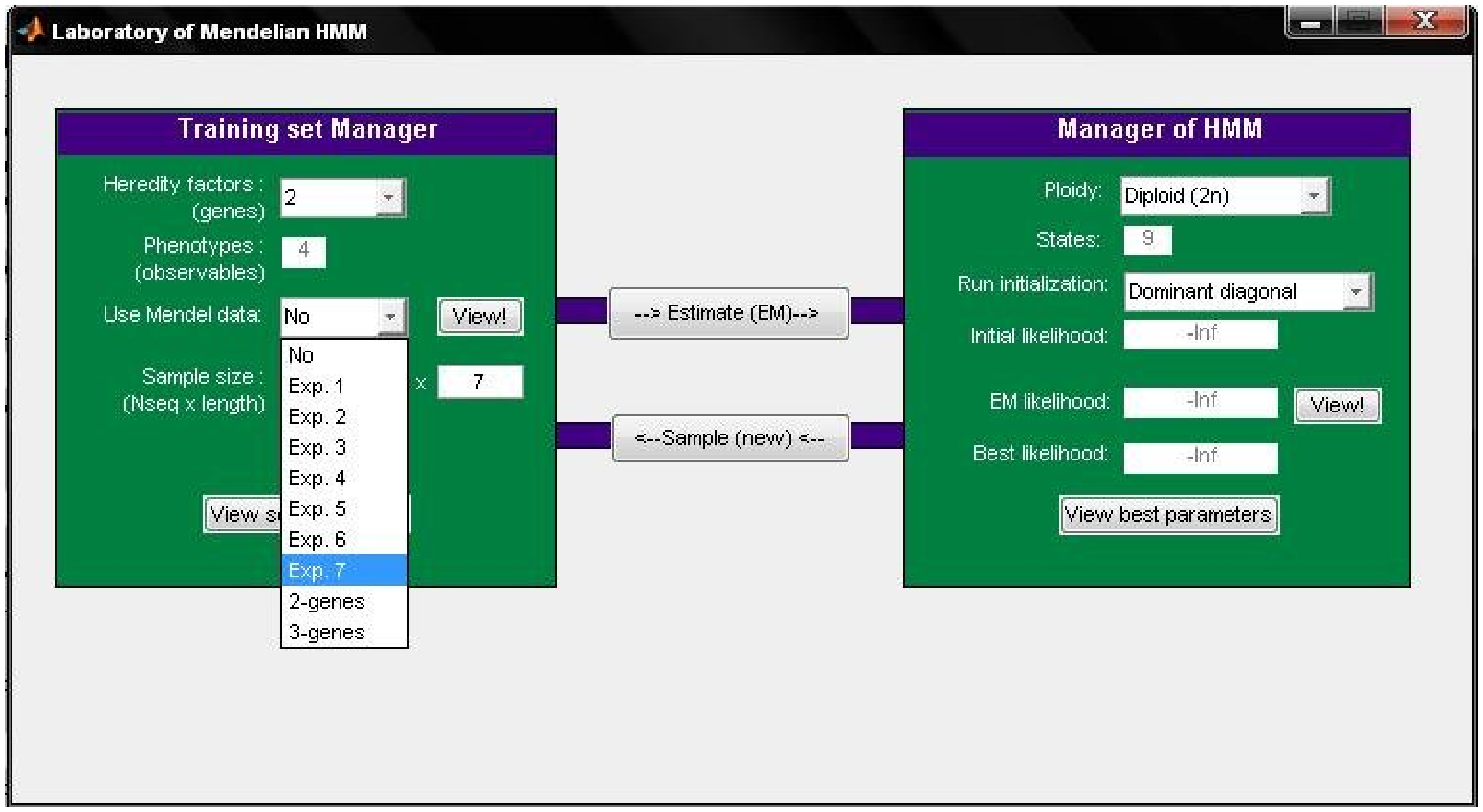

**Selection of Data Set**

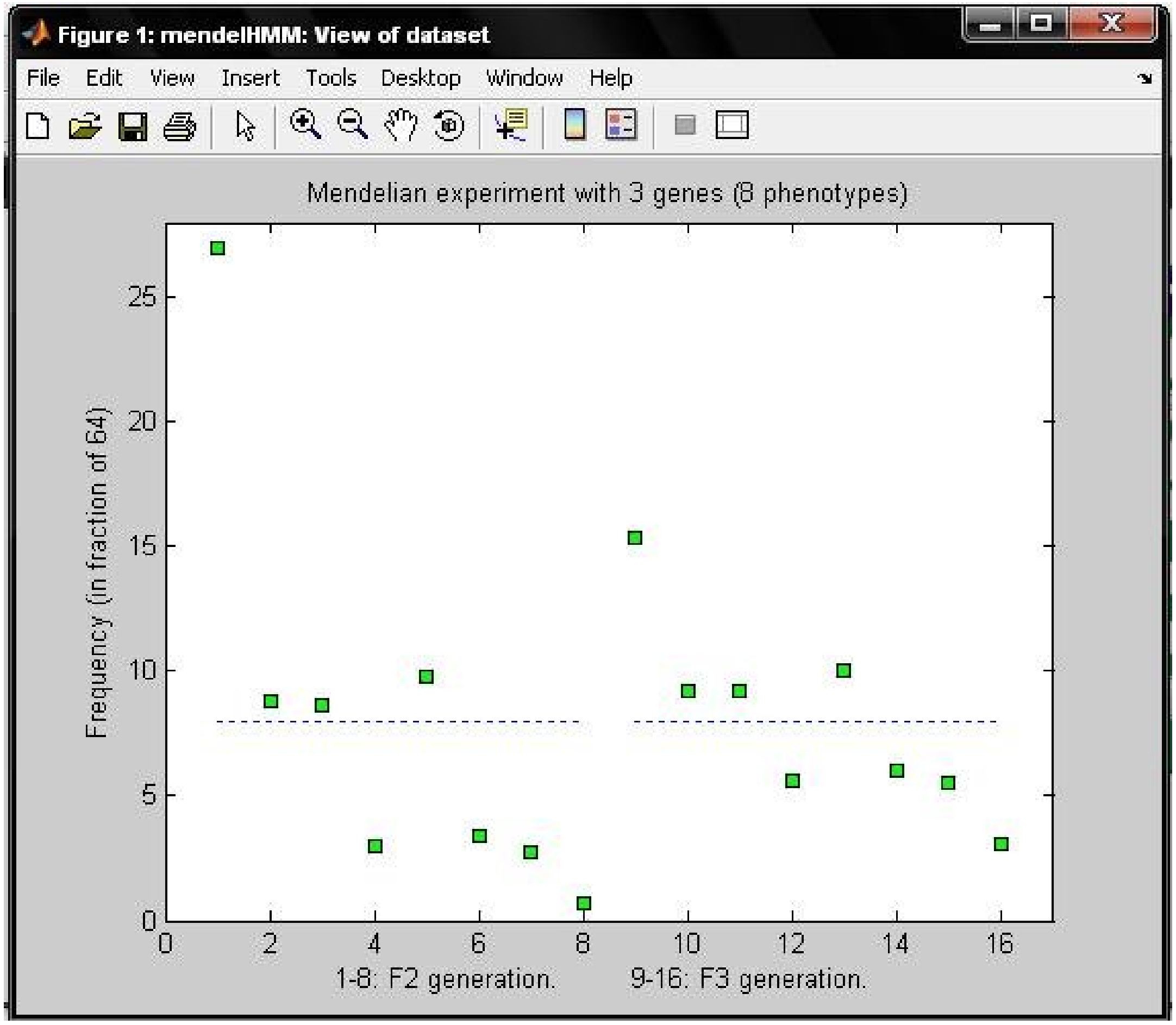

**View of Data Set**

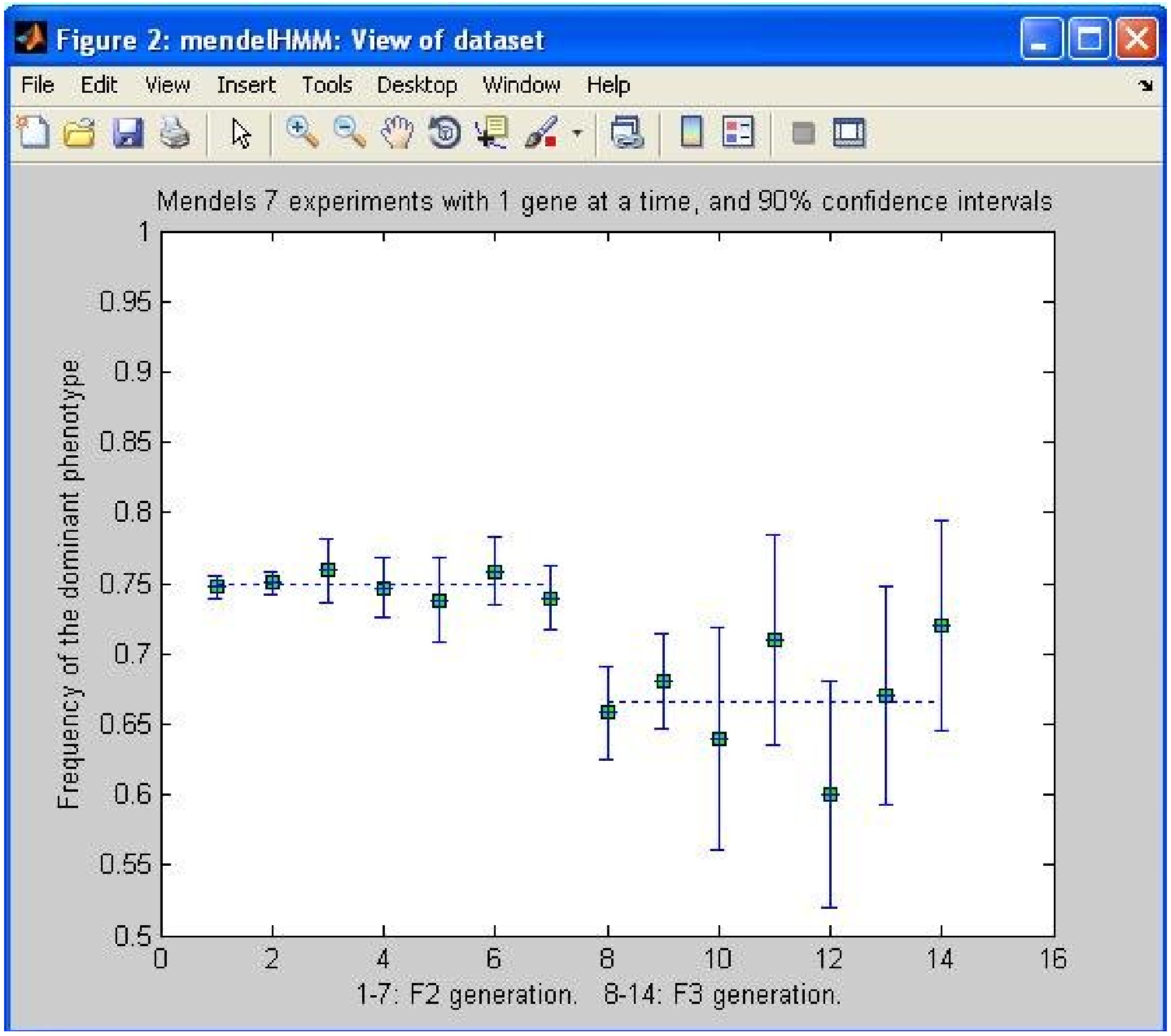

**View of Data Set with confidence intervals**

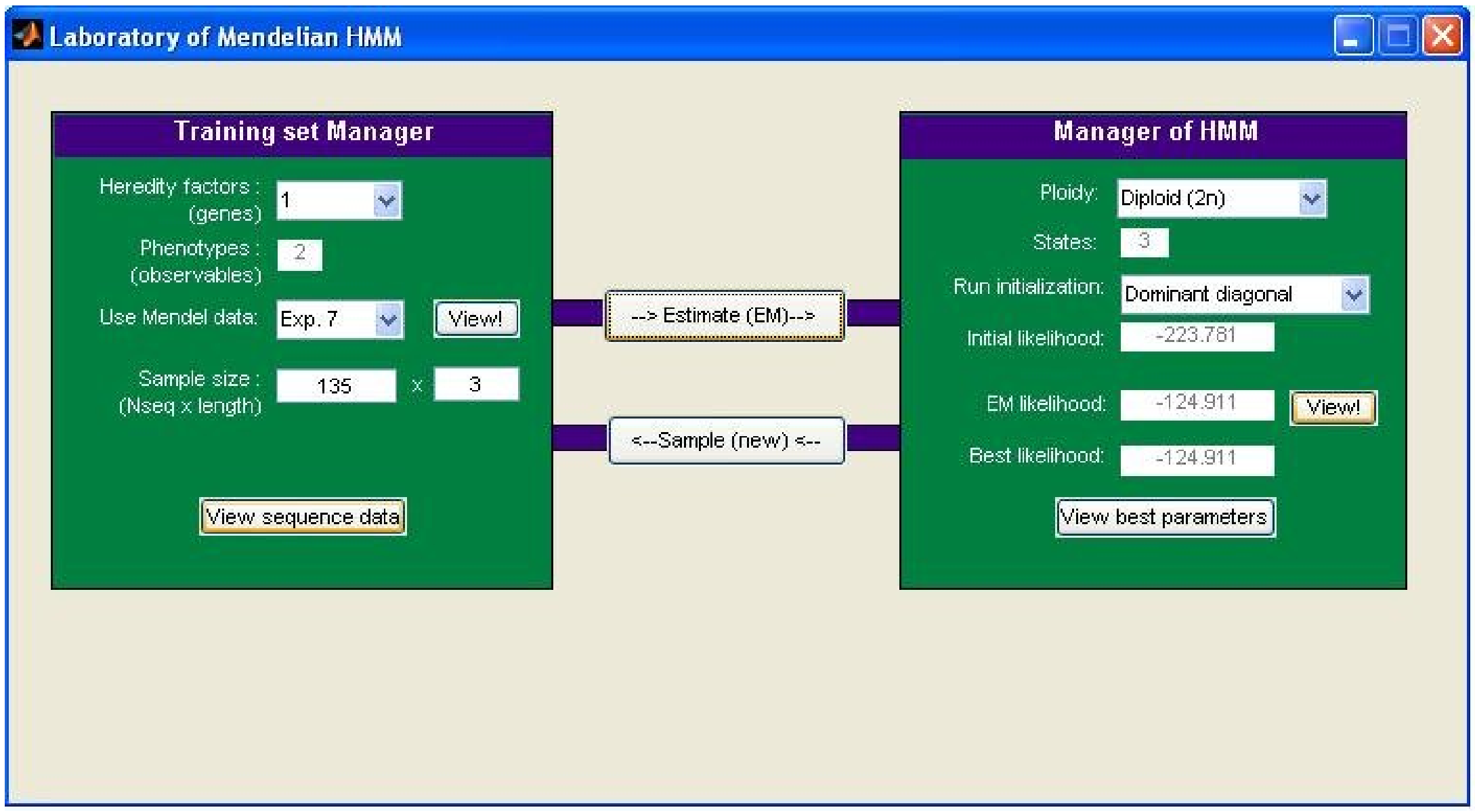

**Estimation**

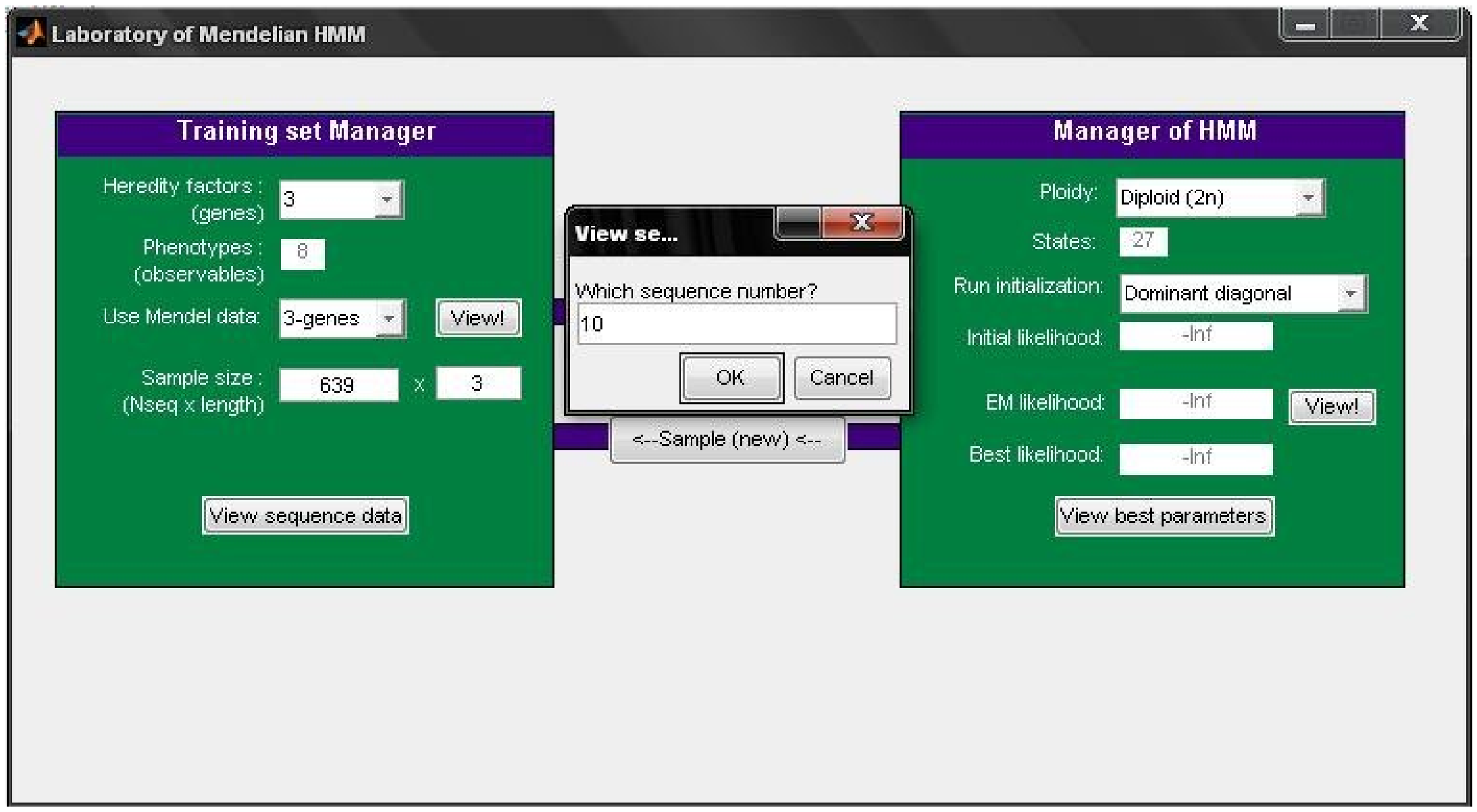

**Selection of Sequence Number**

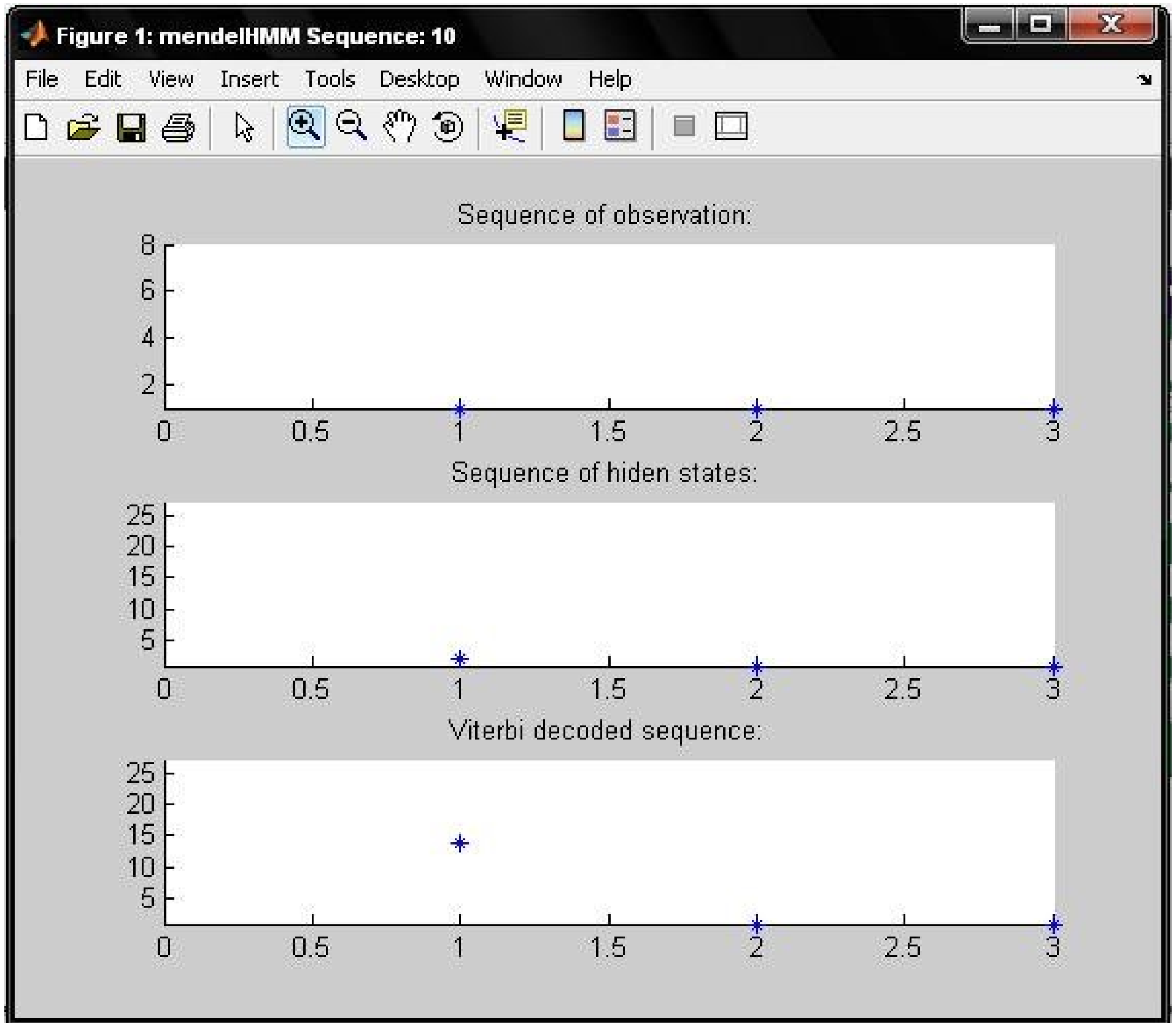

**Sequence of Data Set**

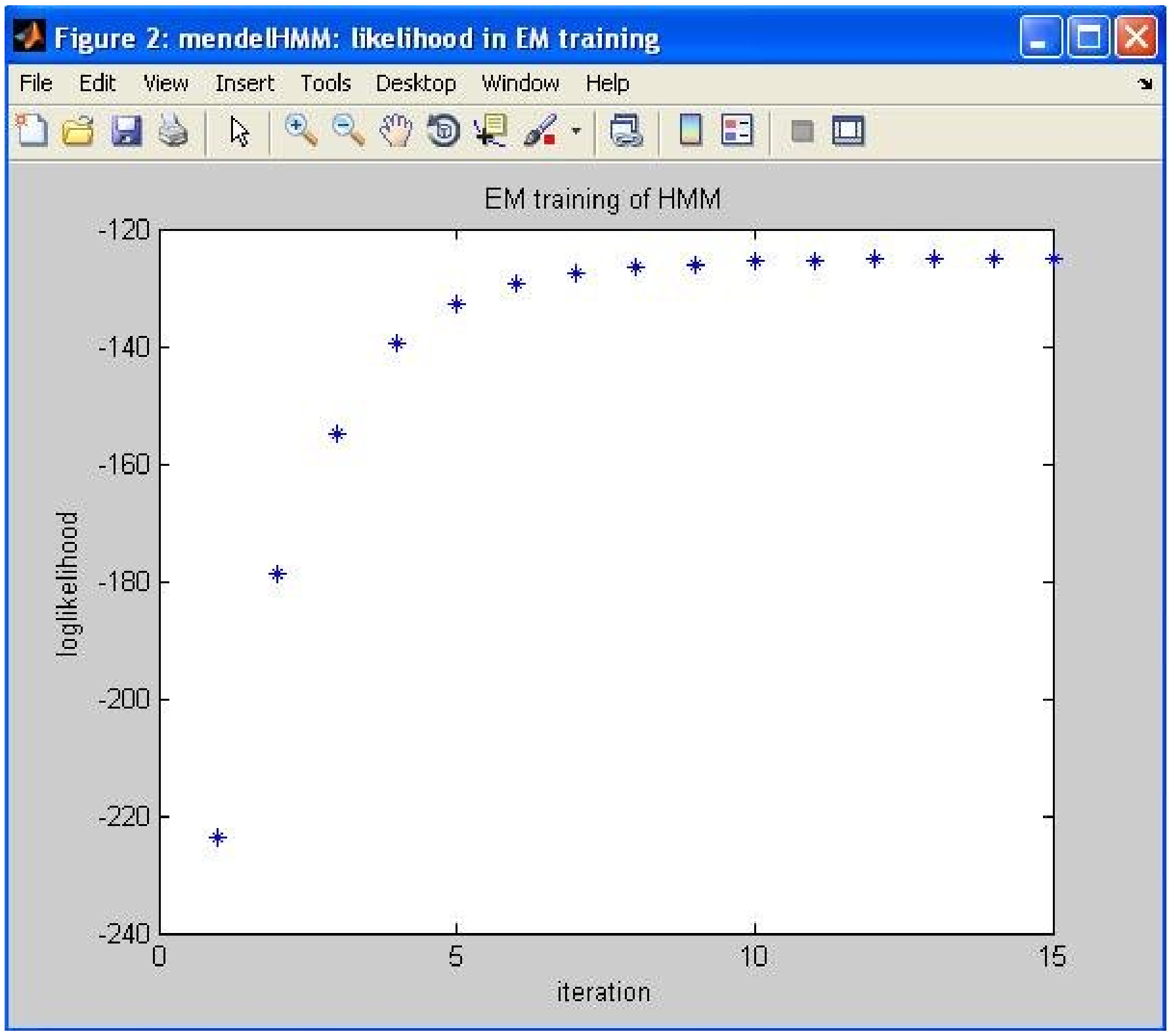

**EM Likelihood**

## 6.5 GANTT CHART

Gantt Charts are mainly used to allocate resources to activities. The resources allocted to activities, include staff, hardware, software.

Gantt Chart is a special type of bar chart where each bar represents an activity. The bars are drawn along time line. The length of the each bar is proportional to the duration of the time plan for the corresponding activity.

Gantt Charts used in s/w project mgmt. are exactly an enhanced version of the standard Gantt Charts.

Gantt Charts used for the s/w project mgmt., each bar consists a white part & a shaded part. The shaded part of the bar shows the length of the time each task is estimated to take. The white part shows the slck time i.e. the latest time by which a task must be completed.

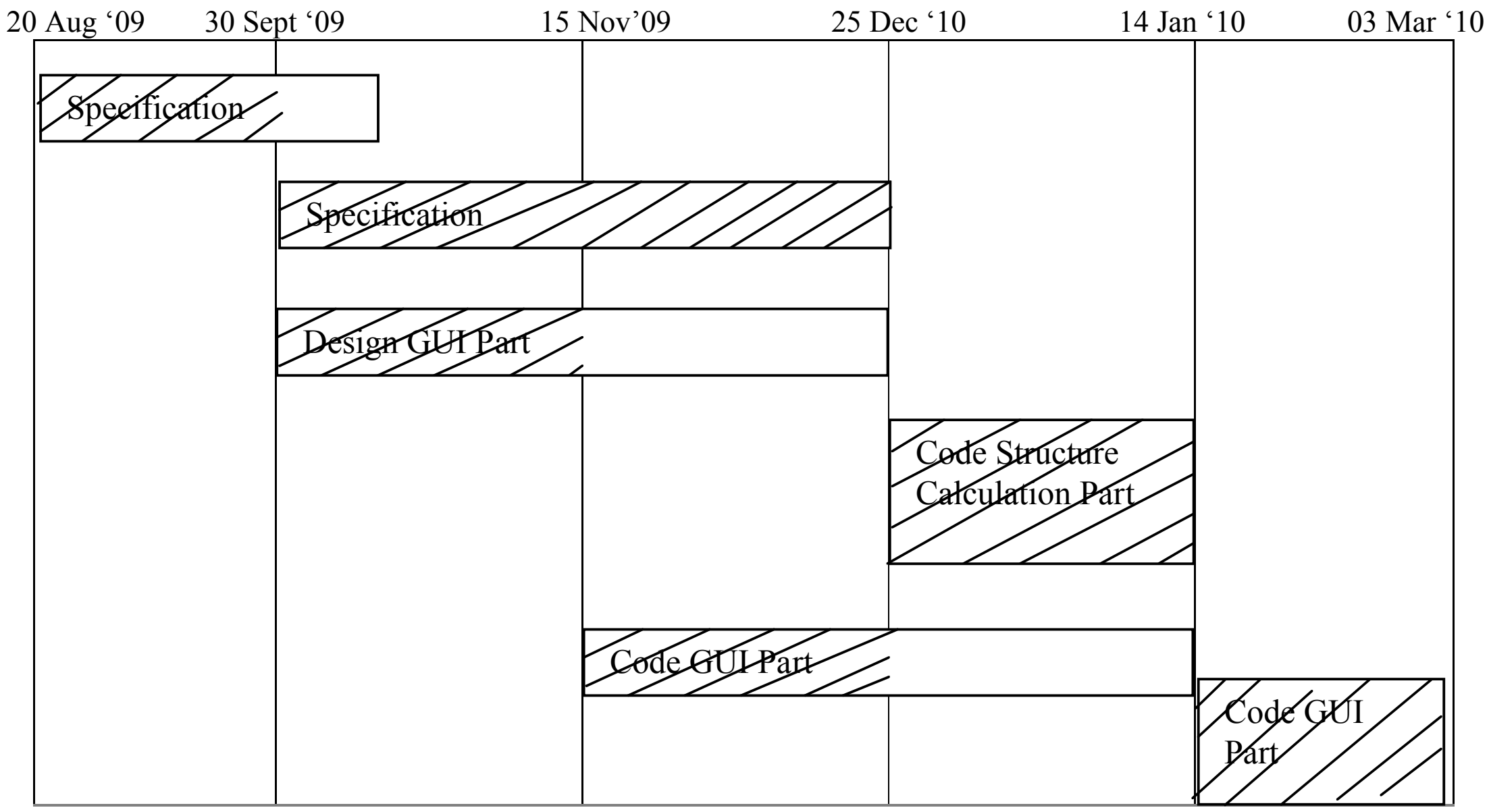

# TESTING & RESULT

## 7.1 TEST CASES

**Hidden Markov Model**

**Model-1: Prediction of protein secondary structures from protein sequences.**
**Observed state**: Protein sequence
**Hidden state**: Protein secondary structure

Table-1: Efficiencies of HMM model 1 for different training and test dataset

| Training dataset index | Test dataset index | Efficiency |
|---|---|---|
| 101 to 507 | 1 to 100 | 47.0800 |
| 1 to 100 and 201 to 507 | 101 to 200 | 46.7600 |
| 1 to 200 and 301 to 507 | 201 to 300 | 45.9400 |
| 1 to 300 and 401 to 507 | 301 to 400 | 46.5500 |
| 1 to 400 and 501 to 507 | 401 to 500 | 44.8800 |

**Model-2: Prediction of protein sequence from protein secondary structures.**
**Observed state**: Protein secondary structure
**Hidden state**: Protein sequence

Table-2: Efficiencies of HMM model 2 for different training and test dataset

| Training dataset index | Test dataset index | Efficiency |
|---|---|---|
| 101 to 507 | 1 to 100 | 13.1000 |
| 1 to 100 and 201 to 507 | 101 to 200 | 12.8600 |
| 1 to 200 and 301 to 507 | 201 to 300 | 13.3100 |
| 1 to 300 and 401 to 507 | 301 to 400 | 13.6300 |
| 1 to 400 and 501 to 507 | 401 to 500 | 13.660 |

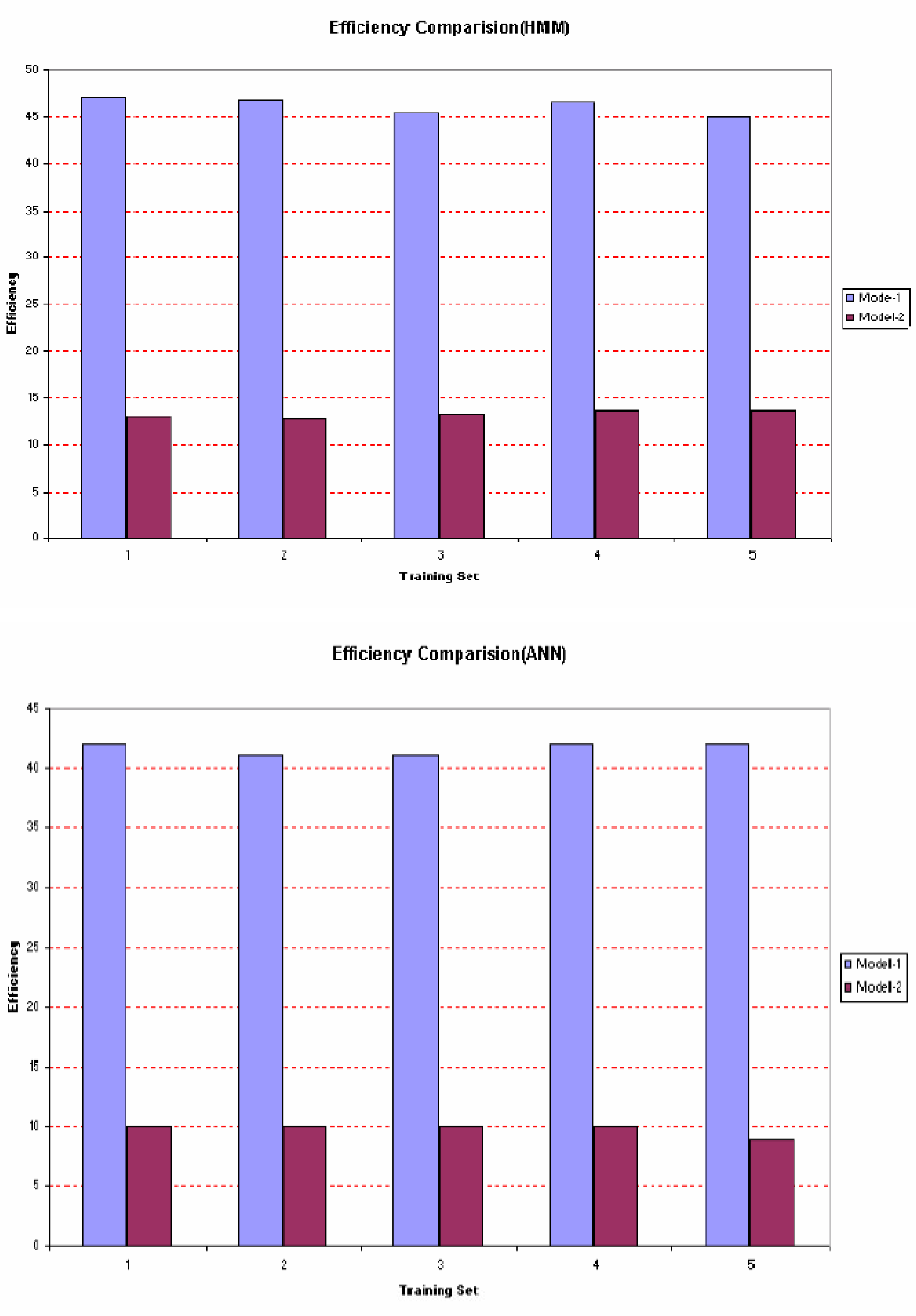

Figure (8) Efficiency comparison of HMM and ANN

## 7.2 DISCUSSION

From the work on HMM we find that the model designed on the basis of “hidden secondary structure giving rise to emission of observed protein primary structure i.e., its sequence” is giving better efficiency. From the HMM based model we get the following natural inferences:

1) To draw better efficiency in secondary structure prediction we should start with observed or known state as its primary structure i.e., its sequence. That means the protein sequence should be known (i.e., observed) to build such a prediction model.
2) The hidden pattern of protein secondary structure is in some way controlling the emission of observed pattern of protein primary structure (i.e., protein sequence).

To substantiate the above hypothesis, we performed an experiment with the help of simple feed forward ANN model of computing with starting point of two types:

i) Protein sequence is known (or observed in the jargon of HMM) which is fed into the ANN model as input AND protein secondary structure is unknown (or hidden in the jargon of HMM) which is fed into the ANN model as output.

ii) Reverse of the above.

From the work on ANN we find that the first type of starting point is giving significantly better efficiency than its second type – thus confirming the finding we obtained from HMM based model.
The two computing models employed by us are giving us the hint about the fact that “hidden pattern of protein secondary structure is organized earlier than its sequence” is more probable than its reverse one.

# IMPLEMENTATION & MAINTENANCE

## 8.1 TRAINING SETS

A **training set** is a set of data used in various areas of information science to discover potentially predictive relationships. Training sets are used in artificial intelligence, machine learning, genetic programming, intelligent systems, and statistics. In all these fields, a training set has much the same role and is often used in conjunction with a test set.
Training sets are used in supervised learning procedures in data mining (i.e. classification of records, or prediction of target values that are continuous.)

*Training set can be made easily directly from the time series. Certain number of measured values is used as inputs and the value to be predicted (i.e., the value in the future, in some chosen distance after these input measured values) is used as required output. Input part of the time series is called window, the output part is the predicted value.*

**Supervised vs. unsupervised learning**

From a theoretical point of view, supervised and unsupervised learning differ only in the causal structure of the model. In supervised learning, the model defines the effect one set of observations, called inputs, has on another set of observations, called outputs. In other words, the inputs are assumed to be at the beginning and outputs at the end of the causal chain. The models can include mediating variables between the inputs and outputs.

In unsupervised learning, all the observations are assumed to be caused by latent variables, that is, the observations are assumed to be at the end of the causal chain. In practice, models for supervised learning often leave the probability for inputs undefined. This model is not needed as long as the inputs are available, but if some of the input values are missing, it is not possible to infer anything about the outputs. If the inputs are also modelled, then missing inputs cause no problem since they can be considered latent variables as in unsupervised learning.

## 8.2 CONCLUSION

The pattern analysis above and the result shown there in above suggest that, the efficiency of the model-1 is better due to the fact that irrespective of the sequence variation the conservation is maintained in structural class of protein. The figure (8) shows the efficiency of both models over five classes of training sets, randomly generated from the total 507 dataset. In each class of training set of size 407, the model-1 outperforms model-2. This result support the fact that hidden pattern of protein secondary structure is organized earlier than its sequence and the patterns in sequence are less conserved in comparison to the structural pattern conservation so sometimes structures from the same family have less than 10% of sequence identity, yet are structurally similar.

Additionally, since structure is more conserved than sequence, structural data can be used to reconstruct many of the deeper evolutionary branches that would be difficult or impossible to determine with sequence data alone. This result can be supported in light of the structural constraints on different secondary structure elements, and their role in protein structural stabilization and topology.

## 8.2 FUTURE WORK

Only using the evolutionary information it is not possible to make any statistical model which can give enough protein secondary structure prediction accuracy on which we can rely on. Commonly the best accuracy given by these models is up to 70 %. This much accuracy is not good enough to use the predicted secondary structure for further analysis. So our prime object will be to improve the prediction accuracy by feeding the chemical properties of the amino acids to the model.
There can be no doubt that artificial neural networks are very effective at combining multiple non-linear factors, such as those affecting folding. They have been applied successfully to medical diagnosis, financial prediction, and face recognition. Given their success in other areas, it is very likely that with more research, neural nets will prove themselves worthy for protein structural prediction and will solve the protein folding problem.

**This page intentionally left blank**

# APPENDIX-A

## A SUBSET OF DATA SET (from 1 to 20):

```
seq{1}='MFKVYGYDSNIHKCVYCDNAKRLLTVKKQPFEFINIMPEKGVFDDEKIAELLTKLGRDTQIGLT
MPQVFAPDGSHIGGFDQLREYFK'
str{1}='UEEEEEUUTTTSUUHHHHHHHHHHHHHTTUUEEEEEESUSBTTBUUHHHHHHHHHHHHTUSUSSSUU
SUEEEUTTSUEEESHHHHHHHTU'
seq{2}='APAFSVSPASGASDGQSVSVSVAAAGETYYIAQCAPVGGQDACNPATATSFTTDASGAASFSFT
VRKSYAGQTPSGTPVGSVDCATDACNLGAGNSGLNLGHVALTFG'
str{2}='UUEEEEEEUUSSUUSSUEEEEEEEESUUSEEEEEEEEUEETTEEUUUTTTUUEEEUUSSUUUEEEEE
UUSEEEEEUTTSUEEEEEEETTTSUUEEEEEUSSUUUUUUUUUUU'
seq{3}='TPAFNKPKVELHVHLDGAIKPETILYFGKKRGIALPADTVEELRNIIGMDKPLSLPGFLAKFDY
YMPVIAGCREAIKRIAYEFVEMKAKEGVVYVEVRYSPHLLANSKVDPMPWNQTEGDVTPDDVVVNQGLQEGE
QAFGIKVRSILCCMRHQPSWSLEVLELCKKYNQKTVVAMDLAGDETIEGSSLFPGHVEAYEGAVKNGIHRTV
HAGEVGSPEVVREAVDILKTERVGHGYHTIEDEALYNRLLKENMHECPWSSYLTGAWDPKTTHAVVRFKNDK
ANYSLNTDDPLIFKSTLDTDYQMTKKDMGFTEEEFKRLNINAAKSSFLPEEEKKELLERLYREYQ'
str{3}='UUSUUSUEEEEEEEEGGGSUUHHHHHHHHHHHTUUUSUSSHHHHHHHHSUSSUUUHHHHTTGGGG
THHHHTTUHHHHHHHHHHHHHHHHHTTEEEEEEEEUSGGGUSSSUSSUGGGUUUUSUUHHHHHHHHHHHHHH
HHHUUEEEEEEEEEETTUTTTHHHHHHHHHHTBTTTEEEEEEEESUTTSTTGGGUHHHHHHHHHHHHHTUEEEE
EESSSSUHHHHHHHHHTTUUSEEEEUGGGGGSHHHHHHHHHTTUEEUHHHHHHTSSSUTTSUUHHHHHHHTT
UEEEEUUBUHHHHTUUHHHHHHHHHHHTTTUUHHHHHHHHHHHHHHTSUUUHHHHHHHHHHHHHHHTU'
seq{4}='SEACPLILDYHVALDNAREKARGAKAIGTTGRGIGPAYEDKVARRGLRVGDLFDKETFAEKLKE
VMEYHNFQLVNYYKAEAVDYQKVLDDTMAVADILTS'
str{4}='UTTUBBUUHHHHHHHHHHHHHHUSUUUUUSSTTHHHHHHHHHHTTUUUBGGGGSUHHHHHHHHHH
HHHHHHHHHHHTSUUUUUUHHHHHHHHHHHHHHHHT'
seq{5}='VPSLATISLENSWSGLSKQIQLAQGNNGIFRTPIVLVDNKGNRVQITNVTSKVVTSNIQLLLNT
RNI'
str{5}='UUUHHHHHHHHHHHHHHHHHHHHHHTTTTTEEEEEEEEEUUSSSSUEEEEEETTSHHHHHTBUUBUUG
GGU'
seq{6}='TPEMPVLENRAAQGNITAPGGARRLTGDQTAALRNSLSDKPAKNIILLIGDGMGDSEITAARNY
AEGAGGFFKGIDALPLTGQYTHYALNKKTGKPDYVTDSAASATAWSTGVKTYNGALGVDIHEKPTILEMAKA
AGLATGNVSTAELQDATPAALVAHVTSRKCYGPSATSQKCPGNALEKGGKGSITEQLLNARADVTLGGGAKT
FAETATAGEWQGKTLREEAEARGYQLVSDAASLNSVTEANQQKPLGFADGNMPVRWLGPKATYHGNIDKPAV
TCTPNPQRNDSVPTLAQMTDKAIELLSKNEKGFFLQVEGASIDKQDHAANPCGQIGETVDLDEAVQRALEFA
KKEGNTLVIVTADHAHASQIVAPDTKAGLQALNTKDGAVMVMSYGNSEEDSQEHTGSQLRIAAYGPHAANVV
GLTDQTDLFYTMKAALGLK'
str{6}='UUUUUUUUUUSUUSUTTSTTTTUUUSSUUHHHHHHHUUUSUUSEEEEEEEETTUUHHHHHHHHHH
HHUTTUUUTTGGGUUEEEEEEEUUEEUTTTUSEESSUUHHHHHHHHHHSUUUUTTUBSBUTTUUUUHHHHHHH
TTUEEEEEEEEEETTSHHHHTTTUUBSUTTUUSHHHHHHHUGGGUTTTTUSUUHHHHHHHHHUUSEEEEEUGGG
GGUBBSSGGGBTSBHHHHHHHHHTUEEEUSHHHHHHUUUUBTTBUEEUUSSSUUUSEEUUUUEETHHHHSUUB
UUEEUTTSUSSSUUHHHHHHHHHHHHHHTUTTUEEEEEEEUHHHHHHHTTUHHHHHHHHHHHHHHHHHHHHHH
HHHSSEEEEEEEUSSBUSUEEEUTTUUUSEEEEEUTTSSEEEEEEEUSUSSSSUUUBUUUEEEEEEESTTGGGGS
EEEEHHHHHHHHHHHHHTUU'
```

```
seq{7}='SAISFHSGYSGLVATVSGSQQTLVVALNSDLGNPGQVASGSFSEAVNASNGQVRVWR'
str{7}='UEEEEUTTSSSEEEEEEEUSSUEEEEEEESUUUUUGGGTUUSUUUEEEEEETTTTEEEEU'
seq{8}='MPPITQQATVTAWLPQVDASQITGTISSLESFTNRFYTTTSGAQASDWIASEWQALSASLPNAS
VKQVSHSGYNQKSVVMTITGSEAPDEWIVIGGHLDSTIGSHTNEQSVAPGADDDASGIAAVTERVLSENNFQ
PKRSIAFMAYAAEEVGLRGSQDLANQYKSEGKNVVSALQLDMTNYKGSAQDVVFITDYTDSNFTQYLTQLMD
EYLPSLTYGFDTCGYACSDHASWHNAGYPAAMPFESKFNDYNPRITQDTLANSDPTGSHAKKFTQLGLAYAI
EMGSATG'
str{8}='UUUUUUHHHHHHGGGUUHHHHHHHHHHHHHTSSUUUTTSHHHHHHHHHHHHHHHHHHHTTSTTEE
EEEEUUTTSUUUEEEEEEEUUSSEEEEEEEEEEEUUUUSSTTUUTTUUUUUTTTTHHHHHHHHHHHHHHHTTUU
USEEEEEEEEESUGGGTSHHHHHHHHHHHHHTTUEEEEEEEEUUUUSUUUSSSSEEEEUSSSUHHHHHHHHHHHH
HHUTTSUEEEEEUUSSUUSTHHHHHHHTTUUEEUEESSUGGGSUTTTSTUUGGGSUTTUHHHHHHHHHHHHHHH
HHHSUUU'
```

```
seq{9}='MYGNWGRFIRVNLSTGDIKVEEYDEELAKKWLGSRGLAIYLLLKEMDPTVDPLSPENKLIIAAG
PLTGTSAPTGGRYNVVTKSPLTGFITMANSGGYFGAELKFAGYDAIVVEGKAEKPVYIYIKDEEIRDASHIW
GKKVSETEATIRKEVGSEKVKIASIGPAGENLVKFAAIMNDGHRAAGRGGVGAVMGSKNLKAIAVEGSKTVP
I'
str{9}='UUSSUSEEEEEETTTTEEEEEEUUHHHHHHHUSHHHHHHHHHHHHSUTTSUTTSTTSUEEEEEU
TTTTSSSTTUUUEEEEEEUTTTSSEEEEEEUSSHHHHHHHHTUSEEEEESUUSSUEEEEEETTEEEEUTTTT
TUUHHHHHHHHHHHHTTUSSUEEEEUUHHHHTTUTTBUEEETTTEEEUSSSHHHHHHHTTEEEEEEEUUUUUU
U'
seq{10}='VAGGGLPKYGTAVLVNIINENGLYPVKNFQTGVYPYAYEQSGEAMAAKYLVRNKPCYACPIGC
GRVNRLPTVGETEGPEYESVWALGANLGINDLASIIEANHMCDELGLDTISTGGTLATAMELYEHIKDEELG
DAPPFRWGNTEVLHYYIEKIAKREGFGDKLAEGSYRLAESYGHPELSM'
str{10}='HHHTHHHHHUGGGHHHHHHHTTUUUBTTTTBSUUTTGGGGSHHHHHHHTEEEEEUUTTUSSUU
EEEEEETTTEEEEUUUHHHHHHHTGGGTUUUHHHHHHHHHHHHHHTBUHHHHHHHHHHHHHHHHSSUHHHHT
TSUUUUTTUTHHHHHHHHHHHTTUTTHHHHTTUHHHHHHTTTUGGGUU'
seq{11}='SQIRHYKWEVEYMFWAPNCNENIVMGINGQFPGPTIRANAGDSVVVELTNKLHTEGVVIHWHG
ILQRGTPWADGTASISQCAINPGETFFYNFTVDNPGTFFYHGHLGMQRSAGLYGSLIVDPPQGKP'
str{11}='UUEEEEEEEEEEEEEEUTTSSUEEEEEEETTBSSUUUEEEEETTUEEEEEEEEEUUSSUUBUEEEET
UUUTTUGGGSUUBTTTBUUBUTTUEEEEEEEEUUSUEEEEEEEEUSTTGGGGTUEEEEEEEEUSTTUS'
seq{12}='FHYDGEINLLLSDWWHQSIHKQEVGLSSKPIRWIGEPQTILLNGRGQFDCSIAAKYDSNLEPC
KLKGSESCAPYIFHVSPKKTYRIRIASTTALAALNFAIGNHQLLVVEADGNYVQPFYTSDIDIYESYSVLIT
TDQNPSENYWVSVGTRARHPNTPPGLTLLNYLPNSVSKLPTSPPPQTPAWDDFDRSKNFTYRITAAMGS'
str{12}='SUUSEEEEEEEEEEEUSSUHHHHHHHHHSSSUUUUUSUSEEEEETTBUUSSSBTTGGGUTTSUBU
UUUSUSTTSUUUEEEUTTUEEEEEEEEEUUSSUEEEEEEETTBUEEEEEEETTEEEEEEEEESUEEEUUEEEEEEE
UUSUTTSUEEEEEEEEESSUUUSUUEEEEEEEETTSUTTUUUSSUUUUUUUTTUHHHHHHHHHHUUBUTTS'
seq{13}='PKPPVKFNRRIFLLNTQNVINGYVKWAINDVSLALPPTPYLGAMKYNLLHAFDQNPPPEVFPE
DYDIDTPPTNEKTRIGNGVYQFKIGEVVDVILQNANMMKENLSETHPWHLHGHDFWVLGYGDGKAEEESSLN
LKNPPLRNTVVIFPYGWTAIRFVADNPGVWAFHCHIEPHLHMGMGVVFAEGVEKVGRIPTKALACGGTAKSL
INNPKNP'
str{13}='UUUUSSUSEEEEEEEEEEEEEETTEEEEEEETTEEEUUUSSUHHHHHHTTUTTSSUUSUUUUUUUT
TUUTTSUUUUTTUEEEUUUEEEUTTUEEEEEEEEUUUUSSTTUUUUEEEEEETTUUEEEEEEEEESSGGGGGGSU
SSSUUEESEEEEUTTEEEEEEEEEUUSUEEEEEEEESSHHHHHTTUEEEEEEEUGGGUUUUUHHHHUSHHHHHHH
SUUUSUU'
seq{14}='REPRGLGPLQIWQTDFTLEPRAPRSWLAVTVDTASSAIVVTQHGRVTSVAAQHHWATAIAVLG
RPKAIKTDNGSCFTSKSTREWLARWGIAHTTGIPGNSQGQAVERANRLLKDKIRVLAEGDGFKRTSKQGELL
AKAYALNHFER'


str{14}='UUUSTTUUUUEEEEEEEEEUUUUUSUEEEEEEEETTTUUEEEEEEESSUUHHHHHHHHHHHHHHHU
UUSEEEUUSUHHHHSHHHHHHHHHHHTUEEEUUSSSUSUUUUUHHHHHHHHHHHHHHHHHHTTUUUHHHHHHHH
HHUUTTTUUUU'
seq{15}='TTYADFIASGRTGRRNAIHD'
str{15}='UHHHHHHTSSUUSUUUUUUU'
seq{16}='RDPDAGIDEAQVEQDAQALFQAGELKWGTDEEKFITIFGTRSVSHLRKVFDKYMTISGFQIEE
TIDRETSGNLEQLLLAVVKSIRS'
str{16}='UUUUUUUUHHHHHHHHHHHHHHHHTTTUSUHHHHHHHHHHHSUHHHHHHHHHHHHHHHHSSUGGG
GSUTTTUUHHHHHHHHHHHHHHU'
seq{17}='IPAYLAETLYYAMKGAGTDDHTLIRVMVSRSEIDLFNIRKEFRKNFATSLYSMIKGDTSGDYK
KALLLLCGEDD'
str{17}='HHHHHHHHHHHTSSSSSUUHHHHHHHHHHHTTTTTHHHHHHHHHHHHHSSUHHHHHHHHHUUTHHH
HHHHHHHUSUU'
seq{18}='MRRWFHPNITGVEAENLLLTRGVDGSFLARPSKSNPGDFTLSVRRNGAVTHIKIQNTGDYYDL
YGGEKFATLAELVQYYMEHHGQLKEKNGDVIELKYPLN'
str{18}='UUTTBUSSUUHHHHHHHHHHHTUUTTEEEEEEEUSSSTTUEEEEEEEETTEEEEEEEEEEUSSUEES
SSSSUBSSHHHHHHHHHHHSTTUEEBTTUUEEUUUEEUU'
seq{19}='CSVDIQGNDQMQFNTNAITVDKSCKQFTVNLSHPGNLPKNVMGHNWVLSTAADMQGVVTDGMA
SGLDKDYLKPDDSRVIAHTKLIGSGEKDSVTFDVSKLKEGEQYMFFCTFPGHSALMKGTLTLK'
str{19}='UUEEEEUUSSSUUSUSSEEUUSSSSEEEEEEEEUUSSUUTTTSUBUUEEEETTTTHHHHHHHHH
HUHHHHSSUSSUTTUSBUUUUBUTTUUEEEEEEESSSUUSSUUEEEEEUUSTTTTTTSEEEEEUU'
seq{20}='MEVEKEFITDEAKELLSKDKLIQQAYNEVKTSICSPIWPATSKTFTINNTEKNCNGVVPIKEL
CYTLLEDTYNWYREKPIDVYKEFIENSELKRVGMEFETGNISSAHRSMNKLLLGLKHGEIDLAIMPIKQLAY
YLTDRVTNFEELEPYFELTEGQPFIFIGFNAEAYNSNVPLIPKGSDGMSKRSIKKWKDKVENK'
```

```
str{20}='UEEEEEEEUHHHHHHTTUHHHHHHHHHHHHHHHUUUBSTTUSSBEEEUSSTTUBUSHHHHHH
HHHHHHHHHUUEEUUUUSEEEEEETTEEEEEEEEEUUSUHHHHHHHHHHHHHHHHTTSUSEEEEEUHHHHT
TBUSSUUUHHHHGGGGGGTTSUEEEEEEEUSEEETTSUUUUUSTTSSUHHHHHHHHHHHHTU'
```